\documentclass{article}

\usepackage[final,nonatbib]{neurips_data_2024}

\usepackage[utf8]{inputenc} 
\usepackage[T1]{fontenc}    
 \usepackage[colorlinks=true,linkcolor=red,citecolor=green]{hyperref}
\usepackage{url}            
\usepackage{booktabs}       
\usepackage{amsfonts}       
\usepackage{nicefrac}       
\usepackage{microtype}      
\usepackage{xcolor}
\usepackage{pifont}
\usepackage{multirow}
\usepackage{macros}
\usepackage{graphicx}
\usepackage{subcaption}
\usepackage{tabularx}
\usepackage{xspace}

\usepackage{algorithm}
\usepackage{algorithmicx}  
\usepackage[noend]{algpseudocode}
\usepackage{wrapfig}
\usepackage{listings}
\usepackage{forest}
\usepackage{adjustbox}

\usepackage{framed} 
\usepackage{enumitem}

\usepackage{tikz}
\newcommand*\circled[1]{\tikz[baseline=(char.base)]{
            \node[shape=circle,draw,inner sep=2pt] (char) {#1};}}

\definecolor{codegreen}{rgb}{0,0.6,0}
\definecolor{codegray}{rgb}{0.5,0.5,0.5}
\definecolor{codepurple}{rgb}{0.58,0,0.82}
\definecolor{backcolour}{rgb}{0.95,0.95,0.92}

\definecolor{tablegreen}{rgb}{0.9,0.99,0.9}
\definecolor{tablered}{rgb}{0.99,0.9,0.9}
\definecolor{darkgreen}{rgb}{0.0, 0.5, 0.0}

\lstdefinestyle{mystyle}{
    backgroundcolor=\color{backcolour},   
    commentstyle=\color{codegreen},
    keywordstyle=\color{magenta},
    numberstyle=\tiny\color{codegray},
    stringstyle=\color{codepurple},
    basicstyle=\ttfamily\footnotesize,
    breakatwhitespace=false,         
    breaklines=true,                 
    captionpos=b,                    
    keepspaces=true,                 
    numbers=left,                    
    numbersep=5pt,                  
    showspaces=false,                
    showstringspaces=false,
    showtabs=false,                  
    tabsize=2
}

\newcommand{\name}{\textsc{Task-Me-Anything}\xspace}
\newcommand{\nameone}{\name-\textsc{v1.0}\xspace}
\newcommand{\dbname}{\textsc{\name-DB}\xspace}
\newcommand{\uiname}{\textsc{\name-UI}\xspace}
\newcommand{\randomname}{\textsc{\name-Random}\xspace}
\newcommand{\benchname}{\textsc{\name-2024}\xspace}

\newcommand{\twod}{\textit{2D sticker image}\xspace}
\newcommand{\threed}{\textit{3D tabletop scene}\xspace}
\newcommand{\sg}{\textit{Scene Graph}\xspace}

\newcommand{\instructblip}{\textsc{InstructBLIP}\xspace}
\newcommand{\instructblips}{\textsc{InstructBLIP-7B}\xspace}
\newcommand{\instructblipl}{\textsc{InstructBLIP-13B}\xspace}
\newcommand{\qwenvl}{\textsc{Qwen-VL}\xspace}
\newcommand{\qwenvlchat}{\textsc{Qwen-VL-Chat}\xspace}
\newcommand{\llava}{\textsc{LLaVA}\xspace}
\newcommand{\llavas}{\textsc{LLaVA-7B}\xspace}
\newcommand{\llaval}{\textsc{LLaVA-13B}\xspace}

\newcommand{\videollamatwo}{\textsc{Video-LLaMA-2}\xspace}
\newcommand{\videollamatwos}{\textsc{Video-LLaMA-2-7B}\xspace}
\newcommand{\videollamatwol}{\textsc{Video-LLaMA-2-13B}\xspace}
\newcommand{\videochatgpt}{\textsc{Video-ChatGPT}\xspace}
\newcommand{\videochatgpts}{\textsc{Video-ChatGPT-7B}\xspace}
\newcommand{\chatunivi}{\textsc{Chat-UniVi}\xspace}
\newcommand{\chatunivis}{\textsc{Chat-UniVi-7B}\xspace}
\newcommand{\chatunivil}{\textsc{Chat-UniVi-13B}\xspace}
\newcommand{\videollava}{\textsc{Video-LLaVA}\xspace}
\newcommand{\videollavas}{\textsc{Video-LLaVA-7B}\xspace}
\newcommand{\videochattwo}{\textsc{VideoChat2}\xspace}
\newcommand{\videochattwos}{\textsc{VideoChat2-7B}\xspace}

\newcommand{\glmfourv}{\textsc{GLM-4v}\xspace}
\newcommand{\cogvlmtwo}{\textsc{CogVLM2-19B}\xspace}
\newcommand{\ideficstwo}{\textsc{Idefics2-8B}\xspace}
\newcommand{\phivision}{\textsc{Phi-3-vision-3B}\xspace}
\newcommand{\paligemma}{\textsc{PaliGemma-3B}\xspace}

\newcommand{\geminipro}{\textsc{Gemini-Pro}\xspace}
\newcommand{\gptfourv}{\textsc{GPT4V}\xspace}
\newcommand{\gptfouro}{\textsc{GPT4o}\xspace}
\newcommand{\qwenvlmax}{\textsc{Qwen-VL-Max}\xspace}
\newcommand{\internvlchat}{\textsc{InternVL-Chat-1.5-24B}\xspace}
\newcommand{\llavanext}{\textsc{LLaVA-Next-34B}\xspace}

\usepackage[colorinlistoftodos,textsize=tiny]{todonotes}

\newcommand{\goal}[1]{\textbf{G#1}}

\newcommand*\bluecircled[1]{{\color{blue}{\tikz[baseline=(char.base)]{
            \node[shape=circle,draw,inner sep=2pt] (char) {#1};}}}}
\newcommand*\greencircled[1]{{\color{darkgreen}{\tikz[baseline=(char.base)]{
            \node[shape=circle,draw,inner sep=2pt] (char) {#1};}}}}
\newcommand*\redcircled[1]{{\color{red}{\tikz[baseline=(char.base)]{
            \node[shape=circle,draw,inner sep=2pt] (char) {#1};}}}}
\newcommand*\orangecircled[1]{{\color{orange}{\tikz[baseline=(char.base)]{
            \node[shape=circle,draw,inner sep=2pt] (char) {#1};}}}}

\title{Task Me Anything}

\author{%
Jieyu Zhang$^{1}$, 
Weikai Huang$^{1}$\thanks{\ The authors contribute equally to this work.}, 
Zixian Ma$^{1}$\footnotemark[1], 
Oscar Michel$^{2}$, 
Dong He$^{1}$, \\\bf
Tanmay Gupta$^{2}$, 
Wei-Chiu Ma$^{2}$, 
Ali Farhadi$^{1,2}$,
Aniruddha Kembhavi$^{2}$, 
Ranjay Krishna$^{1,2}$\\
$^1$University of Washington, 
$^2$Allen Institute for Artificial Intelligence \\
\centerline{\url{https://www.task-me-anything.org}}
}

\begin{document}

\maketitle

\begin{abstract}

Benchmarks for large multimodal language models (MLMs) now serve to simultaneously assess the general capabilities of models instead of evaluating for a specific capability.
As a result, when a developer wants to identify which models to use for their application, they are overwhelmed by the number of benchmarks and remain uncertain about which benchmark's results are most reflective of their specific use case.
This paper introduces \name, a benchmark generation engine which produces a benchmark tailored to a user's needs. \name maintains an extendable taxonomy of visual assets and can programmatically generate a vast number of task instances. 
Additionally, it algorithmically addresses user queries regarding MLM performance efficiently within a computational budget. 
It contains $113$K images, $10$K videos, $2$K 3D object assets, over $365$ object categories, $655$ attributes, and $335$ relationships. It can generate $750$M image/video question-answering pairs, which focus on evaluating MLM perceptual capabilities.
\name reveals critical insights: open-source MLMs excel in object and attribute recognition but lack spatial and temporal understanding; 
each model exhibits unique strengths and weaknesses; 
larger models generally perform better, though exceptions exist; 
and \gptfouro demonstrates challenges in recognizing rotating/moving objects and distinguishing colors.

\end{abstract}

\section{Introduction}
\label{sec:intro}

Benchmarks in computer vision have traditionally served to evaluate progress towards important research problems. They shepherd the research community's attention towards a specific capability by providing reproducible evaluation protocols to identify the best solution.
For example, the NYUv2 benchmark has served to identify the best model for depth estimation for the last decade~\cite{silberman2012indoor}.
In a surprising twist, the role of recent benchmarks has shifted with the advent of general-purpose large multimodal language models (MLMs)~\cite{chatgpt, gpt-4v}. 
This shift has similarly led to the curation of general-purpose benchmarks that assess the diversity of capabilities and not any one single capability~\cite{liu2023mmbench,yue2023mmmu,li2023seed, li2023seed2, fu2024blink,li2023mvbench,fan2024muffin,prabhu2024lance,lu2023mathvista}.
As a result, they are now less informative to the communities they are meant to serve—researchers, developers, and users.

When a developer wants to identify which models to use for their application, they remain uncertain about which benchmark results are most aligned with their specific use case.
Consider a scenario where an application developer needs a model that can most accurately identify object shapes. They may find there are existing datasets such as SHAPES~\cite{andreas2016deep} and CLEVR~\cite{johnson2017clevr} that contain shape-related task instances, yet the involved objects are simple geometric primitives instead of objects in the real world.
Similarly, consider a team of researchers at a big technology corporation hoping to identify the limitations of their proprietary MLM.
Although MLMs are released with evaluations on benchmarks like MMBench, MMMU, BLINK and SeedBench~\cite{liu2023mmbench,yue2023mmmu,li2023seed,li2023seed2,fu2024blink}, their performance across these holistic benchmarks do not pinpoint which fine-grained capabilities are lacking.

There is a need for a principled benchmark generation process that answers task-specific user queries: ``(\textbf{Q1}) Which model is the best at recognizing the shape of objects?'' or ``(\textbf{Q2}) what are the model's weaknesses that we can further improve on?''. 
To actualize such a process, there are several challenges.
First, we need to define an extendable taxonomy to represent the space of inputs and outputs. For example, to answer Q1, the taxonomy must include objects and their shapes. This taxonomy should be easily extendable so that future queries can evaluate new concepts.
Second, the process must be able to curate a sufficient number of input-output evaluation pairs given a user query. To answer Q1, it must be able to generate thousands of images containing objects with their known shapes.
Third, evaluating MLMs is computationally expensive, so the evaluation process should estimate an MLM's performance given a computation budget.

\begin{figure}[!t]
\centering
\includegraphics[width=\linewidth]{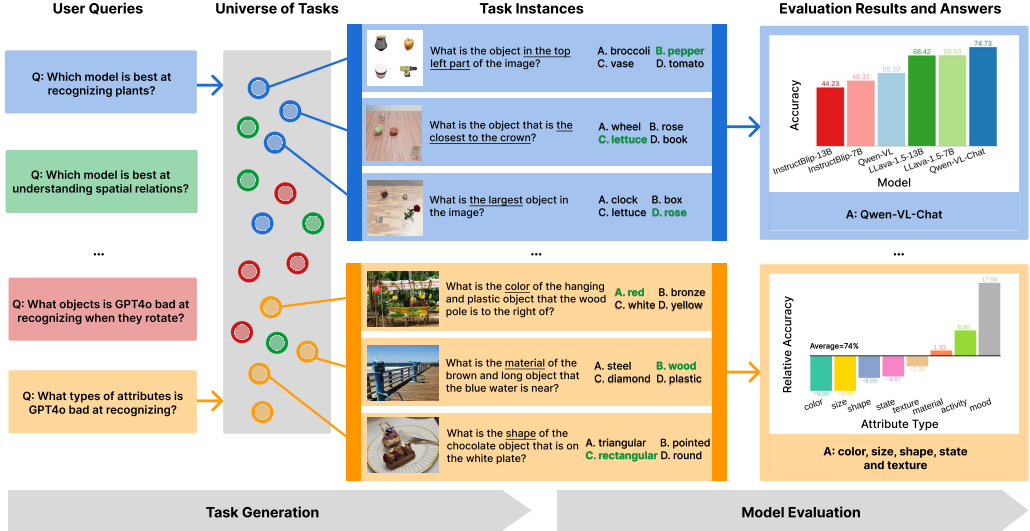}
\caption{We present examples of user queries, corresponding task instances generated by \name as well as the evaluation results on them that answer the queries.}
  \label{fig:teaser}
\end{figure}

We present \name, a benchmark generation engine that curates a custom benchmark given a user query (Figure~\ref{fig:teaser}).
First, \name maintains a extendable taxonomy with corresponding visual assets (\eg~images with scene graphs~\cite{krishna2017visual}, 3D object assets~\cite{deitke2024objaverse}, videos with spatio-temporal annotations~\cite{ji2020action}, rendering softwares~\cite{blender}, \etc.). 
It is implemented as an extendable library where new concepts and their corresponding assets and annotations can be easily added.
Second, \name contains programmatic task generators which sub-select from the taxonomy to curate a large number of input-output pairs.
Image/videos are either from existing datasets or programmatically generated with specific configurations.
With our current taxonomy, \name can generate over $750$ million tasks. In comparison, existing benchmarks for MLMs have fewer task instances: MME (2,194), MMBench (3,217), BLINK (3,807), MMMU (11,550), SeedBench (19,242). 
Programmatic task generation is not new—CLEVR~\cite{johnson2017clevr} and GQA~\cite{hudson2019gqa} were also programmatically generated. While their contribution is the final generated benchmark, our contribution is the benchmark generation process itself.
Third, \name allows users to specify a computation budget. It contains algorithms to approximate the results of user queries via predicting the model performance across a large number of input-output pairs without actually invoking the MLM on each task instance.

The current version of \name's library contains $122,866$ scene graphs~\cite{hudson2019gqa, GrundeMcLaughlin2021AGQA} associated with $113,018$ real images and $9,848$ real videos, $1,996$ 3D object assets~\cite{deitke2023objaverse, deitke2024objaverse} with manual annotations, can curate $28$ types of tasks (counting ``how many $\ldots$?'', color questions ``what color $\ldots$?'', etc.), $365$ object categories, $335$ relationships, $655$ attributes, and $14$ spatial positions. 
With this, we extensively evaluate $13$ open-source MLMs over 1M task instances and $18$ open-source/proprietary MLMs over $8,400$ task instances, both generated by \name. 
We then address the following questions:
(1) ``What perceptual capabilities do open-sourced MLMs still lack?'';
(2) ``Do all models lack the same perceptual capabilities?'';
(3) ``Do larger (or proprietary) models always exhibit superior perceptual capabilities than smaller (or open-source) ones?'';
(4) ``What specific capabilities does \gptfouro, the recently introduced proprietary MLM, still lack?''.

Our analyses produce the following takeaways: 
(1) open-sourced MLMs exhibit strong object and attribute recognition abilities but struggle at counting, spatial and temporal understanding.
(2) while most models perform similarly across different capabilities, individual models showcase different strengths and weaknesses (\eg,  \qwenvlchat is good at spatial relation understanding whereas \instructblips is exceptionally good at understanding emotional relations). 
(3) Larger MLMs do tend to perform better than smaller ones with a few exceptions (\eg, \instructblips outperforms \instructblipl on relation understanding).
(4) The best open-source MLM is on par with if not better than the best proprietary model across skills, with a nontrivial performance margin up to 7 and 8\% on spatial and 3D attribute understanding.
(5) We found that recognizing rotating/moving ``furniture'', ``food'', and ``plants'' is more challenging for \gptfouro than for other object categories like animals and vehicles, likely because these objects are typically static in the real world, and \gptfouro struggles more with distinguishing colors than other attributes.

\section{\name}
\label{sec:sys-deign}

\begin{figure}[!t]
  \centering
  \includegraphics[width=\linewidth]{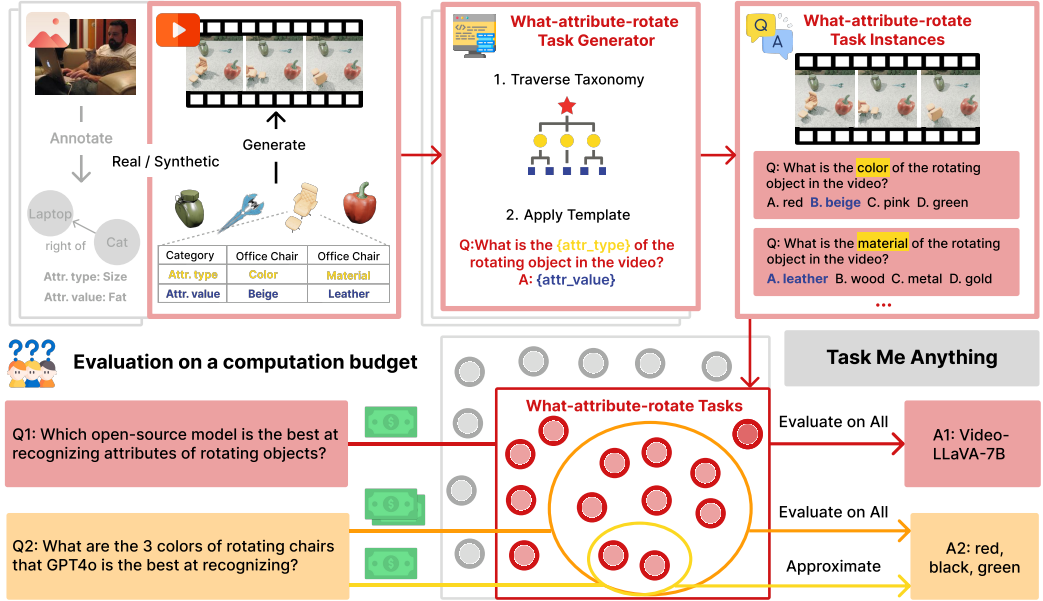}
  \caption{We present the key components in \name. The top part illustrates the task generation process with an example video synthesized with 3D objects and their annotations, and the task generator for generating questions about rotating objects' attributes. The bottom part depicts the model evaluation process, which selects the relevant tasks based on the user's query and their budget and performs either full evaluation or results approximation to answer the query.  }
  \label{fig:generation-process}
\end{figure}

Consider a user who wants to know ``Which open-sourced MLM is best at recognizing objects even if the object is rotating?''. \name provides an interface for the user to pose such questions and provides them with an answer (Figure~\ref{fig:generation-process}). 
It contains a taxonomy to symbolically represent visual content. A query identifies the relevant portion of the Taxonomy required to answer the query.
It also contains task generators that create input-output pairs that test for a specific capability. The Taxonomy subset is used to select the appropriate task generator.
We adopt the common input-output format used in existing benchmarks, \ie, all the task instances in \name contain an image/video, a question, and multiple options with one ground truth answer. 
MLMs will be evaluated on these generated task instances and the results will be returned back to the user.
Finally, it also supports queries that ask for, not just the best performing model, but also task instances (``Find top-10 task instances that \gptfouro performs the worst'') or taxonomy concepts (``Find the objects that \gptfouro's performance is higher than a threshold''), as well as on-budget results approximation methods for such fine-grained queries.  
unlike most existing procedural data systems, we design \name so that the generation space of tasks can be expanded by adding new source data and/or task generator code.
More details in Appendix~\ref{app:task-generation} and~\ref{app:query}. 

\subsection{Taxonomy} 
We adopt a spatio-temporal scene graph as a representation of concepts represented in an image or video~\cite{krishna2017visual,ji2020action}. In a scene graph, objects and their corresponding attributes are nodes and relationships between objects are edges. Scene graphs have already been utilized in programmatic generation of VQA task instances in datasets like GQA~\cite{hudson2019gqa} and AGQA~\cite{GrundeMcLaughlin2021AGQA,gandhi2022measuring}.
For example, the object nodes of the scene graph can be used to create counting tasks, relationships edges can encode relative locations and generate spatial understanding tasks, and attributes can ask about color, material, physical states like rotation, etc. 
The scene graph representation is generic: it can be extended to incorporate concepts like lightning conditions and ask questions about the light source, illumination, and shadows~\cite{bashkirova2023lasagna}.
In fact, we extend traditional scene graphs with 3D object assets from Objaverse~\cite{deitke2023objaverse, deitke2024objaverse}, enabling us to ask questions about any objects with available 3D models and their spatial positions, \etc.

\subsection{Task generators} 
A task generator is a Python program that can generate VQA task instances given a subset of the taxonomy. It generates questions using templates of the type: “How many <target object> are there in the image?”, where the <target object> can be filled with objects in the scene graph such as “telephone”.
Also, it programmatically produces the ground truth answer based on the scene graph. It synthesizes incorrect yet plausible options for each question~\cite{zellers2019hellaswag}.
For the visual input associated with every question, we use the images~\cite{hudson2019gqa} and videos~\cite{GrundeMcLaughlin2021AGQA} annotated with scene graphs. However, scene graph data is expensive and therefore, limited.
To facilitate diverse user queries, we programmatically generate images/videos from scene graph representations~\cite{cascante2022simvqa,andreas2016deep}.
Since image/video generation models can introduce potential errors into our evaluation pipeline, we leave the use of generative models to future work.
Instead, we programmatically generate image/video layouts and render them using Blender~\cite{blender} with 3D object models~\cite{deitke2023objaverse, deitke2024objaverse} via the following two approaches:
1) \twod (abbreviated to 2D): Inspired by the SHAPES dataset~\cite{andreas2016deep}, we position individual 2D rendering images of 3D object models in a grid (either 2x2 or 3x3) to compose an image, which is fast to generate but lack realism, \eg, plausible object co-occurrences, lighting, shadows, \etc are absent. 
and 2) \threed (abbreviated to 3D): To overcome the limitations of the 2D approach, we render tabletop scenes to generate images after placing the 3D object assets on the table~\cite{michel2024object}. 
Similarly, we generate videos and adjust the position and angle of the objects across different key frames to make objects move and rotate.
Such rendered images/videos are more realistic since Blender also supports lightning and collision controls.

Concretely, we use the term \emph{task plan} to refer to the ingredients that a task generator requires for task generation, which contain the necessary task metadata and configurations.
For example, in tasks involving counting, the task plan specifies the categories of objects, their total numbers in the scene, and their positions in the image—such as two apples, one on the top right and one on the bottom left. 
The \emph{task instance} then features an actual image/video, question, options, and ground truth answer tuple that comprises a single evaluation test case and is generated by a task generator with a specific task plan.
One such task instance might be an image with two apples, the question: “How many apples are there in the image?”,  and the answer: “2”. Multiple task instances can be generated from a single task plan because other elements such as the image background and types of distractor objects can be randomized, as they are not specified in the task plan.
We refer to this family of task instances that can be generated by a task generator with a specific task plan as \emph{task class} or \emph{task}, a conceptual abstraction of all task instances derived from the same task plan.
Finally, each task generator is implemented for a specific type of task, \eg, the 2D how-many task generator is for generating counting tasks with \twod images and has two major functionalities.
First, it should define the schema of the task plan it can input and be able to enumerate all the possible task plans given the source data, \eg, the 3D object models and annotations; Second, it can generate concrete task instances given a valid task plan.

\subsection{Addressing user queries} 
\label{sec:query}

Given the millions of task instances that \name can generate, it is computationally infeasible to evaluate even a single model on the entire task space. It would also take too long to be useful for everyday users.
We describe how \name supports on-demand task generation and evaluation to address user queries.

Because each task generator supports generating all the task plans without generating the actual task instances and these task plans, once pre-computed, can act as a structured representation of the task space, users can leverage them to identify tasks relevant to their queries and then opt to only generate and evaluate the models on the relevant tasks.
For example, image a user query "Which open-source model is the best at recognizing shapes of rotating objects?", the user can leverage the what-attribute-rotate task generator to compute all the task plans, select those related to recognizing shapes and then use them to generate actual task instances to evaluate and compare open-source models.
Such a workflow enables \emph{query-centric} task generation and evaluation, avoiding generating and evaluating the entire task space.

\paragraph{Fine-grained user queries.} 
While many user queries can be simply addressed by the aforementioned workflow, 
we additionally support 4 types of fine-grained user queries for investigations regarding individual tasks and taxonomy concepts:

\noindent\circled{1} \textit{Top-K queries} enable users to request the top-K taxonomy concepts or tasks (\eg, ``Return the top-10 colors/tasks that \llaval struggles with'').

\noindent\circled{2} \textit{Threshold queries} allows users to query for taxonomy concepts or tasks where model performance surpasses or falls below a given threshold (\eg, ``Find all the object recognition tasks that both \llavanext and \gptfouro perform below 30\% accuracy?'').

\noindent\circled{3} \textit{Model comparison queries} identify where one model outperforms another by a specified margin, enabling comparative analysis (\eg, ``Which types of tasks does \gptfouro outperform \geminipro?'').

\noindent\circled{4} \textit{Model debugging queries} identify where a model's performance deviates from its average by one standard deviation, facilitating the ability to uncover models' inconsistent behavior. (\eg, What action does \videollamatwos struggle to recognize compared to other actions?). 

\paragraph{Addressing fine-grained queries under a budget.}
These fine-grained user queries might involve a large number of tasks to generate and evaluate to obtain query results. For example, to obtain the top K tasks of a task generator that model performs the worst, we have to evaluate all the possible tasks.
To address this, we draw on active learning literature~\cite{karamcheti2021mind}, to implement 3 efficient query results approximation approaches for these fine-grained user queries:

\noindent\circled{1} \textit{Random} randomly samples a subset of task instances from the total possible for that query. MLMs are evaluated on only this subset.

\noindent\circled{2} \textit{Fitting} similarly samples a random subset and evaluates MLMs. The results are used to train an efficient function approximator for each MLM. This function approximator learns to predict an MLM's performance on a task, by featurizing the task plan—never actually generating the task instance itself. While many model choices are applicable, we adopt the Gaussian Process regressor throughout this work since it renders stable performance in preliminary studies. It uses this function to approximate the MLM's performance on the remaining task space.

\noindent\circled{3} \textit{Active} is similar to \textit{fitting} but iteratively trains each function approximator using active learning. Given a smaller subset, it trains an initial function, which is then used to sample the most task instances that are most likely to lead to improved approximation results. MLMs are evaluated on these samples; the results are used to \textit{re-train} the functions again.

\begin{figure}[!t]
  \centering
  \includegraphics[width=\linewidth]{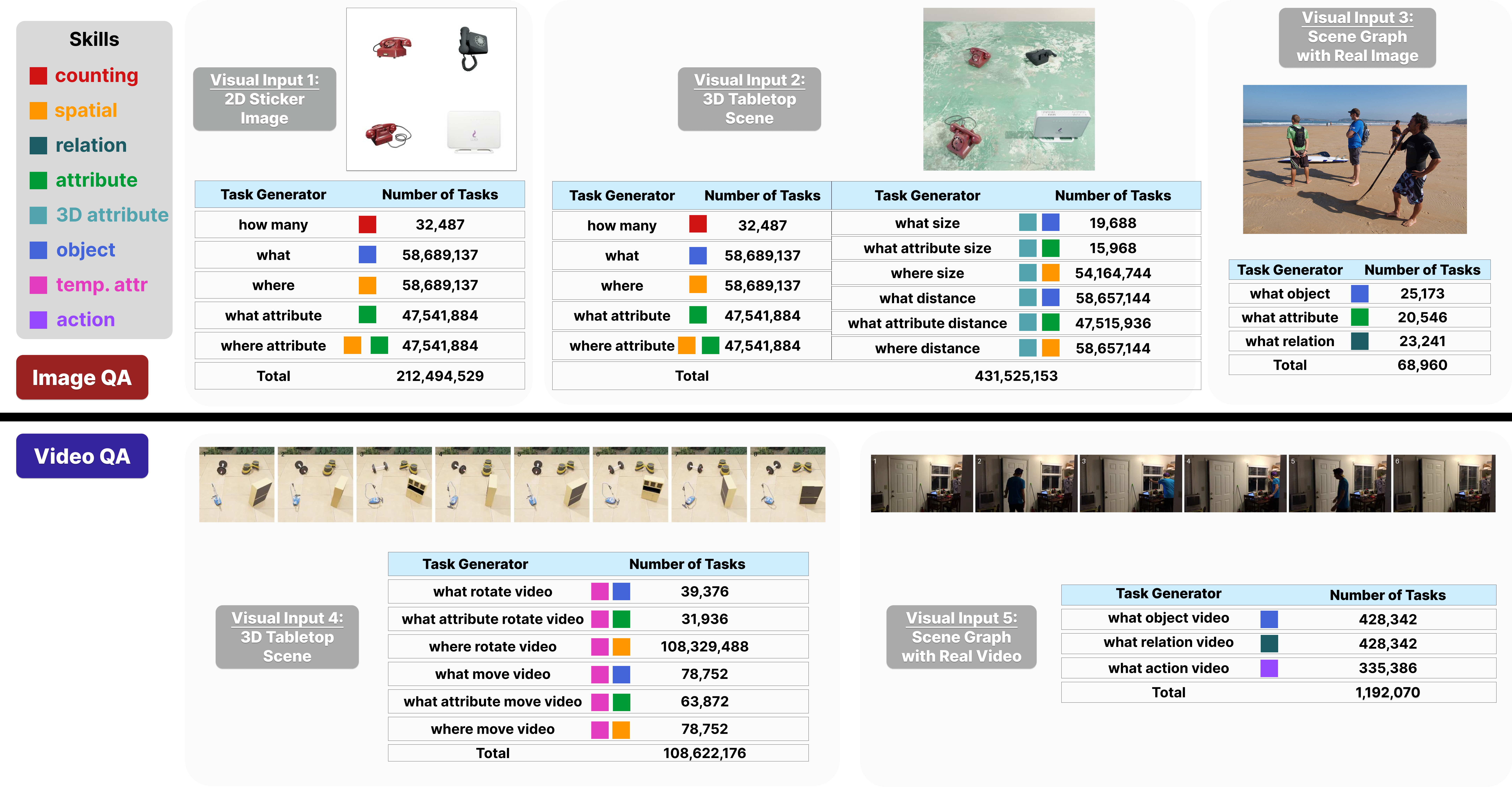}
  \caption{The statistics of generatable tasks of each task generator and example image/video in \name. We each task generator with high-level perceptual skills and this collection of task generators can collectively generate over 750M VQA tasks.}
  \label{fig:task-generator-stats}
\end{figure}

\subsection{Final benchmark engine}
Although \name supports many different kinds of reasoning tasks, it currently focuses on visual perception capabilities. 
We include $28$ different task templates across $5$ types of visual inputs: 2D sticker images (2D), 3D tabletop scene images/videos (3D), and real images/videos with manually-annotated scene graphs. In total, it can generate over $750$ million possible VQA task instances (see Figure~\ref{fig:task-generator-stats} for a breakdown).
We draw image scene graphs from Visual Genome~\cite{krishna2017visual}, and video spatio-temporal scene graphs from Action Genome~\cite{ji2020action}. We also include GQA~\cite{hudson2019gqa} and AGQA~\cite{GrundeMcLaughlin2021AGQA} for their real VQA instances.
For 2D and 3D scenes, we select $1,996$ high-quality 3D objects across $337$ categories from Objaverse-LVIS, the subset of Objaverse 1.0~\cite{deitke2023objaverse} that has been annotated with LVIS~\cite{gupta2019lvis} categories. Each 3D object was manually annotated with attributes such as color, material, shape, and visible angles.
More details can be found in Appendix~\ref{app:system-1.0}.

These $28$ different task generators provide a comprehensive way to evaluate visual understanding capability including object recognition, attribute recognition, relation recognition, localization, spatial reasoning, temporal reasoning, action recognition, \etc (Figure~\ref{fig:task-generator-stats}).
With this diversity of potential questions, \name supports the evaluation at varying desired levels of granularity 

For model users, \name can help decide which model to use for their needs, and for model developers, it can identify the weaknesses of models to improve. For example, a model user wanting to find the best model for distinguishing different breeds of dogs can query: ``What are the top 3 models for distinguishing dogs?'' Similarly, a model developer might query: ``Find the spatial reasoning capabilities that all models lack?'' to identify some general issues in current architecture. Or they might also query: ``Which types of materials do \llava underperform on?'' and then add the corresponding data into training to enhance \llava's material recognition performance. 

This system is not only versatile but also scalable. By adding new task generators, assets like 3D object models, and software like Blender, DALL-E, etc., we can continuously expand its taxonomy. Updating a taxonomy of underlying capabilities is more scalable than collecting sufficient data for the rapid growth in use-cases for MLMs.

\section{Evaluating MLMs using \name}
\label{sec:eval}

In this work,  we extensively evaluate $13$ open-source MLMs over 1M task instances and $18$ open-source/proprietary MLMs over $8,400$ task instances, both generated by \name, for validating \name and analyses.

\begin{figure}[h]
  \centering
  \includegraphics[width=\linewidth]{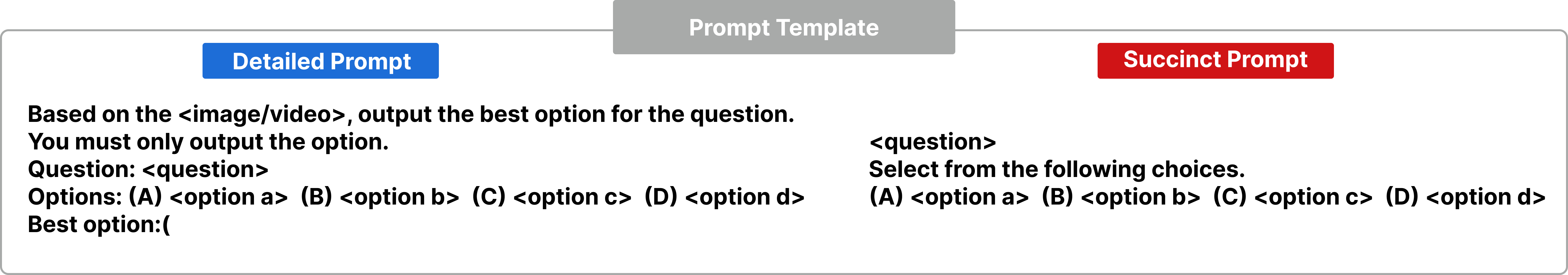}
  \caption{We adopt two distinct prompts, the detailed prompt and the succinct prompt, in our evaluation to assess models' sensitivity to different prompts.} 
  \label{fig:prompt}
\end{figure}

\paragraph{Model evaluation protocol.}
We adopt the accuracy of the model on a task to capture the model's performance. However, one task can contain numerous concrete task instances. In practice, we randomly generate $n$ task instances for a task and then use the model's accuracy on the $n$ task instances as a proxy of the model's accuracy on the task.
For prompts used to evaluate the models, to fairly evaluate the model's performance and enhance the robustness of the results, we use two versions of prompts: a succinct prompt and a detailed prompt. The succinct version simply adds 'Select from the following choices' between the question and the options~\cite{fu2024blink}, while the detailed prompt includes more instructions such as: 'Based on the image/video", and also enclose the options within parentheses (e.g., “(A) camera (B) telephone”)' and ends the prompt with 'Best Option: (' to guide the model to output the option only~\cite{li2023mvbench}. The exact prompt template can be found in Figure~\ref{fig:prompt}."
For option extraction, we match the model output to three types of option representations: 1) option identifier, \eg, “(A)”, 2) option name, \eg, “camera”, and 3) option identifier and name, \eg, “(A) camera” in order to increase the recall of the option extraction.

\paragraph{\randomname: A random set of tasks.}
To offer an overview of the task space of the current \name, we create a random subset of $100$ tasks from each task generator.
For each task, we randomly generate 3 task instances, resulting in $5,700$ ImageQA task instances and $2,700$ VideoQA task instances.
We refer to this random set as \randomname, which we release as a benchmark. 
We evaluate 18 open-source/proprietary MLMs on this set using both detailed prompt and succinct prompt.

\paragraph{\dbname: A database of model evaluation results.}
We also randomly select over 100K tasks across all the task generators and generate 15 task instances for each task, leading to over 1M task instances in total.
Then we evaluate 13 open-source MLM models on the generated task instances using detailed prompt, leading to a total number of 24,240,780 <model, task instance> evaluation pairs.
We refer to this set of evaluation results as \dbname, which we use to study the query results approximation methods and release for future study of model performance prediction. 

\paragraph{\uiname: A graphical interface for model performance investigation.}
\name allows users to query for tasks that most resemble their application.
As such, \name doesn't have to be limited to a static leaderboard commonly seen with most other benchmarks. Instead, we make \name's findings accessible through an interactive graphical user interface.
Our interface allows users to specify their needs without writing any code. They can select parts of the taxonomy that best represent their application.
We use the evaluation results in \dbname obtained our explorations to build a simple example interface: \uiname\footnote{\url{https://huggingface.co/spaces/zixianma/TaskMeAnything-UI}}.
It consists of four tabs: the \textbf{overall} tab reports model performance across a dozen MLM across different subsets of \name's taxonomy; the \textbf{task embedding} tab visualizes different task instances in a 2D space and allows users to observe model behavior across similar tasks; the \textbf{surprisingness} tab highlights tasks where a model achieves surprisingly better or worse performance compared to similar tasks; and the \textbf{query} interface supports users to conduct query-centric investigation of models' capabilities or limitations using the four types of fine-grained user queries mentioned above (Figure \ref{fig:interface}). More details can be found in the Appendix~\ref{app:interface}.

\begin{figure}[!h]
  \centering
  \includegraphics[width=\linewidth]{imgs/interface/interface.pdf}
  \caption{\uiname Interface.}
  \label{fig:interface}
\end{figure}

\section{Validating and ablating \name}
We validate the accuracy of our generated evaluated data by measuring human performance on our tasks. 
Then, we evaluate the different approximation methods introduced in Section~\ref{sec:query} to demonstrate their effectiveness.

\vspace{-6pt}

\paragraph{Validating with human evaluation.}
To validate \name, we first conduct a ($N=2$) human evaluation on \randomname to check the correctness of the tasks. In these random subsets, annotators achieve an accuracy of $92\%-100\%$ for task instances from different task generators (specifically, humans perform 100\% on the ImageQA 2D how-many tasks while 92\% on the VideoQA 3D what-rotate tasks), indicating that our tasks are accurate and can be solved by humans. By contrast, GQA~\cite{hudson2019gqa} and AGQA~\cite{GrundeMcLaughlin2021AGQA} report a human performance between $70\%-84\%$.

\vspace{-6pt}

\paragraph{Ablating the approximation algorithms.}
We evaluate the proposed query results approximation algorithms on 1,137 queries across the 4 query types (Table~\ref{tab:query-approx}). To measure the quality of the approximation, we use the evaluation results from \dbname as ground truth query results. 
From Table~\ref{tab:query-approx}, we can see that the \textit{Active} method outperforms both the \textit{Random} and \textit{Fitting} methods across nearly all query types, yet there is still room for future improvement. More details of experiments and results are in Appendix~\ref{app:query-approximation}.

\begin{table*}[h]
  \centering
  \small
  \caption{The performance of query results approximation algorithms. Top-K query uses Mean Rank (MR, lower is better) and Hit Rate (HR, higher is better) as metrics, while other queries use Precision (P), Recall (R), and F1-score (F1).}
  \scalebox{0.75}{
    \begin{tabular}{l| ccc|ccc|ccc|ccc} 
    \toprule
     \multirow{2}{*}{\bf Method} &  \multicolumn{3}{c|}{\bf Top-K Query} & \multicolumn{3}{c|}{\bf Threshold Query} & \multicolumn{3}{c|}{\bf Model Compare Query} & \multicolumn{3}{c}{\bf Model Debug Query}\\ \cmidrule(lr){2-4}\cmidrule(lr){5-7}\cmidrule(lr){8-10}\cmidrule(lr){11-13}
     
     & {\bf MR} & {\bf HR (\%)} & & {\bf P (\%)} & {\bf R (\%)} & {\bf F1 (\%)} & {\bf P (\%)} & {\bf R (\%)} & {\bf F1 (\%)} & {\bf P (\%)} & {\bf R (\%)} & {\bf F1 (\%)} \\\midrule

{\it Random} & 46.81 & 42.30 &  & 46.88 & 42.48 & 44.05 & \textbf{100.00} & 24.58 & 37.28 & \textbf{93.39} & 23.27 & 35.04\\ \midrule

{\it Fitting} & 34.43 & 46.77 &  & \textbf{47.45} & 46.34 & 46.46 & 78.42 & 47.44 & 52.59 & 83.27 & 32.04 & 43.86\\ \midrule

{\it Active} & \textbf{10.79} & \textbf{70.55} &  & 47.39 & \textbf{46.83} & \textbf{46.55} & 89.94 & \textbf{54.88} & \textbf{61.87} & 89.95 & \textbf{43.84} & \textbf{56.44}\\ 

\bottomrule 
\end{tabular}
}
  
  \label{tab:query-approx}
\end{table*}

\begin{figure}[!h]
  \centering
  \begin{minipage}{0.52\linewidth}
    \centering
    \includegraphics[width=\linewidth]{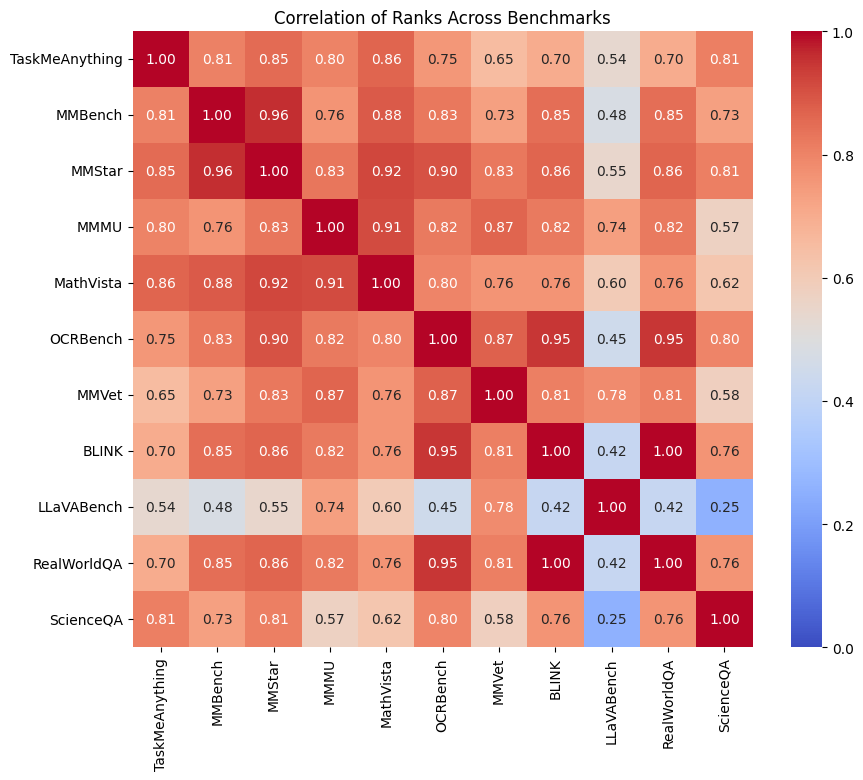}
    \caption{Correlations of \name with other popular MLM benchmarks.}
    \label{fig:cor-with-other-benchmarks}
  \end{minipage}
  \hfill
  \begin{minipage}{0.38\linewidth}
    \centering
    \includegraphics[width=\linewidth]{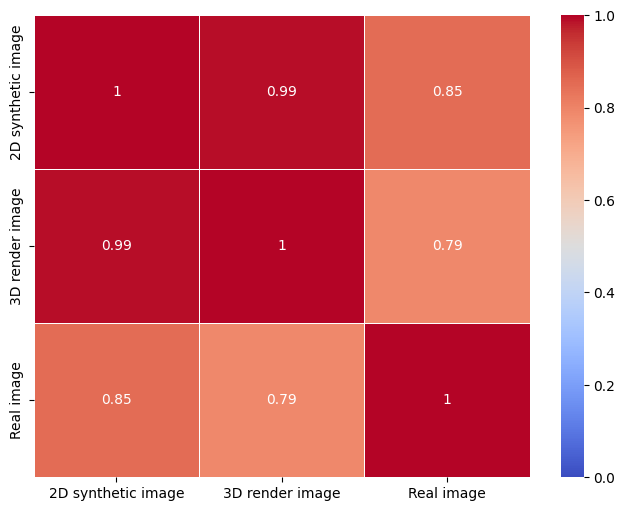}
    \caption{Correlations of synthetic data with real data in \name.}
    \label{fig:cor-within-tma}
  \end{minipage}%
\end{figure}

\paragraph{Comparison with existing benchmarks.}
According to Figure~\ref{fig:cor-with-other-benchmarks}, we found that there is a strong positive correlation between models’ rankings on \name and on other benchmarks, with strong correlations ($\geq$0.8) on the most commonly used ones such as MMMU and MMBench. Notably, the average correlation between \name and the other benchmarks is 0.77, which is greater than that of some other benchmarks such as ScienceQA and LLaVABench (0.70 and 0.57). These results suggest that the evaluation results on our benchmark align with those on existing benchmarks.

\paragraph{Analysis on synthetic data in \name.}
\name leverages both synthetic and real images/videos for evaluating MLM models. While synthetic data are controllable, low-cost, its effectiveness compared to real data remains unclear, particularly given that most benchmarks rely heavily on real datasets.
To quantitatively justify the transferability between evaluations on synthetic and real-world images, we compared 18 models’ rankings on synthetic vs. realistic images and obsered strong positive correlations between them. Across the three different image sources in Task-Me-Anything, we found that the correlation between models’ rankings on 2D and 3D synthetic images is the strongest (r=0.99). While the correlations between models’ rankings on real and 2D images, and between real and 3D images are slightly smaller (r=0.85 and 0.79 respectively), these numbers still suggest strong positive correlations between models’ performance on synthetic images and on real images (Figure~\ref{fig:cor-within-tma}).

\section{Analysing MLMs with \name}
\label{sec:ana}
We use \name to conduct multiple analyses to highlight its different use cases, while simultaneously drawing insights about today's MLMs
(More details in Appendix~\ref{app:analysis}).
Specifically, we evaluated 18 MLMs on \randomname for Query 1 and 4 and reused the evaluation results of \dbname for Query 2, 3, 5, and 6.
Finally, we leverage \name to provide an in-depth analysis on \gptfouro as Query 7.

\subsection{Query 1: How do models perform over a random subset of all possible questions?}
We evaluated 18 MLMs on the \randomname test set (Figure~\ref{fig:random-eval}) to gain an overview of model performance. (Details in Appendix~\ref{app:random-result}).

For ImageQA tasks, the latest open-sourced models, such as \internvlchat and \llavanext, surprisingly outperform popular proprietary models, achieving state-of-the-art performance as shown in recent benchmarking results~\cite{2023opencompass}.

For VideoQA tasks, in addition to natural VideoQA models, we evaluated larger or proprietary ImageQA models, such as \gptfourv, by combining four frames of a video into a single picture. We observed that natural VideoQA models, such as \videollavas, still have a significant performance gap compared to large ImageQA models, even though ImageQA models are not trained on video and only use a combined image rather than the full video.

\begin{figure}[!h]
  \centering
  \includegraphics[width=\linewidth]{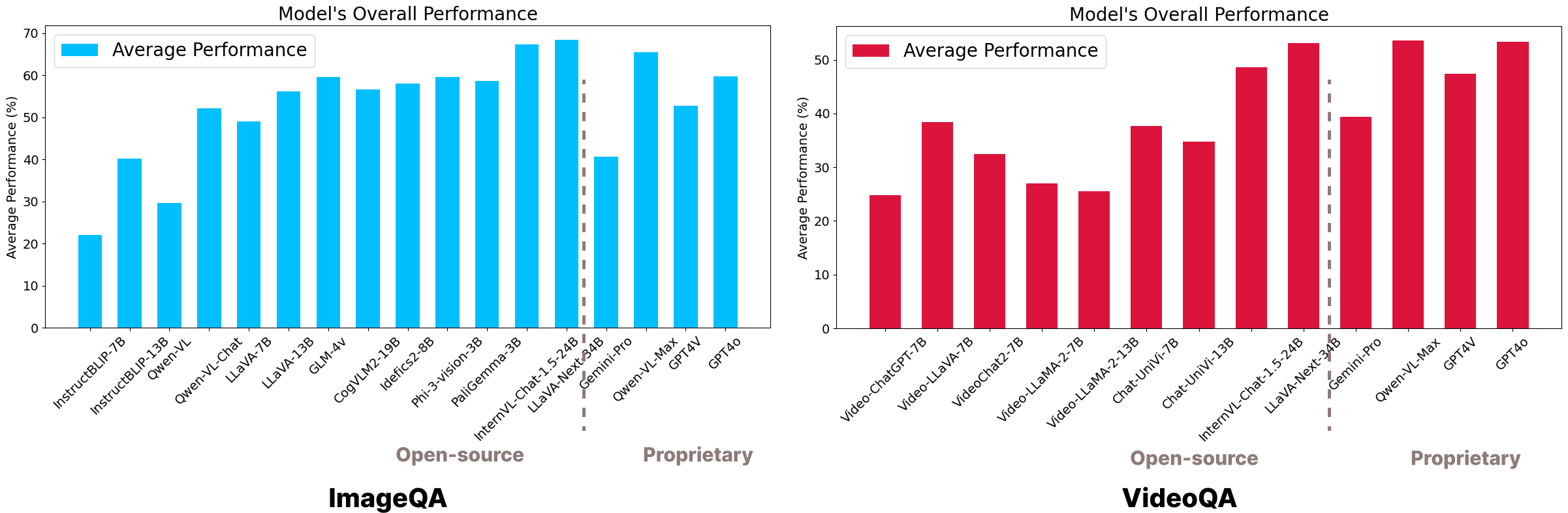}
  \caption{Model performance on the \randomname, a random subset of tasks from \name.}
  \label{fig:random-eval}
\vspace{-5mm}
\end{figure}

\subsection{Query 2: Are current models sensitive to prompts?}

\begin{figure}[!h]
  \centering
  \includegraphics[width=\linewidth]{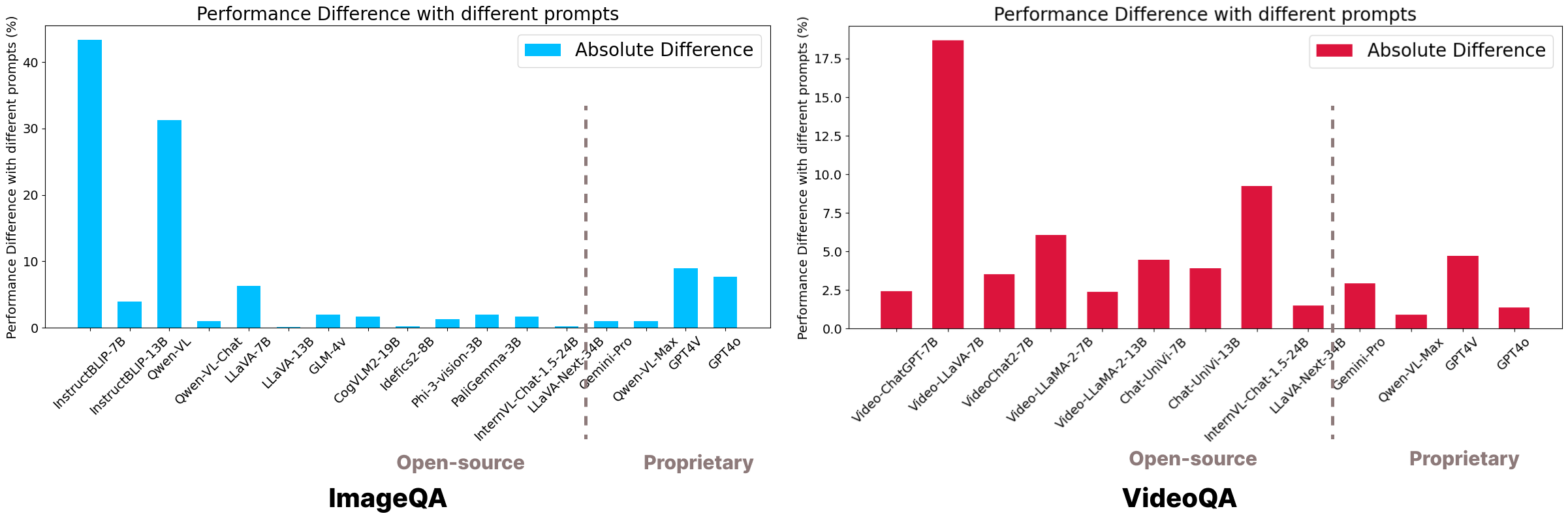}
  \caption{The absolute difference of model performance in different prompts in \randomname.}
  \label{fig:random-prompt-difference}
\vspace{-5mm}
\end{figure}

Our evaluation indicates that current models are still prompt-sensitive (Figure \ref{fig:random-prompt-difference}). We tested our model on two versions of prompts: a succinct version that only describes the question and a detailed version that includes more instructions.

First, we observed that MLMs are still highly prompt-sensitive, and using different prompts might result in over 40\% performance differences. 

While detailed prompts generally lead to better results, certain models, such as \gptfourv, perform notably better with more succinct prompts. In contrast, models like \instructblips and \qwenvl exhibit a marked improvement with detailed prompts compared to succinct ones. (Details in Appendix~\ref{app:random-result}).

\subsection{Query 3: What skills are MLMs best and worst at?}
We analyze performance across different perceptual capabilities to answer: what skills are all models good or bad at? We conduct this study for both ImageQA and VideoQA tasks respectively. 
We find that no specific skill appears to be the best or worst across (both image and video) models (Figure \ref{fig:best-worst-skills}). 
We see that all models struggle in spatial reasoning, counting objects, and 3D attribute understanding on ImageQA tasks, and object recognition, temporal understanding on VideoQA tasks. They perform well on object, attribute, and other relationship recognition instances.
Surprisingly, we find that most MLMs perform the best at relationship understanding between objects, scoring high if not perfectly on interactional relations such as ``riding'', ``looking into'', ``lying next to'' etc. On the other hand, these models struggle the most in spatial reasoning in synthetic images, performing poorly especially on questions that ask about objects in the ``middle'', ``bottom'' or ``back'' (for 3D images) part of the image. Nevertheless, some models behave differently. For example, \llaval is worst at recognizing 3D attributes, failing at identifying the ``smallest'' or ``closest'' 3D objects correctly. Meanwhile, \llavas is best at object recognition and worst at relation understanding, struggling to understand simple actions such as ``touching'' that other models perform well on.

\begin{figure*}[ht!]
\normalsize
\centering
\begin{minipage}{0.32\linewidth}
\centering
  \includegraphics[width=\linewidth]{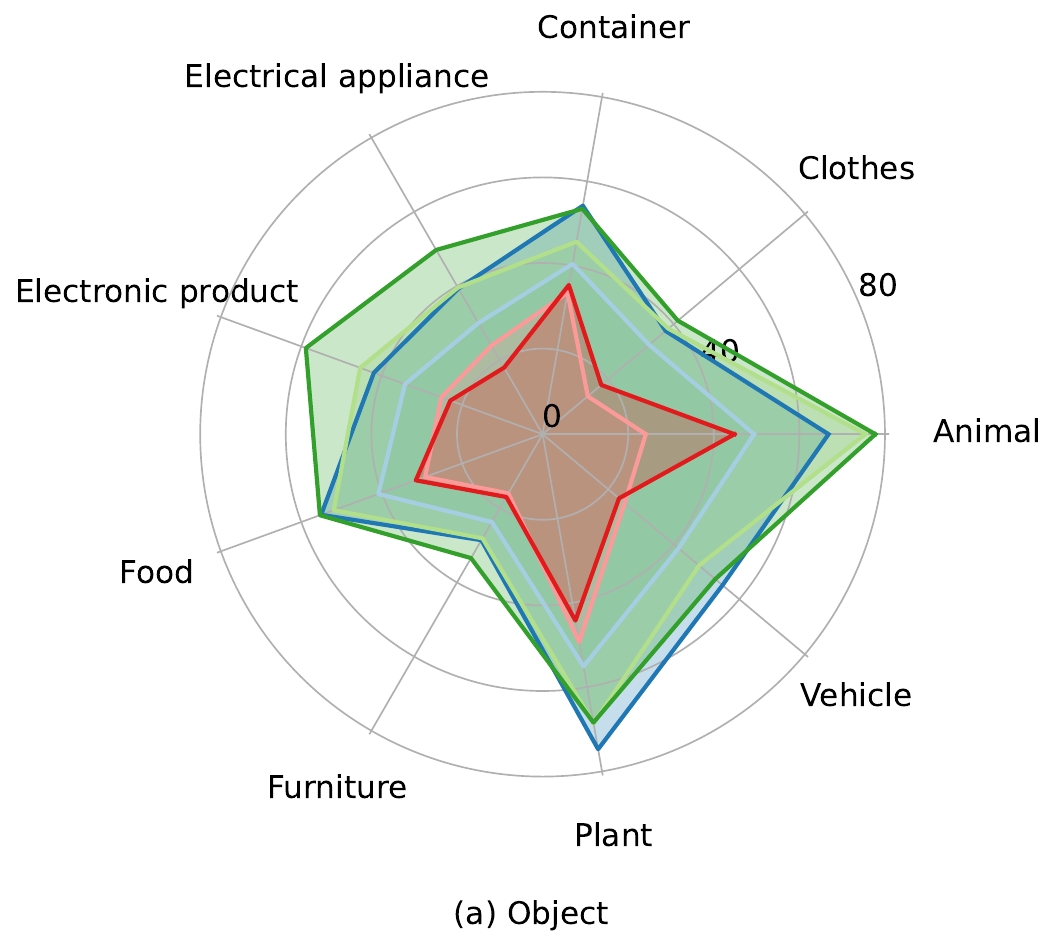}
\end{minipage}\hfill
\begin{minipage}{0.32\linewidth}
\centering
 \includegraphics[width=\linewidth]{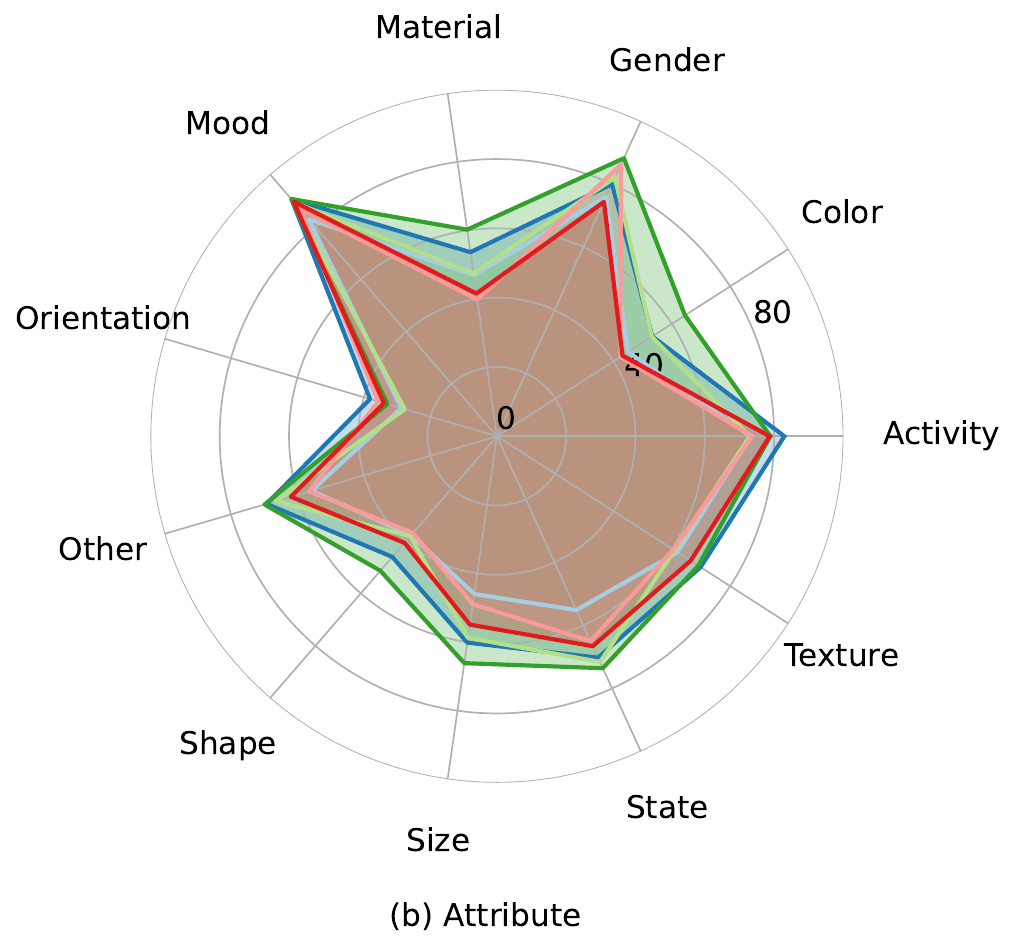}
\end{minipage}
\begin{minipage}{0.35\linewidth}
\vspace{3mm}
\centering
  \includegraphics[width=\linewidth]{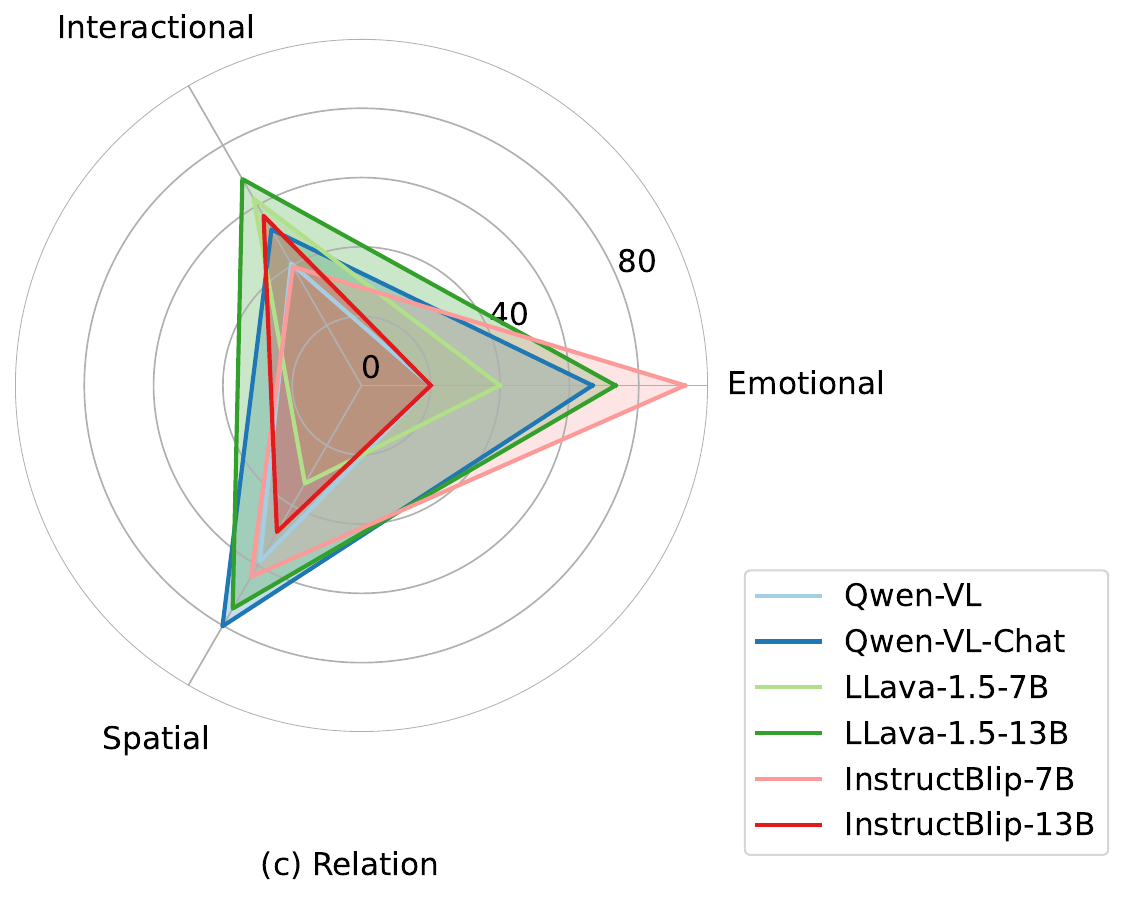}
\end{minipage}\hfill
\caption{\textbf{ImageQA, fine-grained skills, all models.} We also analyze models' performance on ImageQA tasks across fine-grained skills and find that models are good at recognizing plants, understanding mood, and comprehending spatial relations between real-world objects on average despite differences in individual models.}
\label{fig:finegrained-skills}
\end{figure*}

\begin{figure*}[ht!]
\normalsize
\centering
\begin{minipage}{0.47\linewidth}
\centering
  \includegraphics[width=\linewidth]{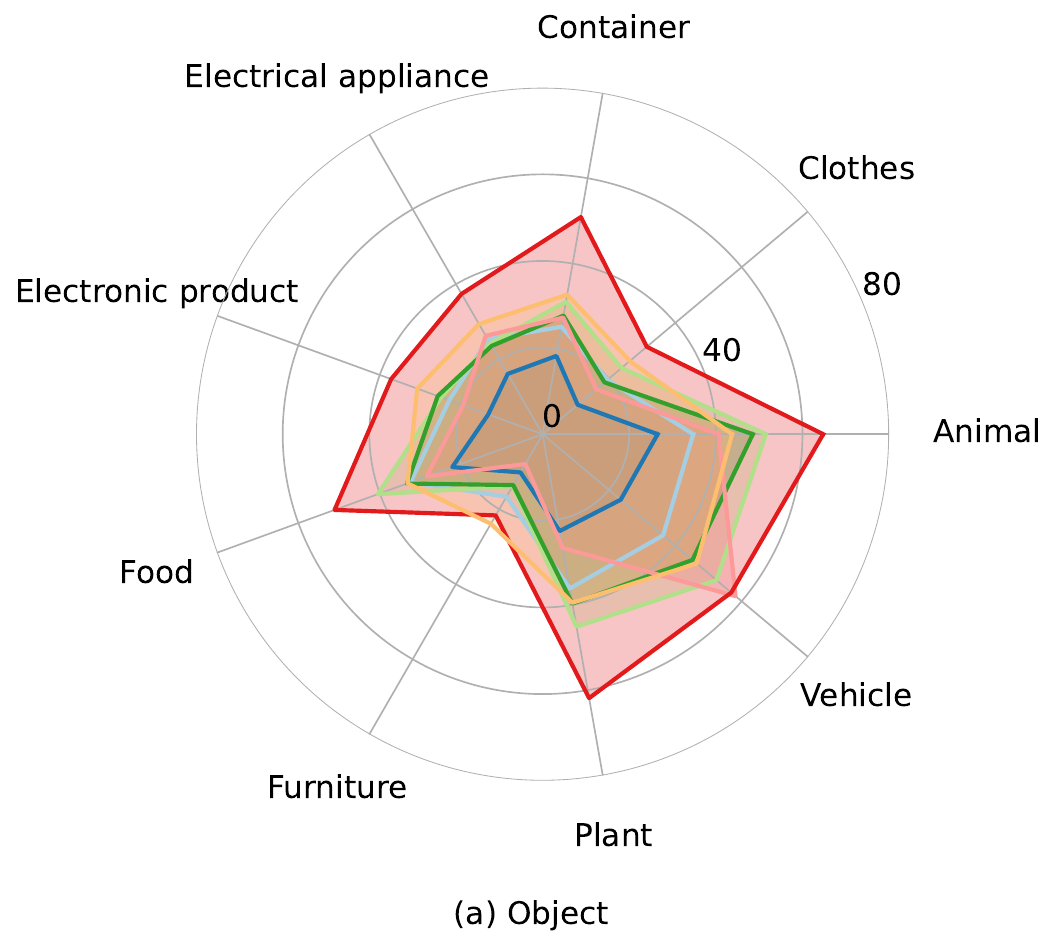}
\end{minipage}\hfill
\begin{minipage}{0.51\linewidth}
\vspace{3mm}
\centering
 \includegraphics[width=\linewidth]{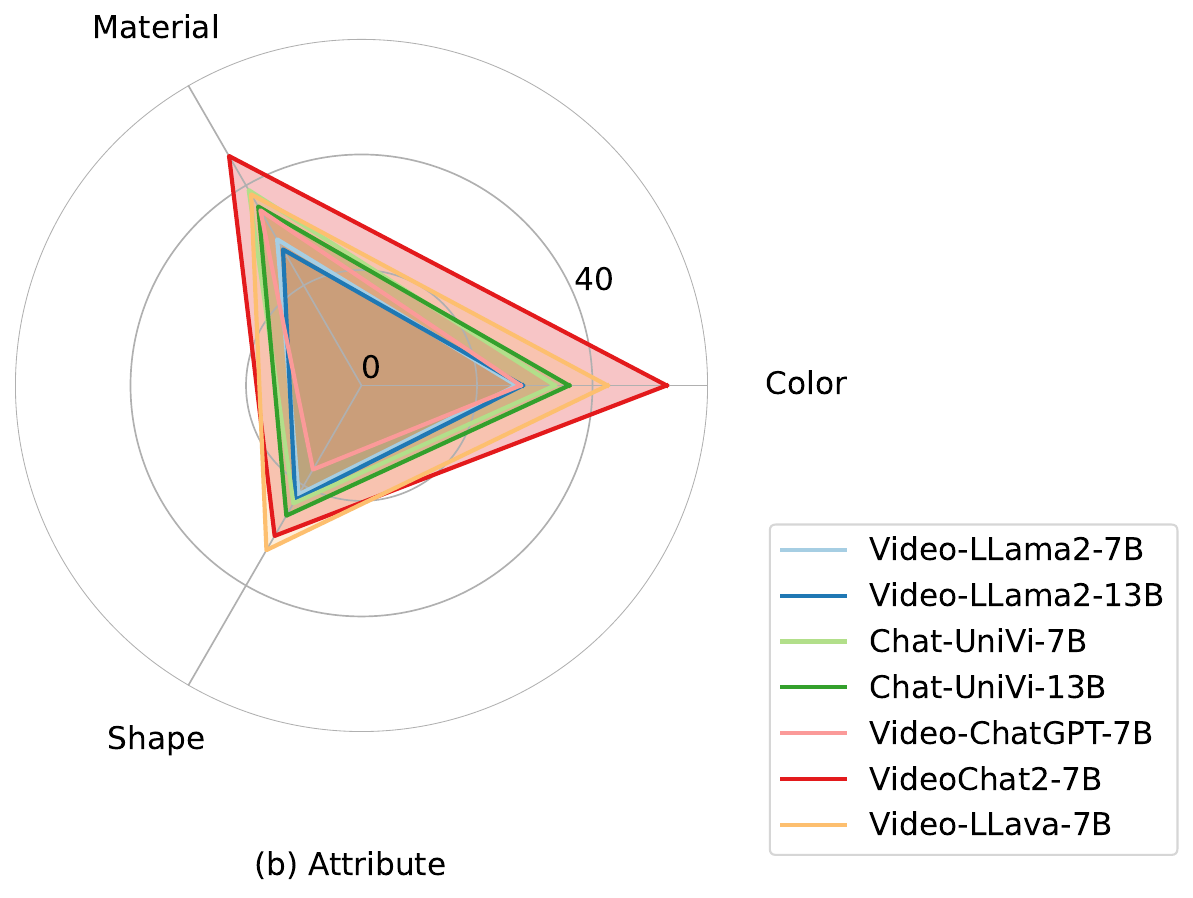}
\end{minipage}
\caption{\textbf{VideoQA, fine-grained object and attribute skills, all models.} We present models' performance on VideoQA tasks across fine-grained skills and find that, on average, models are good at recognizing vehicles and understanding materials in videos.}
\label{fig:finegrained-skills-video}
\end{figure*}

\begin{figure*}[ht!]
\normalsize
\centering
\begin{minipage}{0.5\linewidth}
\centering
  \includegraphics[width=\linewidth]{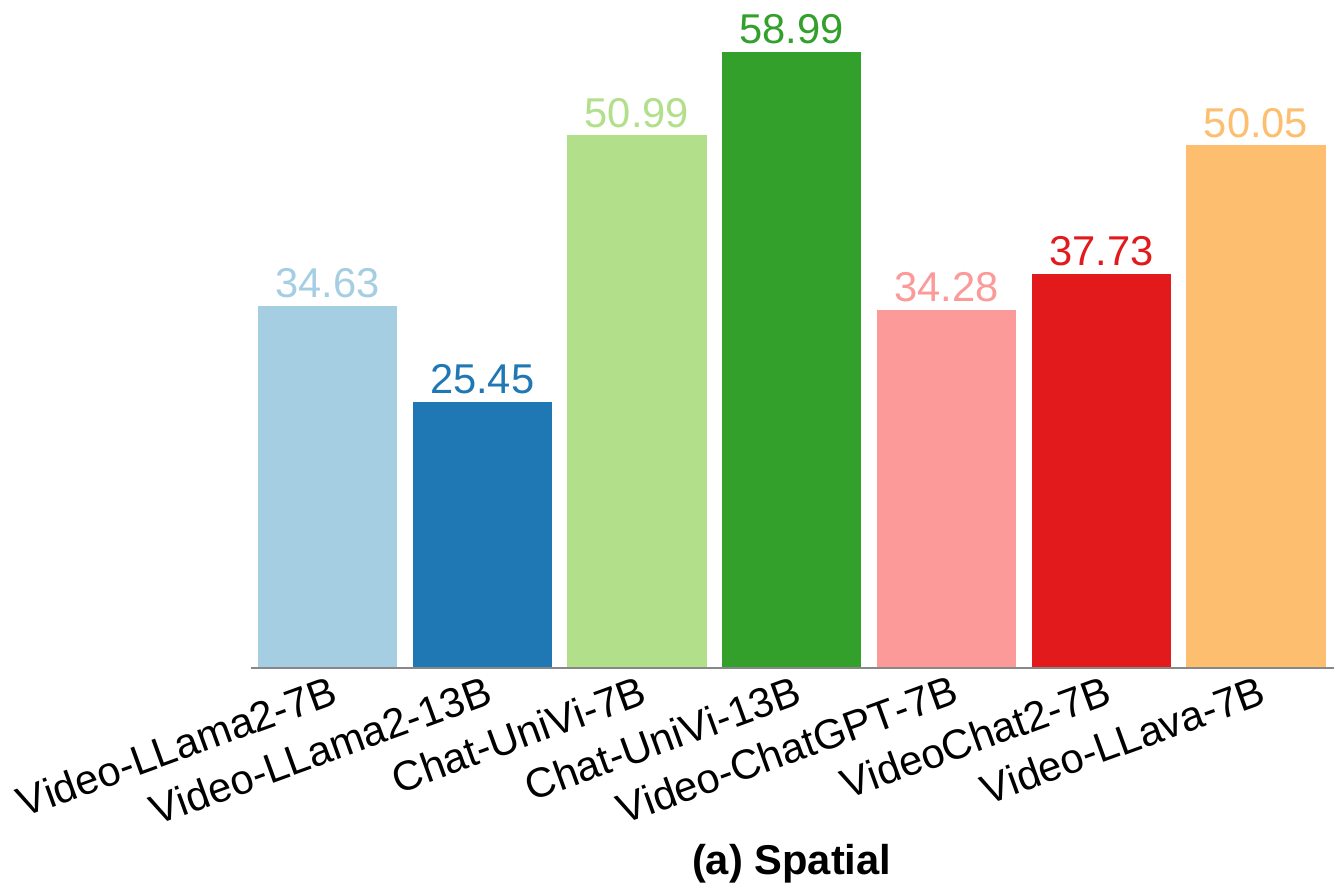}
\end{minipage}\hfill
\begin{minipage}{0.5\linewidth}
\vspace{3mm}
\centering
 \includegraphics[width=\linewidth]{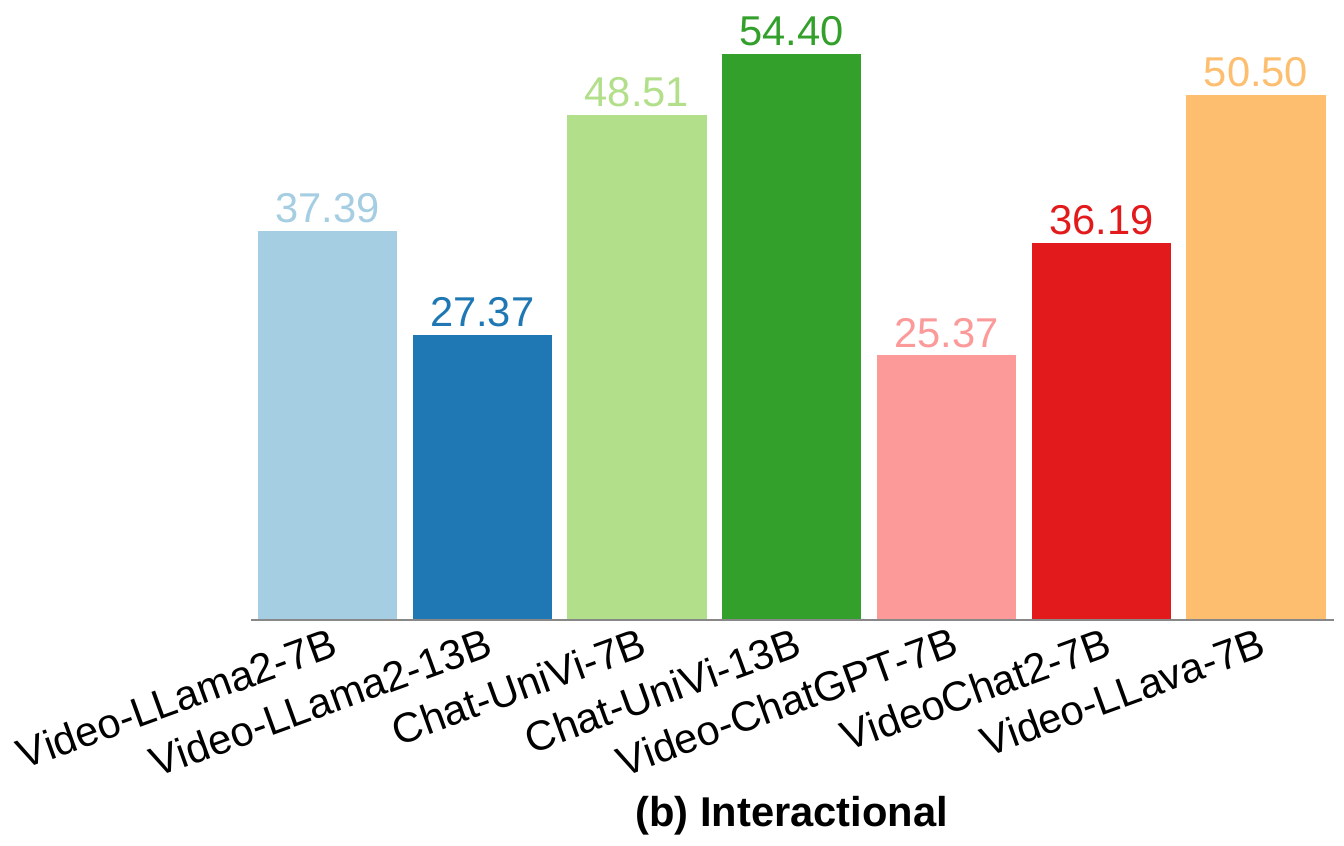}
\end{minipage}
\caption{\textbf{VideoQA, fine-grained relation skills, all models.} On VideoQA tasks, we find that models are better at understanding spatial relations than interactional ones on average.}
\label{fig:finegrained-relation-video}
\end{figure*}

Further, \name also enables us to conduct analyses of models' fine-grained skills such as recognizing a specific type of object, attribute, or relation. For example, on ImageQA tasks, we find that on average models are better at recognizing plants, understanding mood and comprehending spatial relations between real-world objects (Figure \ref{fig:finegrained-skills}). Nevertheless, some models might showcase different strengths: \llaval is better at recognizing animals (Figure \ref{fig:finegrained-skills} (a)), and \instructblips is better at understanding emotional relationships (Figure \ref{fig:finegrained-skills} (c)). On the other hand, for VideoQA tasks, we learn that models are better at recognizing vehicles, material and understanding spatial relationships (Figures \ref{fig:finegrained-skills-video} and \ref{fig:finegrained-relation-video}).

\subsection{Query 4: what is the best MLM for each specific skill?}
\begin{figure*}[ht!]
\normalsize
\centering
    \begin{minipage}{0.49\linewidth}
    \centering
  \includegraphics[width=\linewidth]{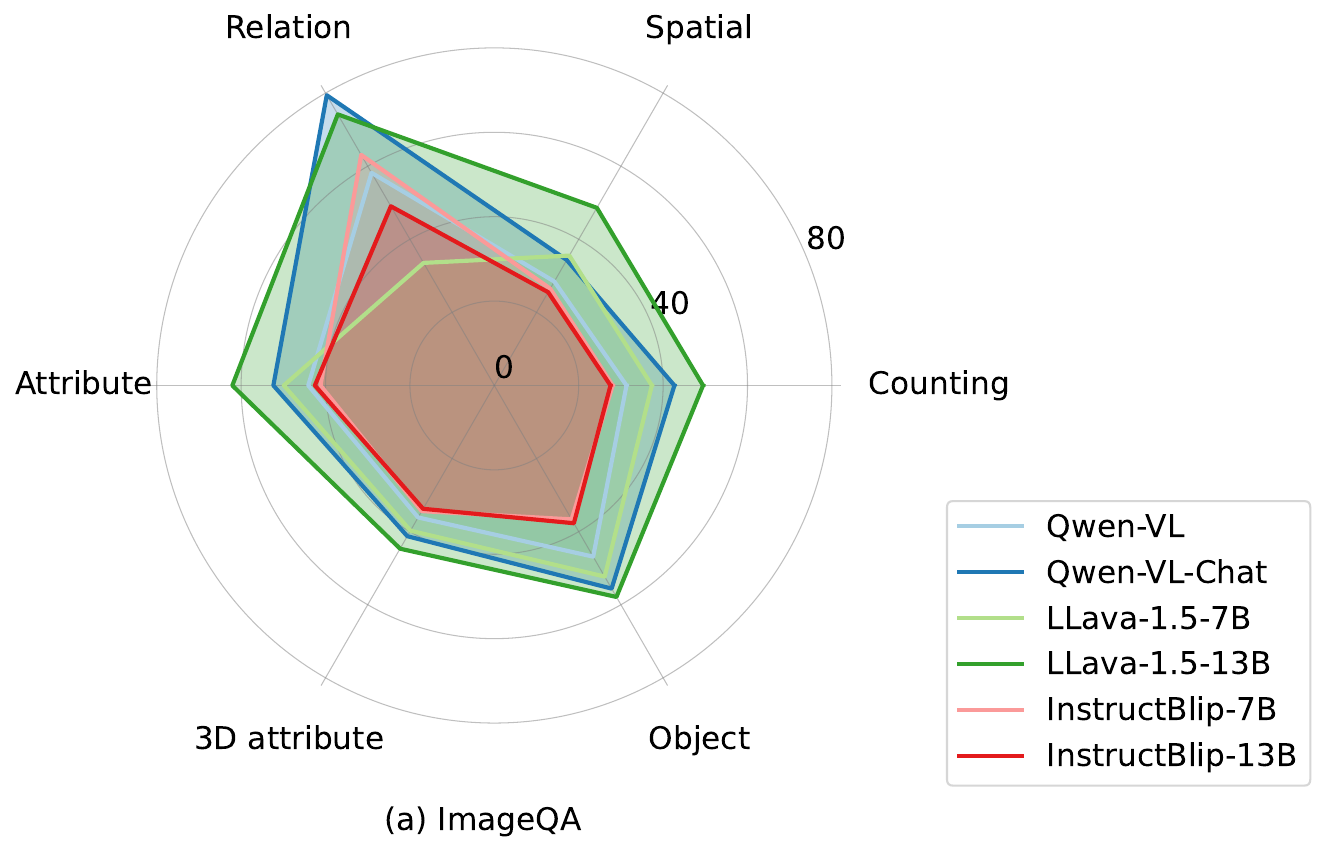}
    \end{minipage}
    \begin{minipage}{0.49\linewidth}
\centering
  \includegraphics[width=\linewidth]{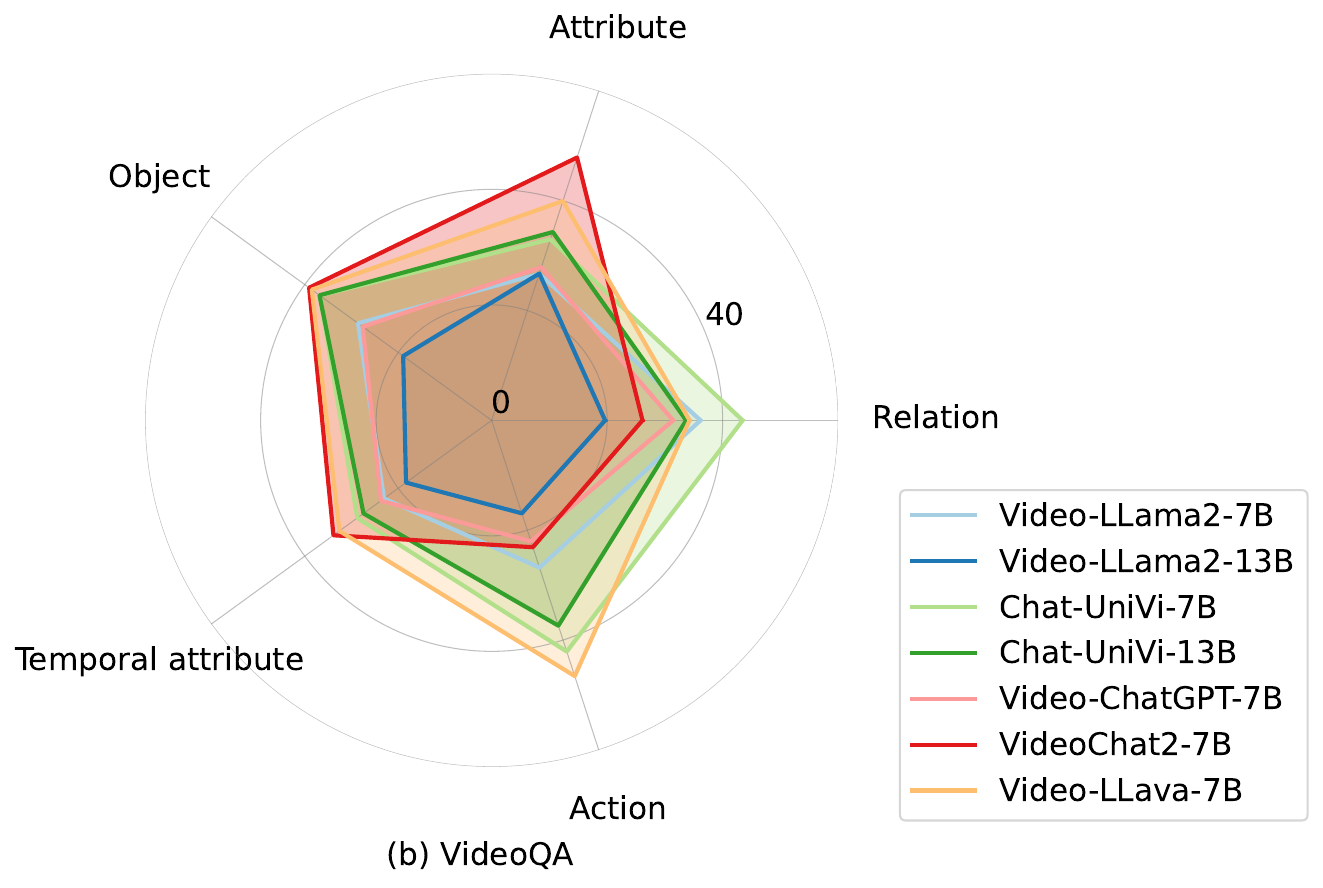}
\end{minipage}\hfill
   \caption{\textbf{Image and VideoQA, high-level skills, all models.} We plot models' performance on Image and VideoQA tasks across all skills. We learn that models are relatively good at object and attribute recognition in both Image and VideoQA and relation understanding in ImageQA but still struggle at others.}
    \label{fig:best-worst-skills}
\end{figure*}

\llaval stood out as the strongest model on ImageQA tasks, achieving the best performance on all skills except for relation understanding; and \videollavas is the overall winner on VideoQA tasks, scoring the highest on action understanding and second or third elsewhere. Specifically, we find that \llaval performs consistently better than other multi-modal models on all skills except for relation understanding, where \qwenvlchat performs better (Figure \ref{fig:best-worst-skills} (a)). On VideoQA tasks, in addition to \videollavas, \chatunivis is also relatively well-rounded, positioning in the top 3 models across all skills except for Attribute understanding (Figure \ref{fig:best-worst-skills} (b)). On the other hand, while \videochattwos specializes in object, attribute, and temporal attribute understanding, it falls short on Action and Relation reasoning (Figure \ref{fig:best-worst-skills} (b)). 





\begin{figure*}[ht!]
\normalsize
\centering
    \begin{minipage}{0.47\linewidth}
    \centering
  \includegraphics[width=\linewidth]{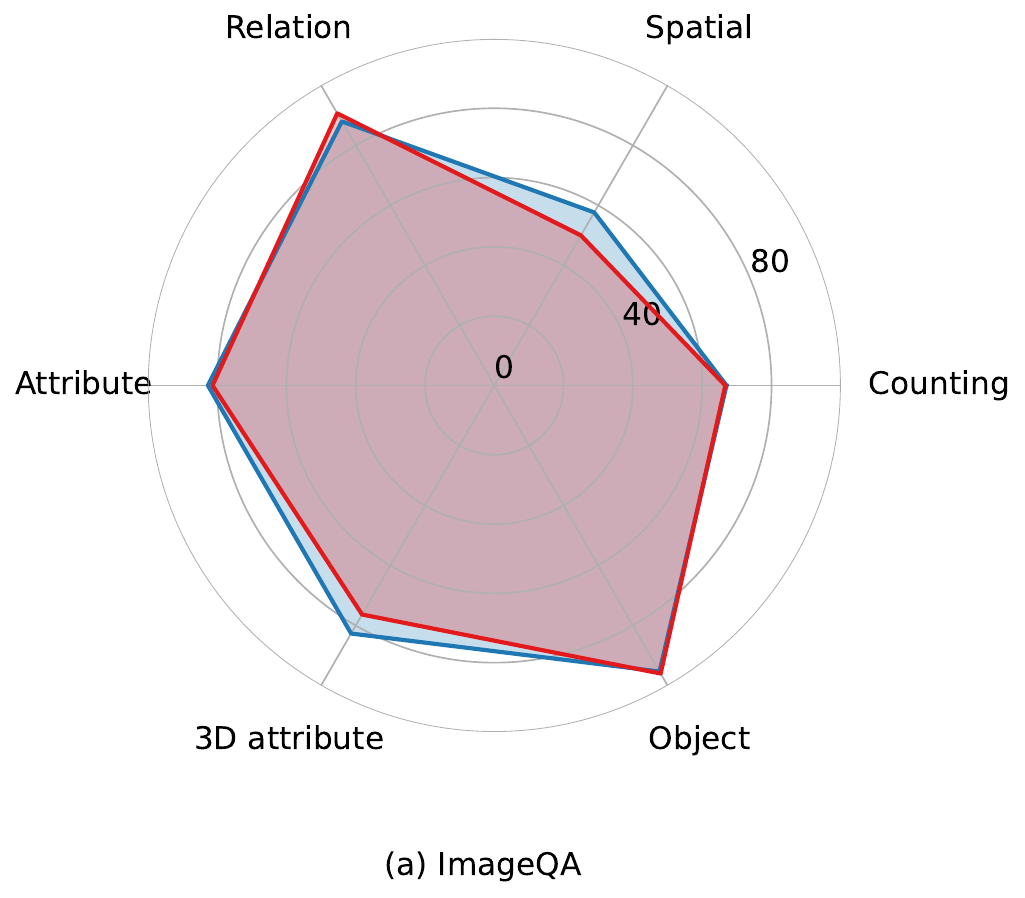}
    \end{minipage}
    \begin{minipage}{0.51\linewidth}
\centering
  \includegraphics[width=\linewidth]{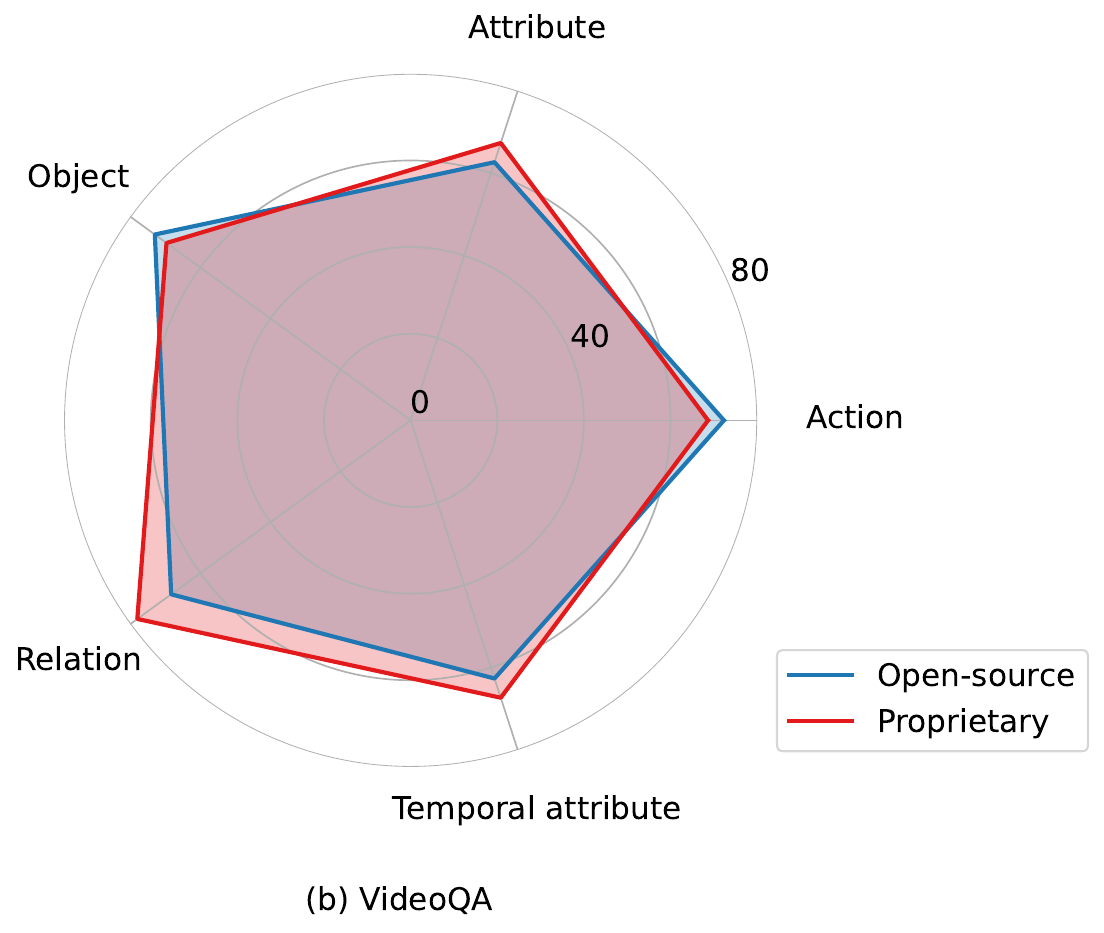}
\end{minipage}\hfill
\caption{\textbf{Image and VideoQA, high-level skills, open-source vs. proprietary best models.} We plot the performance of the best open-source and proprietary model for each skill on ImageQA and VideoQA tasks.}
  \label{fig:open-vs-close}
\end{figure*}

\subsection{Query 5: How does the best open-source model compare against the best proprietary model across skills?}
Moreover, we find that on ImageQA tasks, the best open-source model (\llavanext on object recognition, \llaval on relation understanding and \internvlchat else where) is on par with if not better than the best proprietary model (\gptfouro on attribute recognition, \gptfourv on counting and \qwenvlchat else where) for most skills (Figure \ref{fig:open-vs-close}). Notably, the best open-source model outperforms the best proprietary one on spatial reasoning by around 8\% and 3D attribute by 7\%. On VideoQA tasks, the best open-source model \internvlchat surpasses the best proprietary one \qwenvlmax on object and action recognition but lags behind proprietary models by 5-10\% on attribute, temporal attribute and relation understanding.

\subsection{Query 6: How do small models compare against large models?}
\begin{figure}[!h]
  \centering
  \includegraphics[width=\linewidth]{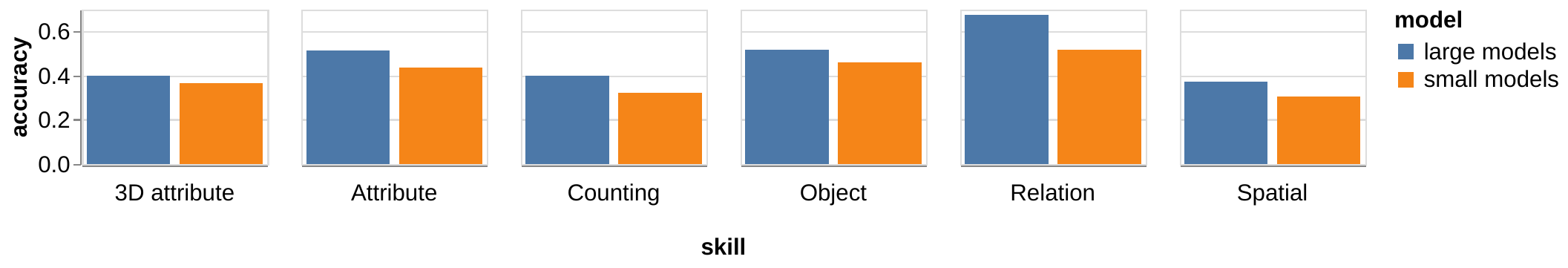}
  \caption{Skill comparison: small vs. large models on ImageQA.}
  \label{fig:image-small-vs-large}
\end{figure}

\begin{figure}[!h]
  \centering
  \includegraphics[width=\linewidth]{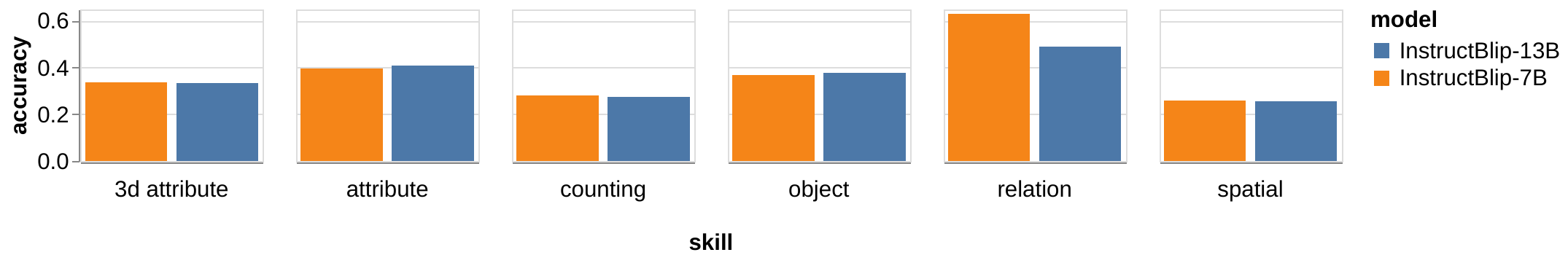}
  \caption{Skill comparison: \instructblips vs. \instructblipl.}
  \label{fig:instructblip-small-vs-large}
\end{figure}

We are also interested in the relative performance of small versus large models with the same skills. On ImageQA tasks, for example, we observe that large multi-modal models collectively perform better than smaller models on ImageQA tasks (Figure \ref{fig:image-small-vs-large}). Nevertheless, this finding might not always hold for individual models. Through t-tests with pairs of small and large models from the same source, we find one exception: \instructblips ($\mu$ = 0.63) significantly outperforms \instructblipl  ($\mu$ = 0.49) on relation understanding with $p$-value $< 1e-5$ (Figure \ref{fig:instructblip-small-vs-large}). 

\begin{figure}[!h]
  \centering
  \includegraphics[width=\linewidth]{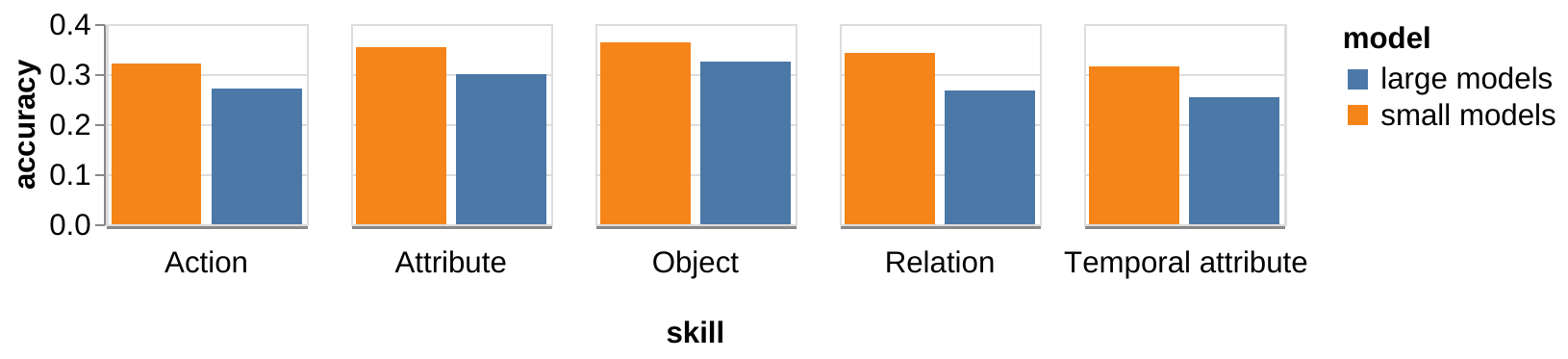}
  \caption{Skill comparison: small vs. large models on VideoQA.}
  \label{fig:video-small-vs-large}
\end{figure}

\begin{figure}[!h]
  \centering
  \includegraphics[width=\linewidth]{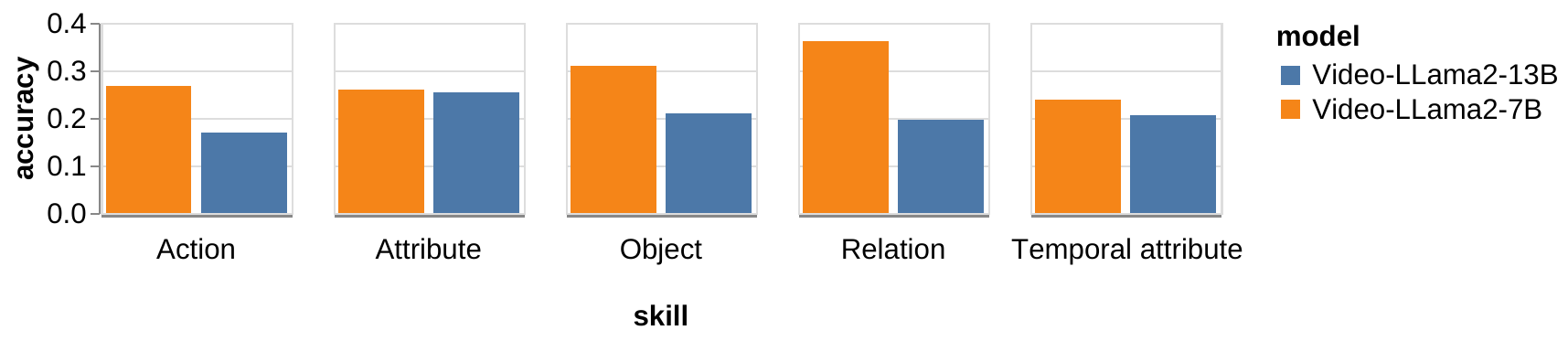}
  \caption{Skill comparison: \videollamatwos vs. \videollamatwol.}
  \label{fig:videollama-small-vs-large}
\end{figure}

\begin{figure}[!h]
 \centering
  \includegraphics[width=\linewidth]{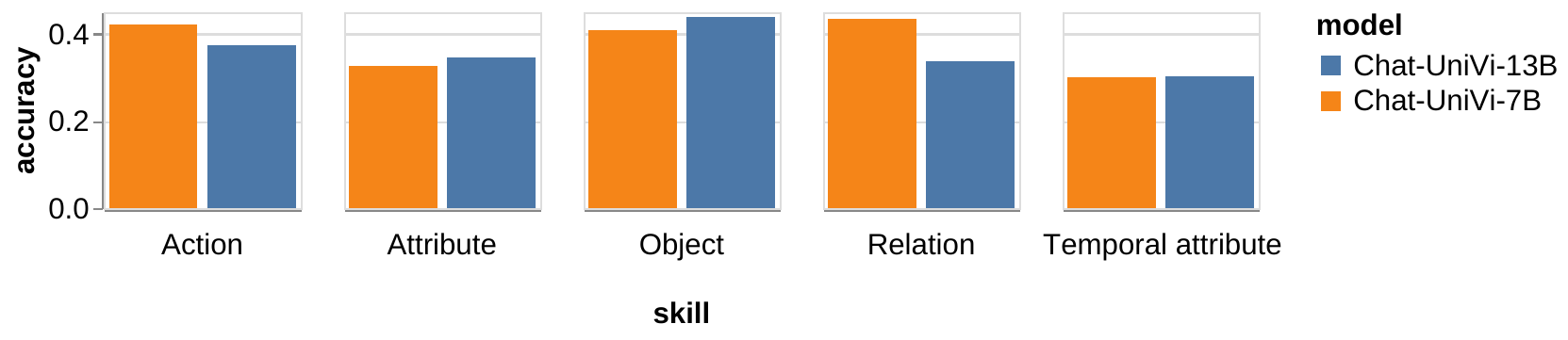}
  \caption{Skill comparison: \chatunivis vs. \chatunivil.}
  \label{fig:video-chat-small-vs-large}
\end{figure}

On VideoQA tasks, interestingly, we find that small models beat larger models on VideoQA tasks on average (Figure \ref{fig:video-small-vs-large}). We hypothesize that this is because we included some strong small video models in our evaluation. For example, we see that \videollamatwos achieves a higher score than \videollamatwol in all skills with $p$-value $< 3e-5$ (Figure \ref{fig:videollama-small-vs-large}), and \chatunivis outperforms \chatunivil on action and relation understanding with $p$-value $< 1e-5$ (Figure \ref{fig:video-chat-small-vs-large}).

\subsection{Query 7: Are models' strengths and weaknesses consistent across visual inputs?}
\begin{figure}[!h]
  \centering
  \includegraphics[width=\linewidth]{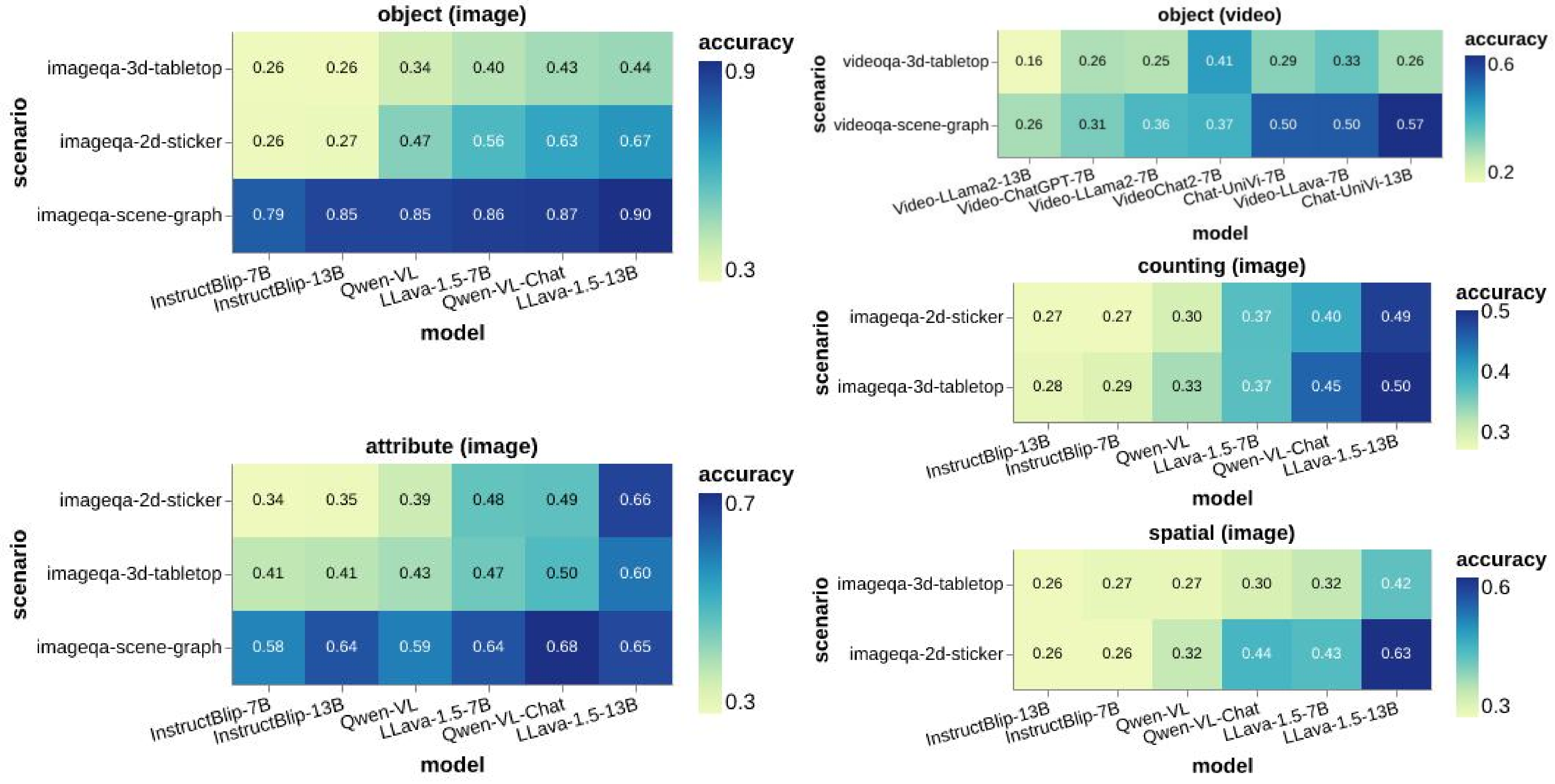}
  \caption{We present models' performance for each skill across visual inputs.}
  \label{fig:skills-across-scenarios}
\end{figure}

Further, we are curious if the models’ strong and weak skills are consistent across visual inputs. To this end, we look at models’ performance across visual inputs for object, attribute, spatial understanding, and counting as these skills involve tasks in multiple visual inputs such as 2D and 3D. We find that for the same skill, the rankings of models remain largely consistent across visual inputs (Figure \ref{fig:skills-across-scenarios}). We observe strong correlations (with Spearman coefficients of 0.77-0.94)  between models’ accuracy scores for different visual inputs in the same skill with only one exception: the video models’ performance on object understanding in 3D tabletop tasks is only weakly correlated (coefficient = 0.64) with their performance in scene graph tasks. This finding suggests our definition of skills is orthogonal to visual inputs and enables us to find models’ inherent strengths and weaknesses. 

\subsection{Query 8: What is today's popular proprietary model, \gptfouro, bad at?}
Finally, we investigate \gptfouro, today's popular proprietary model:
what \emph{objects} are \gptfouro bad at recognizing when rotating/moving?
what \emph{relations} are \gptfouro bad at understanding?
and what \emph{attributes} of objects are \gptfouro bad at recognizing?
To answer these questions, we first identify task generators for each question that can generate relevant tasks to evaluate, based on which we provide both the object/relation/attribute categories and individuals that \gptfouro are bad at.
Note that these are just example questions, and many more of this type can be addressed by \name.

\begin{figure}[!h]
  \centering
  \includegraphics[width=\linewidth]{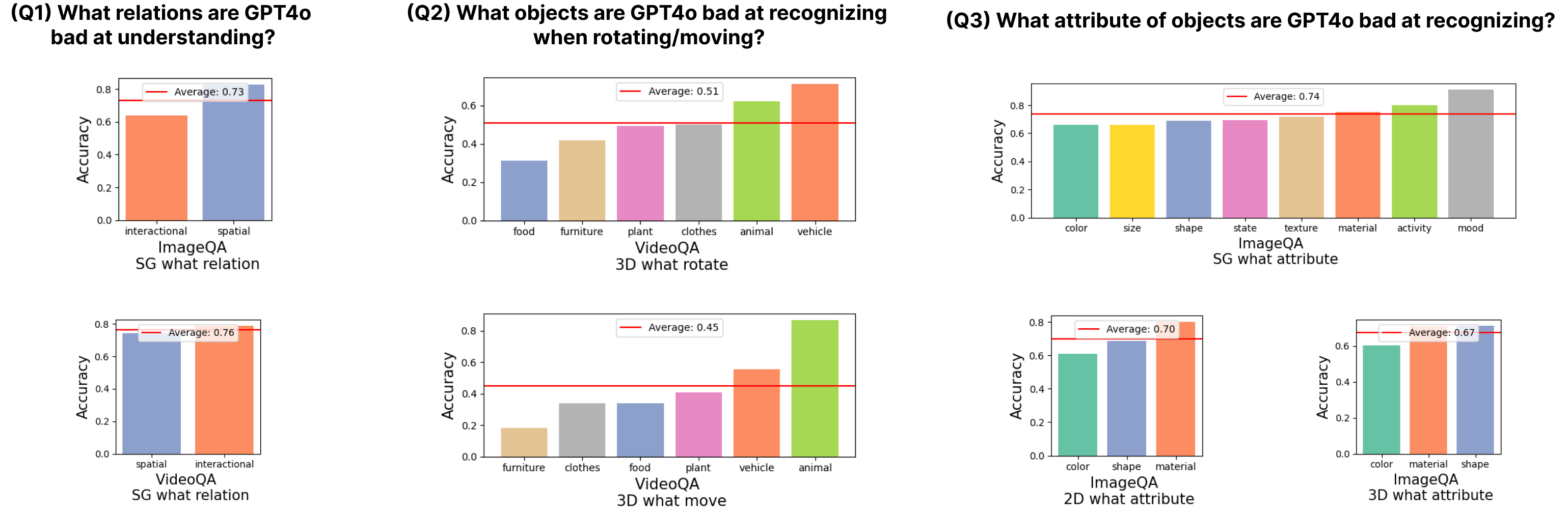}
  \caption{Answering Q1-Q3 with \gptfouro performance on randomly generated task instances relating to coarse-grained object/relation/attribute categories.}
  \label{fig:case-study-type}
\end{figure}

\noindent\textbf{Answering with object/relation/attribute categories.}
First, we answer these questions by comparing \gptfouro's performance across different coarse-grained object/relation/attribute categories and their average, as shown in Figure~\ref{fig:case-study-type}.
We can see that
1) \gptfouro does not perform well in recognizing “interactional” relations in images and “spatial” relations in videos, 
2) recognizing rotating/moving “furniture”, “food”, and “plant” is more challenging for \gptfouro than other object categories such as animal and vehicle, 
and 3) \gptfouro is worse at recognizing “color” than other attributes.

\noindent\textbf{Answering with individual objects/relations/attributes.}
To pinpoint the specific objects/relations/attributes that \gptfouro can't do well,
we convert each question to a Top-K query regarding individual objects/relations/attributes, and employ our \textit{Active} method for query results approximation with a budget of \gptfouro calls.
We found that \gptfouro's performance drops by a large margin ($-5$\% to $-50$\%) on the Top-5 objects/relations/attributes founded by \name, indicating they remain challenging for \gptfouro (Table~\ref{tab:case-study}).
This example use case of \name demonstrates how to leverage the system for locating the model weakness regarding fine-grained concepts.

\begin{table*}[!h]
  \centering
  \small
  \caption{Answering Q1-Q3 with Top-K query regarding individual objects/relations/attributes. We also present the \gptfouro performance drop ($\Delta$ Perf. (\%)) on task instances involving found task elements as ground truth answers compared to random task instances, and show that performance drops by a large margin.}
  \scalebox{1.0}{
  \resizebox{\linewidth}{!}{
    \begin{tabular}{llll} 
    \toprule
    {\bf Question} & {\bf Task generator}  & {\bf Top-K objects/relations/attributes}  & {\bf $\Delta$ Perf. (\%)}\\ 
    \midrule

    \multirow{1}{*}{what objects are \gptfouro bad at recognizing}  
    &  VideoQA 3D what rotate &  \emph{fermentation product, hamper, tool, computer keyboard, mathematical instrument} & -21.67\\
    \multirow{1}{*}{when rotating/moving?}  & VideoQA 3D what move & \emph{towel, bathtub, furniture, air conditioner, desk} & -19.33 \\
 
    \midrule

    \multirow{2}{*}{what relations are \gptfouro bad at understanding?}  
    &  ImageQA SG what relation & \emph{taller than, exiting, pushing, pushed by, between} & -51.05 \\
     & VideoQA SG what relation & \emph{beneath, covered by, carrying, above, standing on} & -16.66\\
    \midrule

    \multirow{3}{*}{what attributes are \gptfouro bad at recognizing?}  
    &  ImageQA 2D what attribute & \emph{purple, brown, red, gray, beige} & -5.33\\
    &  ImageQA 3D what attribute & \emph{stone, rubber, textile, leather, plastic} & -10.67\\
    &  ImageQA SG what attribute & \emph{crooked, power, lower, steep, glowing} & -45.45\\
\bottomrule
\end{tabular}
}
}
  \label{tab:case-study}
\end{table*}

\subsection{Which models should I use?}

When considering which models to use for your tasks, proprietary models generally offer superior performance due to their extensive training on diverse datasets and robust architectures. For tasks requiring high accuracy in object recognition, relation understanding, or attribute identification, \gptfouro is currently among the best options. Its ability to handle complex queries and produce refined outputs makes it a top choice for applications demanding reliability and depth.

For those seeking open-source alternatives, \llavanext and \internvlchat stand out as strong contenders, often achieving performance levels comparable to proprietary models while being much smaller in size (20-30B) and capable of running on a single A100 GPU or two A6000 GPUs. In VideoQA tasks, current models are still in their early stages and may underperform compared to popular ImageQA models that only takes several video frames instead of full videos. However, for research and study purposes, \videollavas are currently the state-of-the-art for these tasks.

\section{\benchname Benchmark}

Finally, we introduce \benchname, a benchmark specifically designed to highlight tasks that popular MLMs are still struggling with. With \name system, we are able to automatically identify the tasks challenging for given MLMs within a given budget. The resultant \benchname provides a comprehensive picture of the limitations of current MLMs. 

\subsection{\benchname Generation Process.}
We conducted Top-K queries with active approximation algorithms on popular open-source MLMs across all task types. For each task type, we use a  budget of 300 model inferences for each model to query the worst-performing tasks.

In particular, for ImageQA, we use the following models: \instructblips, \instructblipl, \qwenvl, \qwenvlchat, \llavas, \llaval, \glmfourv, \cogvlmtwo, \ideficstwo, \phivision, \paligemma, and \internvlchat; 
for VideoQA, we use: \videochatgpts, \videollavas, \videochattwos, \videollamatwos, \videollamatwol, \chatunivis, \chatunivil, and \internvlchat (by taking combined frames).

For each task type (e.g., 3d-how-many), we collected the Top 10 worst-performing tasks for each model in succinct and detailed prompts. By combining the worst-performing tasks of each model, we obtained 12,270 ImageQA and 3,567 VideoQA questions that current popular MLMs are struggling with as our \benchname benchmark.

\subsection{Results and analysis.}
\begin{figure}[h]
  \centering
  \includegraphics[width=\linewidth]{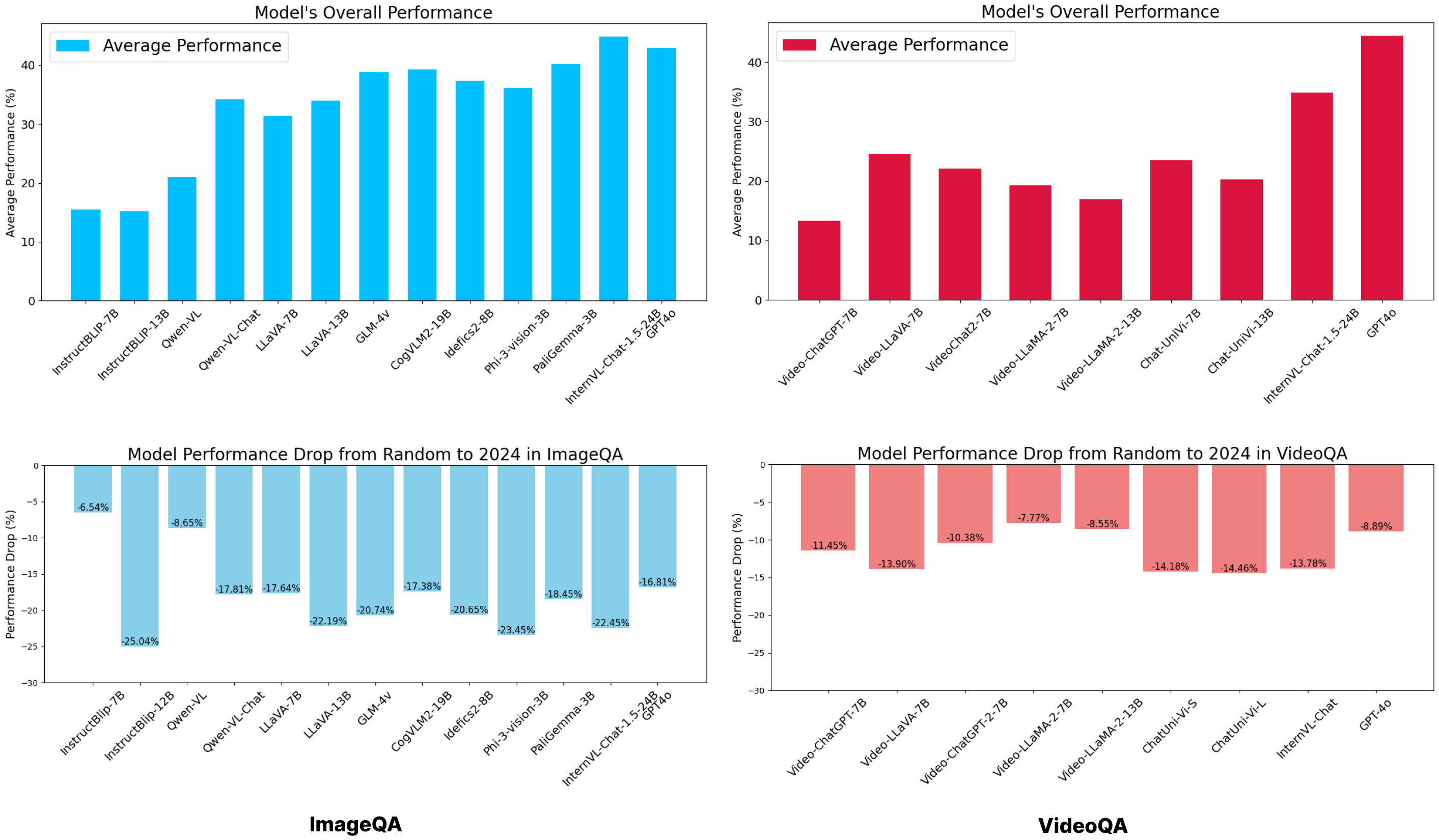}
  \caption{Model's overall performance on \benchname and averaged model's performance drop from \randomname to \benchname.}
  \label{fig:2024-result}
\end{figure}

We then evaluate open-sourced models and \gptfouro on \benchname dataset (Figure \ref{fig:2024-result}). Comparing the models’ performance on \randomname and \benchname, we observed a 10-30\% performance drop for each model. Importantly, \gptfouro, which wasn’t involved in the \benchname generation process, also performed significantly worse. This validates that \benchname is more challenging than \randomname, and our generation process indeed found the questions where popular visual models broadly suffer. Additionally, we observe that the relative performance of models still holds on \benchname. Larger or proprietary models consistently perform better than smaller ones.



\section{Related Work}
\label{sec:rw}

We situate our work amongst existing work on large multimodal language models, programmatic task generation, and model-adaptive testing and debugging.

\paragraph{Large multimodal language models (MLMs).}
In recent years, large multimodal language models, by integrating visual encoders within various pretrained large languages models~\cite{Wang2024InternVideo2SV, huang2023vtimellm, chen2023videollm, wang2023gpt4video, sun2023finegrained, lyu2023macawllm, tang2023llmvagebc, wang2023chatvideo, lin2023mmvid, bi2023misar, chen2023groundingprompter, liu2024prismer, peng2023kosmos2, chen2023pali3, shukor2023unival, lin2023mmvid, lu2023chameleon, li2023mimicit, sun2024emu, moor2023medflamingo, awadalla2023openflamingo, sun2024generative}, have progressively driven advancements in visual-language learning. With ubiquitous open-sourced LLM backbones and the increasing data for visual instruction tuning. Models like InstructBlip~\cite{dai2024instructblip}, QwenVL~\cite{bai2023qwen}, LLaVA~\cite{liu2024visual}, InternVL~\cite{chen2023internvl}, etc, have achieved unprecedented visual understanding performance in nearly all kind of visual tasks. Not only for static images, in the filed of video, by adding temporal information into the training and fine-tuning process. Models like VideoLLaMA~\cite{damonlpsg2023videollama}, VideoChatGPT~\cite{Maaz2023VideoChatGPT}, ChatUnivi~\cite{jin2023chatunivi}, VideoLLaVA~\cite{lin2023video}, and VideoChat2~\cite{li2023mvbench} have extended their capabilities to encompass video. These models, take both visual content and language as input and output language, are being considered as a new type of foundation model. 
The rise of large multimodal models has catalyzed the evolution of multimodal benchmarks~\cite{Fu2024VideoMMETF, Wang2023InternVidAL, Zhang2023ASL, Lei2018TVQALC, Zhang2023MoVQAAB, Rawal2024CinePileAL, Huang2023VBenchCB, Mangalam2023EgoSchemaAD, zhang2024benchmarking, tong2024eyes, fu2023challenger, cai2023benchlmm, zhang2023m3exam, cui2023holistic, huang2023sparkles, ge2024mllmbench, liu2023mmc, ning2023videobench, liu2024tempcompass}, making them both broader and deeper. On the breadth axis, works such as MMBench\cite{liu2023mmbench}, SEED-Bench \cite{li2023seed, li2023seed2} and MMMU \cite{yue2023mmmu} provide comprehensive and integrated VQA benchmarks to evaluate a model's performance overally. On the depth axis, efforts like MathVista~\cite{lu2023mathvista}, Blink \cite{fu2024blink}, MultipanelVQA \cite{fan2024muffin}, and Lance \cite{prabhu2024lance} focus on specific areas of visual tasks, such as spatial reasoning, multipanel images understanding, counterfactual images understanding, etc. To evaluate the models' ability in specific domains or tasks.

\paragraph{Programmatic task generation.} 
Leveraging program to generate scalable and controllable benchmark data to evaluate models has been explored in various tasks, Within the task of VQA. Early attempts like the CLEVR~\cite{johnson2017clevr} dataset, which generates simple 3D shapes to test models' visual reasoning, GQA dataset\cite{hudson2019gqa}, using programs to generate questions from real images have achieved great success. The advent of vision models has given them the ability to tackle more complicated and compositional vision tasks, and the need for comprehensive and complex programmatic benchmarks has emerged. SimVQA\cite{cascante2022simvqa}, integrated 3D models and simulated 3D environments, to generate photo-realistic, multi-physics synthetic scenarios with questions. Moreover, leveraging the advantages of programmatic benchmark generation, such as those used in 3DB~\cite{leclerc20223db}, allows for precise targeting and identification of subgroups where models underperform.

\paragraph{Model-adaptive testing and debugging.}
In the past decades, we used the static "training set, test set" paradigm to evaluate the model's performance. However, as the foundation models are all trained on a wide spectrum of datasets, this paradigm might face overfitting and data contamination issues, which makes it hard to evaluate the performance of a model fairly and truly.
Model-adaptive testing and debugging, consequently, emerges to solve this problem. The key idea is 1): dynamically update the test data to prevent overfitting and data contamination. Dynabench~\cite{dynabench}, for instance, uses human and model collaboration to create challenging benchmarks. Additionally, LatestEval~\cite{li2023avoiding} uses the latest texts to evaluate the model, avoiding training data overlap, and \cite{ying2024have} automates dataset updates through stylistically similar samples generated by LLMs.
2): adaptively identify subgroups where models underperform and adjust task ratios accordingly. AdaVision~\cite{gao2023adaptive}, an interactive tool for iterative testing and refinement of computer vision models, pinpoints and addresses their systematic failures with user involvement. Moreover, \cite{van2024can}'s 3S Testing employs synthetic data to focus evaluations on minority subgroups and distributional shifts. Lifelong Benchmarks~\cite{prabhu2024lifelong} proposes dynamically expanding benchmarks and an innovative algorithm to handle the increasing data and evaluation demands efficiently.

\section{Conclusion}

In this work, we introduce \name, a task generation and evaluation system designed to address user queries with different evaluation objectives. 
We conduct various analyses and case studies based on \name and existing MLMs, and offer many insights to the headroom for future model improvements. 
There are some limitations in this first version of \name. For example, the current task space is more about models' perceptual capabilities and don't test for complex reasoning capabilities, which we plan to address in future versions by adding more task generators into \name.

\section*{Acknowledgement}
This project was partially funded by Toyota Motor Corporation and OpenAI Superalignment Fellowship.

\clearpage

\appendix

\clearpage
{\hypersetup{hidelinks}
\tableofcontents
}
\clearpage

\section{Discussion}
\label{app:discussion}

\subsection{Limitation}

\paragraph{Programmatically generated tasks can be unrealistic and biased.}

Programmatically generated tasks can lack the complexity and variability found in real-world data. These tasks might not capture the nuances of real-world scenarios, leading to models that perform well on synthetic data but fail in practical applications.
The constraints and rules defined in the code may oversimplify the tasks, making them easier for models to solve compared to real-world tasks. This can result in overestimating a model's capabilities.
The rules and logic used to generate tasks can inadvertently introduce biases. For example, if the code disproportionately generates certain types of objects or scenarios, the model may not be adequately tested on a diverse range of tasks.

\paragraph{Designing the task space is challenging.}

Identifying and defining the relevant attributes for each task type (e.g., object recognition) requires deep domain knowledge and understanding of what aspects are critical for evaluating model performance.
The task space must be comprehensive enough to cover various scenarios but not so complex that it becomes infeasible to manage or evaluate. Striking this balance is a significant challenge.
The task space should be designed to ensure comprehensive coverage of all relevant scenarios and diversity in the types of tasks. This requires meticulous planning and consideration of all possible task variations.

\paragraph{Adding new task generators requires coding skills.}
Adding new task generators involves programming and understanding the underlying framework used for task generation. This requires technical expertise, which may not be available for all communities and can be a barrier for non-technical researchers who might have valuable insights and ideas for new tasks but lack the coding ability to implement them.

\paragraph{Query results approximation can be inaccurate.}

Efficient query results approximation within certain budgets might sometimes yield inaccurate results, especially when the budget limits are constrained. This inaccuracy can stem from several factors. First, the models that embed tasks into vectors may not fully capture all the details and nuances between different tasks. Second, the algorithms used for querying might have inherent limitations or room for improvement, affecting the precision of the results. Addressing these issues requires ongoing refinement of both the task embedding models and the query algorithms to enhance their ability to deliver accurate approximations under varying computational budgets.

\subsection{Potential negative social impact}

\paragraph{Misuse for malicious benchmarks.} \name's ability to generate a vast number of tasks could be misused to create benchmarks specifically designed to trick or expose vulnerabilities in AI systems. Malicious actors might use this capability to create benchmarks that mislead researchers or lead to the development of AI models with undesirable biases or vulnerabilities.

\paragraph{Reinforcement of biases and discrimination.} If \name's task generators are not carefully designed and curated, they could inadvertently perpetuate existing biases present in the source data. This could lead to the development of AI models that are biased against certain groups of people or perpetuate harmful stereotypes.

\paragraph{Overreliance on synthetic tasks.} The focus on synthetic task generation could lead to a disconnect between evaluation results and real-world performance. Overreliance on synthetic tasks might create a false sense of progress and hinder the development of AI models that can effectively address real-world challenges.

\paragraph{Data contamination.} Fine-tuning models on synthetic tasks generated by \name could lead to data contamination, where the model learns to exploit the specific patterns and biases of the synthetic data rather than generalizing to real-world scenarios. This could result in models that perform well on synthetic benchmarks but poorly in practical applications.

\paragraph{Access and fairness.} While \name aims to democratize AI evaluation, the technical expertise required to implement new task generators could create barriers for researchers and practitioners from underrepresented groups, leading to a lack of diverse perspectives and potentially reinforcing existing inequalities.

\subsection{Future work}

\paragraph{Supporting natural language user queries.}
We plan to enable natural language queries, allowing users to specify evaluation needs in plain language. This will leverage language models to translate instructions into actionable query commands, making the system more accessible and user-friendly. This enhancement will democratize access to model evaluation, streamline the process, and reduce barriers for non-technical users, fostering a more inclusive evaluation ecosystem.

\paragraph{Expanding the \name system.}
To further enhance the capabilities of \name, we plan to extend it across a broader range of scenarios and model types. This involves integrating support for various generative models, including language models and visual generative models, which can fine-tune the evaluation of generation quality. Also, by incorporating new types of source data, we aim to enrich the diversity and relevance of the tasks generated, ensuring that the evaluation framework remains robust and comprehensive as foundation model capabilities advance. Additionally, developing new task generators will enable the creation of tasks that capture emerging AI challenges and applications, facilitating continuous adaptation to the evolving landscape of AI. This expansion will empower users from different domains to evaluate models in ways that are highly specific to their needs, ultimately contributing to more targeted and effective deployment of AI technologies.

\paragraph{A new workload for database study.}

\name presents new opportunities for the database community to develop efficient query execution techniques on conceptual relations containing model inference results (e.g., task accuracy of many models on many tasks) that are expensive to compute and often unmaterialized when a query is issued. 
The idea of pre-filtering to avoid expensive computation has been proven to be effective in some database problems, such as accelerating similarity joins~\cite{mann2016anempirical, jiang2014string} and video analytics queries~\cite{kang2018blazeit} where computing the similarity function or running model inference on videos is expensive during query execution. 
In a similar vein, recent work~\cite{he2023masksearch, DBLP:journals/pvldb/HeDCB21, vartak2018mistique} has proposed efficient database indexing and query execution techniques to navigate the tradeoffs between storing the model inference results on disk and computing them on-the-fly at query time. 
Some other efforts~\cite{agarwal2013blinkdb} have also proposed trading off query result accuracy for query response time.
Another direction for future work is query result diversification. 
When a practitioner explores a set of MLMs, datasets, and tasks, they may desire to examine a diverse set of result items, e.g., tasks that are dissimilar. 
It would be interesting to how query result diversification techniques~\cite{ge2020efficient, hirata2022solving} could be adapted in \name's setting.


\clearpage

\section{Details of Task Generation}
\label{app:task-generation}

In this section, we describe the details of the programmatic task generation process in \name. We focus on tasks of multiple-choice visual questions answering, including both image question answering (ImageQA) and video question answering (VideoQA).

\subsection{Key concepts}

First, we introduce several key concepts and definitions in our task generation process.

\paragraph{Task instance, task, and task plan.}
A task instance is an image/video, question, options, and ground truth answer tuple that comprises a single evaluation test-case. A task is a conceptual abstraction consisting of all task instances that share the same question and answer.
Tasks are specified via task plans, which contain the required task metadata and configurations to create the actual task instances. For example, in tasks involving counting, the task plan specifies the categories of objects, their total numbers in the scene, and their positions in the image—such as two apples, one on the top right and one on the bottom left. The task instance then features an actual image of the target objects and includes a specific question and answer that is consistent with the arrangement of these objects in the scene. One such task instance might be an image with two apples, the question: "How many apples are there in the image?",  and the answer: "2". Multiple task instances can be generated from a single task plan because other elements such as the image background and types of distractor objects can be randomized, as they are not specified in the task plan.

\paragraph{Source data.} 
We refer to source data as the visual data and annotations that are used to generate task instances, \eg, the 3D objects from Objaverse~\cite{deitke2023objaverse, deitke2024objaverse} and their associated annotations or the real images and scene graphs from GQA~\cite{hudson2019gqa, krishna2017visual}.

\paragraph{Task generator.}
Each task generator is a program that, given source data as input, generates task instances of a certain type. It achieves three main purposes: 1) it defines the schema of the task plan; 2) it can enumerate all possible task plans given the available source data; and 3) given source data and a specific task plan, it can randomly generate a task instance belonging to the task family defined by the task plan. 

\subsection{The generation process}

Given the source data and a task generator, one can readily generate a large number of tasks. The overall generation process consists of the following steps:

\paragraph{Step 1: enumerate the task plans.}
Once the task generator is implemented, one can use it to enumerate and return all the possible task plans based on the defined schema and the source data. As each task plan consists of just the metadata of the task rather than the actual task instances, it is efficient to enumerate all the task plans and store them as a single table. Note that enumerating all possible task plans is a one-time job, since the table of task plans can be stored and reused.

\paragraph{Step 2: generate task instances of a task given its task plan.}
Another core functionality of the task generator is to generate one task instance given a valid task plan. 
Note that the task generator may generate many different task instances because of the randomness, \eg, the negative choices can be randomly sampled from possible candidates, yet since they are all generated by the same task generator with the same task plan, they would share the question and ground truth answer and are considered belonging to the same task.

\paragraph{Properties.}
This task generation process exhibits several key properties:
\begin{itemize}
    \item \textbf{Reproducible:} With our task generation process, the tasks are produced as a combination of the source data and the programs, therefore one can reproduce identical task instances with the same source data and the random seed of the program. 
    \item \textbf{Scalable:}  This task generation process is scalable for two reasons. First, it is \emph{memory-friendly}. One only needs to store the source data and the annotations, as well as our codebase. Even when one aims to evaluate a model on millions of task instances, since the task instances are reproducible, one can choose to generate the task instances on the fly rather than beforehand. Secondly, it is \emph{easy to expand} the space of task that can be generated. One can increase the number of possible tasks by either adding new source data or new task generators. 
    \item \textbf{Easy to update:} Benchmarks can contain unexpected errors, \eg, annotation error~\cite{northcutt2021labelerrors}, so the task generation process must be easy to update once the error is caught. Since our task generation process is transparent to the users, once an error is caught, it can immediately be attributed to either the error of the source data or bugs in the code of the task generators, and then be fixed. We welcome the whole community to report any flaw in our task generation process. 
    \item \textbf{Structured task space:}  Finally, each task generated by our approach is associated with a task plan composed of its metadata. This design offers a natural structure for the tasks so that they can be grouped by certain specifications of task metadata. It enables users to navigate wanted tasks by querying the table of task plans as querying a normal database. Also, it facilitates the diagnosis of models according to the task metadata.
\end{itemize}

\begin{figure}[!h]
\includegraphics[width=\linewidth]{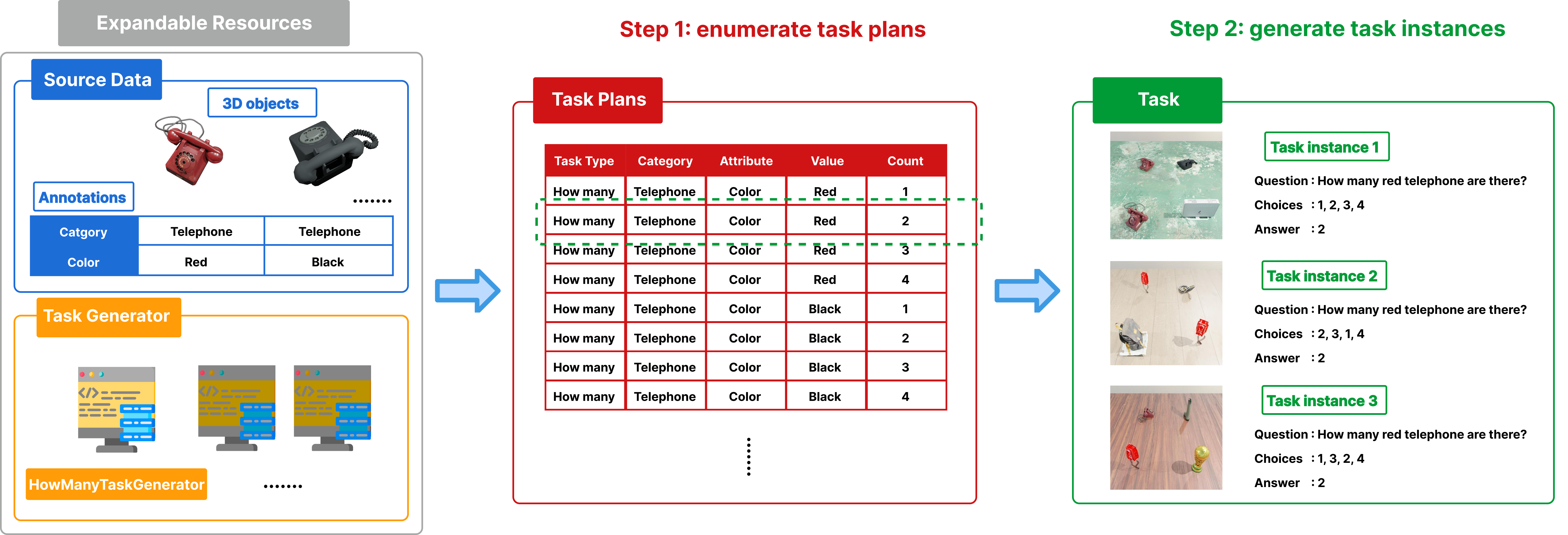}
  \caption{An illustration of core concepts and the task generation process.}
  \label{fig:gen}
\end{figure}

\clearpage

\section{Details of Fine-grained User Query and Query Approximation Algorithms}
\label{app:query}

With \name, most user queries regarding model performance can be simply addressed by identifying the relevant task generators and a subset of the task plans to generate task instances for model investigation.
However, there is a special family of fine-grained user queries regarding individual tasks and taxonomy concepts that may require a large number of tasks to be appropriately addressed.
For example, \emph{the colors that the minimum performance of models M1, M2 is larger than 50\%}; such a query involves tasks related to all the color attributes and concerns the models' performance on each individual color.
In this section, we outline four types of such fine-grained user queries and discuss how to address them with efficient query results approximation.

\subsection{Fine-grained user query}

We introduce four types of fine-grained user query. By default, the target of a query is the tasks, \eg, Top K <task>; 
one can also query different task metadata or their products, \eg, Top K <category> or Top K <category $\times$ attribute>.

\paragraph{Top-K query.}
Users may be interested in knowing the tasks or task metadata (\eg, object category) that the model(s) performs the best or the worst, which can be supported by a Top-K query. 
An example Top-K query in natural language is, 
\emph{(E1) Top 10 “how many” tasks ranked by the maximum performance of the user-specified list of models (the user specifies all models in this case) in descending order}. 
This query finds the top 10 tasks that all models perform the best, measured by the maximum performance of the models on each task.

\paragraph{Threshold query.}
Another useful type of query is the Threshold query, since users may want to know the tasks or task metadata on which the model's performance is larger or lower than a given threshold. 
An example in natural language is, 
\emph{(E2) The color attributes on which the mean of the minimum performance of models M1, M2 is larger than 50\%}. 
The query first groups tasks by their color value attribute and then aims to find the groups where the mean of the minimum performance of M1 and M2 across all tasks in the group is larger than 50\%.

Built upon basic queries, one can develop new types of queries to fulfill specific needs, \eg, comparing models or diagnosing the model. 
Here, we showcase two advanced queries based on the Threshold query: model compare and debug. 

\paragraph{Model Comparison query.}
A useful type of query is to support comparing a model to another. In contrast to the traditional way of comparing models by ranking based on their performance, our \emph{Model Comparison Query} supports finding tasks or patterns where one model performs better than the other by a given threshold. 
An example query is \emph{(E3) The task types on which the mean performance of model M1 is larger than model M2}.

\paragraph{Model Debugging query.}
Model debugging is an important field of study for model evaluation~\cite{}, where the goal is to find patterns or subgroups where the model performs significantly worse or better than its average performance.
To fulfill this need, we support \emph{Model Debugging Queries} by leveraging the Threshold query with the threshold being a function of the model's average performance and a hyperparameter. 
For example, to find tasks where the model performs significantly worse than average, we can use the Threshold query and set the threshold to be $\mu - \sigma$, where $\mu$ is the averaged performance of the model and $\sigma$ is the standard deviation of the model performance. 
An example query is \emph{(E4) The tasks on which the performance of model M1 is lower than its average performance of all tasks by a standard deviation}.

Note that these two types of query can be similarly defined based on the Top-K query, \eg, the Model Debugging query can be the top k tasks that a model performs the worst, and how to define these queries depends on the user need.

\subsection{Query execution}
\label{sec:query-exe}

\begin{figure}[!h]
  \centering
  \includegraphics[width=\linewidth]{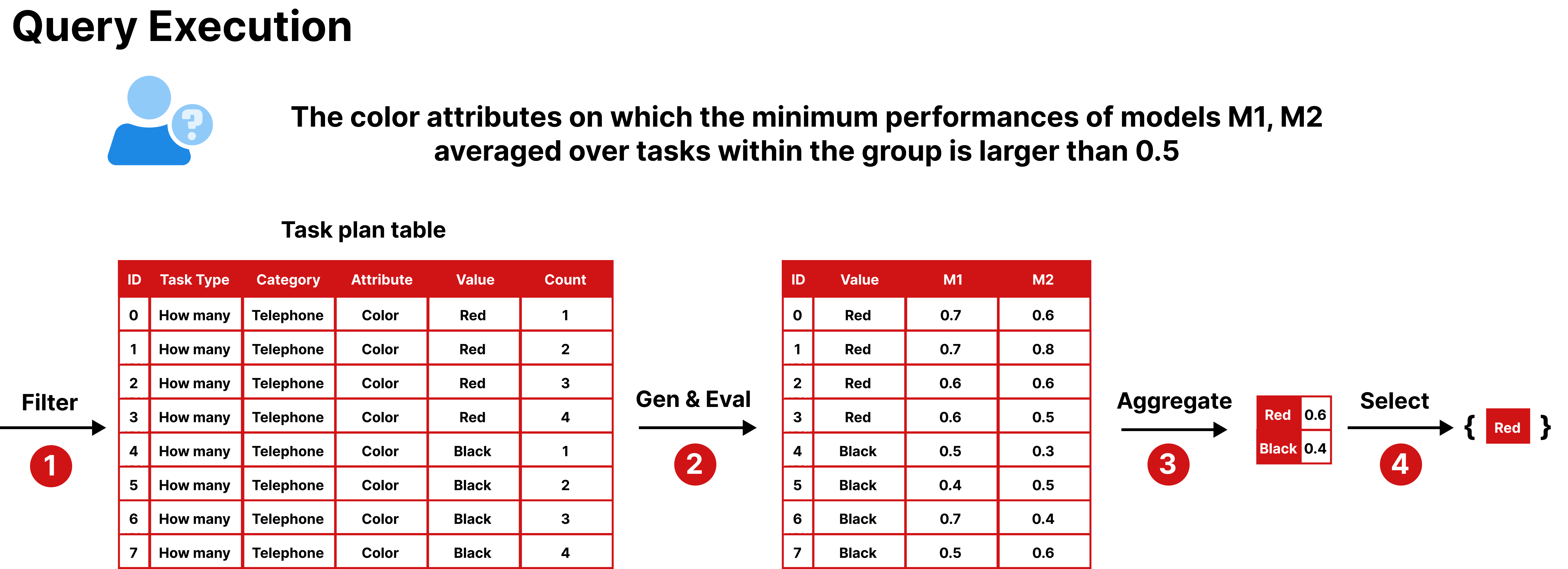}
  \caption{An illustration of the query execution process.}
  \label{fig:qe}
\end{figure}

We provide an example of the conceptual query execution process in Figure~\ref{fig:qe}, which illustrates the steps required to execute query E2. 
Query E2 requires these steps:

\begin{enumerate}
    \item Filter: the query filters the task plans related to “color”. 
    \item Generate and evaluate: the query needs to generate the tasks given the obtained task plans and then evaluate model M1 and M2 against these tasks to collect their accuracy for each task. 
    \item Aggregate: once we obtain models' accuracy on every involved task, we perform some aggregate functions to collect the final results. We first compute the minimum accuracy of models M1 and M2 on each task. Then we average the obtained minimum accuracy over tasks within one color value group, to gather the final results for each color value group.
    \item Select: for each group, the query checks whether the final result is greater than 0.5 and only keeps the groups where this filter condition holds.
\end{enumerate}

\paragraph{Incorporating frequent pattern mining.}
In practice, users may be more interested in knowing the patterns revealed by the returned tasks than the tasks themselves.
Because each task in our system is associated with a task plan, one can apply frequent pattern mining~\cite{han2022data,wang2004bide,han2001prefixspan} to extract frequent patterns from the set of task plans associated with the returned tasks. 
Note that frequent pattern mining can be applied to the results of any type of query as long as there is a set of associated task plans.

\subsection{Efficient Query Approximation Algorithms}

As the fine-grained user queries may involve a large number of tasks to evaluate and therefore likely become computationally infeasible due to the compute-intensive nature of MLMs, we study three algorithms to approximate the query results given a budget of $B$ on the number of tasks to be evaluated.

\paragraph{Subset proxy.}
One straightforward approach to approximate the query results is to spend the budget randomly sampling $B$ tasks and then evaluate the models against them to obtain the results.
Then, we use this sampled subset as a proxy of the whole set of tasks to perform the fine-grained user query.

\paragraph{Fitting.}
Built upon the subset proxy method, the fitting method uses the evaluation results of the $B$ randomly sampled tasks to train a model (referred to as \emph{function approximator}) to approximate the function of interest, and then apply the model to the rest of the tasks to predict the results. 
In particular, the function of interest can be the model's accuracy function which inputs a task and predicts the model's accuracy, or the task aggregate function, \eg, the minimum accuracy of two models as in query E2.
Finally, we perform the query over all the tasks, with both actual evaluation results on $B$ sampled tasks and values of the remaining predicted by the function approximator.

\paragraph{Active evaluation.}
The third approach, active evaluation, builds upon the fitting method but enhances it by strategically selecting tasks to improve the approximation of query results, as opposed to relying on random sampling. This method utilizes an iterative process, where each step involves selecting a batch of unevaluated tasks based on predictions made by the current function approximator. These tasks are then evaluated, and the results are used to re-fit the function approximator with both existing and new data until the evaluation budget is exhausted. Ultimately, the query is executed using a combination of actual results from evaluated tasks and predicted results, similar to the fitting method.
The task selection criteria are tailored to the specific type of query. 
For the Top-K query, it selects the top-K tasks most likely to fulfill the user’s inquiry based on the predicted values, because these tasks are predicted to have the most significant impact on the outcome of the query, and focusing on them could help learn a function approximator with more accurate predictions in areas that are likely relevant to the actual query results.
For the Threshold query, it selects the tasks whose predicted values are closest to the threshold, because these tasks are most likely to influence the decision boundary of the function approximator and thus are critical for accurately determining the boundary’s position within the task space.

\paragraph{Implementation details.}
To learn a function approximator to predict the value of interest, we first need a representation of each task as the input of the approximator. 
We construct such representation using the task plan, question, and answer associated with each task. In particular, we convert these elements into a piece of formulated text and leverage pre-trained embedding models to calculate the text embedding as the task embedding.
We adopt Gaussian Process regressor\footnote{\url{https://scikit-learn.org/stable/modules/generated/sklearn.gaussian_process.GaussianProcessRegressor.html}} because of its stable performance in our preliminary experiments, while any regression model is applicable.

\clearpage

\section{Details of \name 1.0}
\label{app:system-1.0}

In this section, we introduce the task generators implemented in the first version of \name.  
Inspired by the model cards for model reporting~\cite{mitchell2019model}, we make a task generator card for each implemented task generator, including information such as task type, task plan schema, \etc, available in the appendix, and the template can be found in Figure~\ref{fig:task-card}.

\begin{figure}[!h]
\begin{framed}
 \centering
{\Large {\bf Task Generator Card Template}}
\begin{itemize}[leftmargin=*]

\item {\bf Basic Information}. 
\begin{itemize}
\item {\bf Task Type}.  The target type of task, \eg, ImageQA
\item {\bf Question Type}.  The type of generated question, \eg, "how many"
\item {\bf Answer Type}.  The answer type \eg, integer number or object category
\item {\bf The model capability to evaluate}. \eg, counting 
\end{itemize}

\item {\bf Source Data}.  The source data and annotations it requires
\item {\bf Task Plan Schema}.  The schema of the associated task plans

\item {\bf Partitions}.  The partition of the task space. 
\begin{itemize}
\item {\bf Partition 1}.
\begin{itemize}
\item {\bf Template}. Template used to generate question if available
\item {\bf Example}. An example of generated test case
\end{itemize}
\end{itemize}

\item {\bf Limitations}
\item {\bf Recommendations}
\vspace{-.25em}
\end{itemize}
\end{framed}

\caption{Summary of task generator card sections and suggested prompts for each. Task generator cards for all the included task generators can be found in Appendix~\ref{app:cards}. }\label{fig:task-card}

\end{figure}

\subsection{Source data}

\paragraph{3D objects with annotations.}
We start by selecting objects from Objaverse-LVIS, the subset of Objaverse 1.0~\cite{deitke2023objaverse} that has been annotated with LVIS~\cite{gupta2019lvis} categories. From the set of 47K objects spanning 1,230 categories that comprise Objaverse-LVIS, we select 1,996 objects spanning 337 categories. These objects were manually chosen for their high quality and strong category alignment. We use Blender~\cite{blender}, an open-source ray-tracing software, to render each object from a uniform set of surrounding viewpoints and, following manual verification, only keep renderings where the object's category and attributes are discernible. This gives us a set of viewpoint annotations that we also use when constructing 3D scenes, as they allow us to ensure that the object's category and attributes are perceivable from the camera.

\paragraph{Real images and videos with \sg.}
We also collect real images and videos with scene graph~\cite{krishna2017visual} as part of our source data. In particular, we collect real images with scene graphs from the GQA dataset~\cite{krishna2017visual,hudson2019gqa} and real videos with scene graphs from the AGQA dataset~\cite{ji2020action,sigurdsson2016hollywood}.

Additionally, we normalized the object terms across all source data and built a taxonomy containing 927 concepts and 965 edges using Wikidata and human filtering to avoid concept conflicts in options, such as listing both "apple" and "fruit" as choices.

\subsection{Task generators for different scenarios}

\subsubsection{2D sticker image}

The first scenario of \name is \twod, where we compose task instance images by compositing pre-rendered object images into a 2x2 or 3x3 grid.
Such a simple type of image already enables the generation of basic types of visual questions regarding recognizing object categories and attributes, spatial relations, and counting. For example, one task could be \emph{how many red telephones are there in the image?}.
We list the task generators implemented for \twod and the statistics in Table~\ref{tab:task-2d}.

\begin{table*}[h]
  \centering
  \small
  \caption{\twod}
  \resizebox{\linewidth}{!}
  {
    \begin{tabular}{lllr} 
    \toprule
     {\bf Task generator} & {\bf Example question} & {\bf Example answer} & {\bf \# of tasks}\\ 
    \midrule
    
    \multirow{3}{*}{how many}
    & How many blue objects are there in the image? & 2 & 494\\
    & How many tables are there in the image? & 4 & 6,136\\ 
    & How many pink beverages are there in the image? & 2 & 27,027\\ 
    \midrule
    
    \multirow{2}{*}{what}  
    & What is the object in the bottom middle part of the image? & folding chair & 33,163\\ 
    & What is the object to the left of the telephone? & table lamp & 61,648,184\\  
    \midrule
    
    \multirow{2}{*}{where} 
    & Where is the apple in the image? & back left & 33,163\\ 
    & Where is the vacuum cleaner with respect to the backpack? & left & 61,648,184\\  
    \midrule
    
    \multirow{2}{*}{what attribute} 
    & What is the material of the object in the middle part of the image? & plastic & 27,027\\ 
    & What is the color of the object to the left of the silverware? & gold & 50,175,008\\  
    \midrule
    
    \multirow{2}{*}{where attribute}
    & Where is the white object in the image? & top right & 27,027\\ 
    & Where is the gray object with respect to the lollipop? & top & 50,175,008\\  
    \midrule

\multicolumn{4}{c}{\bf Total number of tasks: 223,800,421} \\ 
    
    \bottomrule 
    \end{tabular}
  }
  \label{tab:task-2d}
\end{table*}

\subsubsection{3D tabletop scene}

Although \twod is a useful setting for generating task instances with speed, the artificial way in which the scenes are constructed through image compositing limits their realism. A real-world scene would come from objects existing in a shared 3D space that is rendered through the perspective of a single camera. As such, in \twod we are unable to understand the effects of depth, lighting and occlusion on image understanding. To remedy this, we introduce \threed, a setting analogous to \twod, wherein objects are arranged on a plane in a shared 3D scene and rendered from a fixed camera viewpoint. This allows us to port all of the task generators from \twod while also allowing us to test 3D-specific capabilities such as relative depth.

\paragraph{ImageQA.}
Another way to generate similar yet more realistic images is to compose a 3D tabletop scene using the objects, and then render a 2D image~\cite{johnson2017clevr}.
For this \threed, we can reuse task generators of \twod with some minor modifications regarding the spatial relations.
For example, the spatial relation of "in the bottom of" would become "in front of".
In addition, we identify two families of task generators unique to 3D scenes: tasks regarding the size and distance of objects, which are not suitable for the 2D scenario discussed above.
We list the task generators implemented for ImageQA of \threed and the statistics in Table~\ref{tab:task-3d-image}. 

\paragraph{VideoQA.}
In addition to the aforementioned ImageQA tasks, we also build VideoQA tasks for \threed. We leverage two temporal attributes, rotation and movement, which can only be identified via video, to construct video-specific task generators and evaluate the models' performance in understanding temporal dynamics. 
To generate these videos, we keep the same layout of the 3D tabletop scene as ImageQA, but change the positions and angles of the objects across different frames of the video to make the objects move and rotate. Our task generators then target the model's ability to understand these temporal changes in object position and orientation.  We list the task generators implemented for VideoQA of \threed and the statistics in Table~\ref{tab:task-3d-video}. 

\begin{table*}[h]
  \centering
  \small
  \caption{\threed with images}
  \resizebox{\linewidth}{!}{
    \begin{tabular}{lllr} 
    \toprule
    {\bf Task generator} & {\bf Example question} & {\bf Example answer} & {\bf \# of tasks}\\ 
    \midrule

    \multirow{3}{*}{how many}   & How many blue objects are there in the image? & 6 & 494\\ 
    
    & How many plates are there in the image? & 5 & 6,136\\ 

    & How many black furnitures are there in the image? & 4 & 27,027\\
    
    \midrule
    
    \multirow{2}{*}{what}  
    & What is the object in the front right part of the image? & scale & 33,163\\ 
    & What is the object to the right of the mobile computer? & bucket & 61,648,184\\  
    \midrule
    
    \multirow{2}{*}{where} 
    & Where is the vacuum cleaner in the image? & back left & 33,163\\ 
    & Where is the vacuum cleaner with respect to the wine glass? & left & 61,648,184\\  
    \midrule
    
    \multirow{2}{*}{what attribute} 
    & What is the color of the object in the back left part of the image? & red & 27,027\\ 
    & What is the material of the object behind the plate? & wood & 50,175,008\\  
    \midrule
    
    \multirow{2}{*}{where attribute}
    & Where is the wood object in the image? & front right & 27,027\\ 
     & Where is the white object with respect to the trophy? & left & 50,175,008\\  
    \midrule
    
    \multirow{1}{*}{what size} & What is the smallest object in the image? & spatula & 20,408\\  
    \midrule
    
    \multirow{1}{*}{what attribute size} & What is the color of the smallest object in the image? & black & 16,632\\  
    \midrule
    
    \multirow{2}{*}{where size} 
    & Where is the largest object in the image? & back left & 20,408\\ 
    & Where is the smallest object in the image with respect to the car? & front & 56,906,016\\  
    
    \midrule
    
    \multirow{1}{*}{what distance} & What is the object that is farthest from the optical instrument? & juice & 61,648,184\\  
    \midrule
    
    \multirow{1}{*}{what attribute distance} & What is the color of the object that is closest to the statue? & beige & 50,175,008\\  
    \midrule
    
    \multirow{1}{*}{where distance} & Where is the object that is farthest from the bread in the image? & middle & 61,648,184\\  \midrule

\multicolumn{4}{c}{\bf Total number of tasks: 454,235,261} \\ 
    
\bottomrule 
\end{tabular}
}
  
  \label{tab:task-3d-image}

\end{table*}

\begin{table*}[h]
  \centering
  \small
  \caption{\threed with videos}
  \resizebox{\linewidth}{!}{
    \begin{tabular}{lllr} 
    \toprule
     {\bf Task generator} & {\bf Example question} & {\bf Example answer} & {\bf \# of tasks}\\ 
    \midrule
    
    \multirow{2}{*}{what rotate video}  
    & What is the object that is rotating counterclockwise in the video? & pants & 20,408\\ 
    & What is the rotating object in the video? & jewelry & 20,408\\  
    \midrule
    
    \multirow{2}{*}{what attribute rotate video} 
    & What is the color of the object that is rotating clockwise in the video? & beige & 16,632\\ 
    & What is the color of the rotating object in the video? & yellow & 16,632\\  
    \midrule
    
    \multirow{2}{*}{where rotate video} 
    & Where is the stepladder with respect to the rotating object in the video? & back & 51,631,112\\ 
    & Where is the object that is rotating counterclockwise with respect to the microscope in the video? & front left & 62,221,736\\  
    \midrule
    
    \multirow{2}{*}{what move video}  
    & What is the object that is moving left in the video? & serving tray & 40,816\\ 
    & What is the moving object in the video? & barrel & 40,816\\  
    \midrule
    
    \multirow{2}{*}{what attribute move video} 
    & What is the color of the object that is moving left in the video? & black & 33,264\\ 
    & What is the color of the moving object in the video? & white & 33,264\\  
    \midrule
    
    \multirow{2}{*}{where move video} 
    & Where is the object that is moving down located in the video? & back right & 40,816\\ 
    & Where is the moving object located in the video? & back right & 40,816\\  
    \midrule

\multicolumn{4}{c}{\bf Total number of tasks: 114,176,720} \\ 
    
    \bottomrule 
    \end{tabular}
  }
  \label{tab:task-3d-video}
\end{table*}

\subsubsection{Real images/videos with scene graphs}

We also leverage existing manually-annotated scene graph data, \ie, GQA and AGQA, to construct task generators.
For ImageQA, because there are three types of nodes in the scene graph for images, \ie, object, relation, and attribute, we accordingly implement three task generators to evaluate models' capability in recognizing these basic visual elements.
Similarly, the scene graph for videos consists of three types of nodes, \ie, object, relation, and action, we implement three task generators regarding these visual elements.
We list the task generators implemented for ImageQA and VideoQA leveraging scene graphs and the statistics in Table~\ref{tab:task-sg-image}\&\ref{tab:task-sg-video}.

\begin{table*}[!h]
  \centering
  \small
  \caption{Real images with \sg}
  \resizebox{\linewidth}{!}{
    \begin{tabular}{lllr} 
    \toprule
    {\bf Task generator} & {\bf Example question} & {\bf Example answer} & {\bf \# of tasks}\\ 
    \midrule

    \multirow{1}{*}{what object}  
    & What is the flat object that is on the brown and wood table? & paper & 25,169\\ 
    \midrule
    
    \multirow{1}{*}{what attribute} 
    & What is the material of the smooth object that is to the right of the yellow container? & plastic & 20,554\\ 
    \midrule
    
    \multirow{1}{*}{what relation} 
    & 
    \begin{tabular}[c]{@{}l@{}}What is the relation from the standing object, which the colorful and long snowboard is to the right of, \\ to the blue and long object, which is to the left of the patterned skis?\end{tabular} 
    
     & holding & 23,241\\ 
    \midrule

\multicolumn{4}{c}{\bf Total number of tasks: 68,964} \\ 
    
\bottomrule 
\end{tabular}
}
  \label{tab:task-sg-image}
\end{table*}
\begin{table*}[!h]
  \centering
  \small
  \caption{Real videos with \sg.}
  \resizebox{\linewidth}{!}{
    \begin{tabular}{lllr} 
    \toprule
    {\bf Task generator} & {\bf Example question} & {\bf Example answer} & {\bf \# of tasks}\\ 
    \midrule

    \multirow{1}{*}{what object video}  
    &  What is the spatial relation of the person to the closet while the person closing a closet? & floor & 428,342\\ 
    \midrule
    
    \multirow{2}{*}{what relation video} 
    & What is the object that the person is behind after the person watching something in a mirror? & behind & 211,983\\ 
    & What is the person doing to the blanket before the person putting a phone somewhere? & touching & 216,359\\ 
    \midrule
    
    \multirow{1}{*}{what action video}  
    & What action is the person doing while laughing at something? & sitting at a table & 335,386\\ 
    \midrule

\multicolumn{4}{c}{\bf Total number of tasks: 1,192,070} \\ 
    
\bottomrule 
\end{tabular}
}
  
  \label{tab:task-sg-video}
\end{table*}

\subsection{\name-UI}
\label{app:interface}

\begin{figure}[!h]
  \centering
  \includegraphics[width=\linewidth]{imgs/interface/interface.pdf}
  \caption{\uiname Interface.}
  \label{fig:interface-app}
\end{figure}

The ultimate goal of our query-centric model evaluation framework is to allow diverse users, including ML practitioners and non-technical users, to understand foundation models’ capabilities and limitations comprehensively and dynamically by answering their various case-specific queries. To achieve this overarching goal, we further break it down into three subgoals and aim to design an interactive end-user interface to achieve these goals:  

\goal{1}: Support understanding of the overall task space and model performance;

\goal{2}: Enable deeper understanding of models through query-centric visualization of model performance (especially for common queries);

\goal{3}: Facilitate model debugging via discovery of surprising results. 

To achieve these goals, we implemented a graphical user interface\footnote{\url{https://huggingface.co/spaces/zixianma/TaskMeAnything-UI}} with Gradio’s \cite{abid2019gradio} framework and used Altair \cite{VanderPlas2018, Satyanarayan2017} for all the visualizations. In this section, we describe our interface in detail and how its components aim to address our design goals. Then, we present several case studies using this interface in the next section. 
Our interface consists of four major components organized as different tabs:

\paragraph{Overall.}  As the name suggests, the Overall tab is designed to help users understand the overall task distribution and model performance (\goal{1}). It consists of two horizontal sections for visualizing overall task distribution (\greencircled{A}) and models’ overall performance (\greencircled{B}) respectively. Section A displays a pie chart of the distribution of all tasks by metadata based on user’s choice of task metadata, while Section B visualizes certain models’ aggregated performance in either a bar plot or heat map according to user-selected models, aggregation method and task metadata. We choose these common chart types in hopes of supporting straightfoward understanding of the overall task space and model performance. 

\paragraph{Task embedding.}
In addition to the overall task distribution, we also include the Task Embedding tab to allow users to visualize all tasks at once in a 2D embedding space (\goal{1}). Concretely, the Embedding tab plots the 2D embeddings of all tasks reduced by UMAP as dots in a scatter plot (\orangecircled{C}). Further, we add a descriptive tooltip for each dot that displays an example image or video along with the corresponding question-answer pair for this task (\orangecircled{D}). By visualizing all tasks in one plot and enabling detail of individual tasks on demand at the same time, we hope the interface can help users understand the entire task space well on both high and low levels. 

\paragraph{Fine-grained user query.}
Most importantly, our interface supports query-centric visualizations of model performance under the Query-centric tab. While the space of possible user queries can be infinite, we define four common user queries: top k, threshold, model comparison and model debugging (Section \ref{sec:query}) and support corresponding visualizations (\redcircled{E}). As these queries involve selecting a subset of tasks for visualization, we include a “Find tasks/task metadata” button to first select the relevant tasks based on the user query and return these tasks in a table (\redcircled{F}). If the user selects task metadata, they will have the option to visualize models’ performance on the selected task metadata (\redcircled{G}). If the user chooses to find individual tasks however, they can additionally visualize the task distribution by some metadata, or find frequent patterns among tasks. By specifying a query first and visualizing models’ performance only on selected tasks/task metadata, users can gain a more targted understanding of models based on what they are interested in (\goal{1}). In particular, the model debugging query can help the user find buggy model behaviors by identifying tasks/task metadata where the model’s performance is lower than its global average accuracy by a large margin i.e. one standard deviation (\goal{2}).

\paragraph{Surprisingness.} Last but not least, we include the Surprisingness tab to help users uncover tasks where models achieve surprisingly good or bad performance compared to their performance on similar tasks (\goal{3}). We define the “surprisingness” of a model $M$ on a particular task $T_i$ as the following: 
For a task, $T_i$ and its $K$ nearest neighbors tasks $\{T_{j}^{\prime}\}$, we compute the surprisingness score as

\begin{equation}
\label{eq:csr}
    s_{i}^{M}=\frac{1}{K}  
    \sum_{j=1}^{K}
    \left( 
    \text{sim}(T_i, T_{j}^{\prime})\times (f(T_i, M) - f(T_{j}^{\prime}, M))
    \right)
\end{equation}

A higher score indicates the model $M$ is much better at task $T_i$ than the neighbor tasks, while a lower score means $M$ is worse at  $T_i$ than the neighbors.

Under the Surprisingness tab, we display the tasks where the model achieves the highest surprisingness scores in a bar chart (\bluecircled{H}). We also make the bar chart interactive so that the user can select a particular surprising task. Then, the scatter plot on the side visualizes this model’s performance on the user-selected task accordingly along with the k most similar tasks in the 2D embedding space (\bluecircled{I}). With this interactive visualization of surprising tasks, we hope to allow users to uncover unexpected model behaviors quickly. 

\clearpage

\section{Details of Model and Human Performance on Random Task Instances}
\label{app:random-result}

In this section, we present the full results of our evaluation on \randomname with 18 MLMs and human anntators.

\subsection{Raw results of Figure~\ref{fig:random-eval}}

\begin{table*}[!h]
  \centering
  \small
  \caption{\textbf{\randomname-ImageQA}. The model performance on random subsets of ImageQA tasks using both the detailed prompt and the succinct prompt. Numbers in parentheses are the number of task instances for each set.}
  \scalebox{0.65}{
    \begin{tabular}{l| cc|cc|cc} 
    \toprule
     & \multicolumn{2}{c|}{\bf \twod} &  \multicolumn{2}{c|}{\bf \threed} & \multicolumn{2}{c}{\bf \sg}\\ 
     
     & \multicolumn{2}{c|}{(1,500)} & \multicolumn{2}{c|}{(3,300)} & \multicolumn{2}{c}{(900)}\\ \cmidrule(lr){2-3}\cmidrule(lr){4-5}\cmidrule(lr){6-7}

 & {\bf Detailed prompt} &  {\bf Succinct prompt} & {\bf Detailed prompt} &  {\bf Succinct prompt}  & {\bf Detailed prompt} &  {\bf Succinct prompt}\\\midrule\midrule

\textbf{Human} & \multicolumn{2}{c|}{99.40} & \multicolumn{2}{c|}{99.73} & \multicolumn{2}{c}{97.33}\\\midrule\midrule

\instructblips & 28.27 & 0.60 & 34.48 & 0.45 & 68.33 & 0.11 \\ \midrule

\instructblipl & 28.34 & 23.87 & 33.12 & 24.73 & 65.22 & 66.11 \\\midrule

\qwenvl & 33.40 & 13.33 & 33.48 & 15.91 & 68.78 & 12.56  \\ \midrule

\qwenvlchat & 40.40 & 35.87 & 38.88 & 39.36 & 78.33 & 79.45 \\ \midrule

\llavas & 37.93 & 41.87 & 37.55 & 39.24 & 62.00 & 75.22 \\ \midrule

\llaval & 45.60 & 43.20 & 43.97 & 42.39 & 79.22 & 82.78 \\\midrule\midrule

\glmfourv & 52.74 & 51.53 & 53.91 & 53.70 & 75.00 & 70.56 \\ \midrule
\cogvlmtwo & 46.20 & 48.93 & 49.94 & 53.00 & 71.33 & 70.66 \\ \midrule
\ideficstwo & 49.20 & 49.40 & 50.67 & 50.00 & 74.33 & 74.11 \\ \midrule
\phivision & 53.60 & 55.60 & 47.00 & 47.91 & 76.22 & 77.22 \\ \midrule
\paligemma & 49.27 & 52.47 & 43.79 & 45.42 & 80.00 & 81.22 \\ \midrule\midrule

\internvlchat & 58.60 & 57.40 & 61.06 & 59.64 & 84.67 & 82.33 \\ \midrule

\llavanext & 62.80 & 62.33 & 56.33 & 58.06 & 85.66 & 84.89 \\ \midrule\midrule

\geminipro & 30.60 & 31.47 & 33.03 & 31.09 & 56.78 & 60.89 \\ \midrule

\qwenvlmax & 55.46 & 53.33 & 53.49 & 55.06 & 85.67 & 89.33 \\\midrule

\gptfourv & 34.60 & 52.40 & 36.73 & 47.55 & 73.44 & 71.78 \\ \midrule

\gptfouro & 45.33 & 54.80 & 46.00 & 58.61 & 76.33 & 77.34  \\

\bottomrule 
\end{tabular}
}
  
  \label{tab:random-image}

\end{table*}
\begin{table*}[h]
  \centering
  \small
  \caption{\textbf{\randomname-VideoQA}. The model performance on random subsets of VideoQA tasks using both the detailed prompt and the succinct prompt. Numbers in parentheses are the number of task instances for each set.}
  \scalebox{0.8}{
    \begin{tabular}{l| cc|cc} 
    \toprule
     &  \multicolumn{2}{c|}{\bf \threed} & \multicolumn{2}{c}{\bf \sg}\\ 
     
     &  \multicolumn{2}{c|}{(1,800)} & \multicolumn{2}{c}{(900)}\\ \cmidrule(lr){2-3}\cmidrule(lr){4-5}

 & {\bf Detailed prompt} &  {\bf Succinct prompt} & {\bf Detailed prompt} &  {\bf Succinct prompt}  \\\midrule\midrule
 
\textbf{Human} & \multicolumn{2}{c|}{98.33} & \multicolumn{2}{c|}{99.33}\\\midrule\midrule

\videochatgpts  & 21.44 & 21.39 & 30.45 & 25.67\\ \midrule

\videollavas & 26.00 & 38.78 & 32.11 & 56.67\\ \midrule

\videochattwos & 30.61 & 28.55 & 37.89 & 32.89 \\ \midrule

\videollamatwos & 23.78 & 16.33 & 36.34 & 31.67\\ \midrule

\videollamatwol & 22.67 & 20.23 & 30.78 & 28.45 \\ \midrule

\chatunivis & 29.72 & 25.95 & 50.11 & 45.00 \\ \midrule

\chatunivil & 28.17 & 25.67 & 45.22 & 39.89 \\ \midrule\midrule

\internvlchat & 38.33 & 31.67 & 68.11 & 56.33 \\ \midrule

\llavanext & 40.06 & 41.17 & 67.55 & 63.44\\ \midrule\midrule

\geminipro  & 31.78 & 30.11 & 50.00 & 45.78\\ \midrule

 \qwenvlmax & 38.89 & 39.39 & 69.11 & 66.78 \\ \midrule

 \gptfourv & 30.95 & 36.83 & 59.11 & 62.67\\ \midrule

\gptfouro & 35.67 & 41.72 & 69.56 & 66.22 \\ 

\bottomrule 
\end{tabular}
}
  
  \label{tab:random-video}

\end{table*}
\clearpage

\subsection{A breakdown of Table~\ref{tab:random-image}}
\label{app:random-image}

\begin{table*}[!h]
  \centering
  \small
  \caption{random-\twod}
  \scalebox{0.75}{
    \begin{tabular}{l| cc|cc|cc|cc|cc|cc} 
    \toprule
     & \multicolumn{2}{c|}{\bf how many} & \multicolumn{2}{c|}{\bf what} & \multicolumn{2}{c|}{\bf what attribute} & \multicolumn{2}{c|}{\bf where} & \multicolumn{2}{c|}{\bf where attribute}\\

    \cmidrule(lr){2-3}\cmidrule(lr){4-5}\cmidrule(lr){6-7}\cmidrule(lr){8-9}\cmidrule(lr){10-11}

 & {\bf DP} &  {\bf SP} & {\bf DP} &  {\bf SP}  & {\bf DP} &  {\bf SP}  & {\bf DP} &  {\bf SP}  & {\bf DP} &  {\bf SP} \\\midrule\midrule

 \textbf{Human} & \multicolumn{2}{c|}{100.00} & \multicolumn{2}{c|}{98.00} & \multicolumn{2}{c|}{100.00} & \multicolumn{2}{c|}{100.00} & \multicolumn{2}{c|}{99.00} \\\midrule\midrule

\instructblips &  23.67 & 0.00 & 24.33 & 0.00 & 39.67 & 0.00 & 27.00 & 1.00 & 26.67 & 2.00 \\ \midrule

\instructblipl & 26.67 & 30.67 & 23.67 & 24.33 & 41.67 & 40.67 & 23.67 & 22.00 & 26.00 & 1.67\\ \midrule

\qwenvl & 30.67 & 9.00 & 36.67 & 9.00 & 47.00 & 17.67 & 27.33 & 15.00 & 25.33 & 16.00\\ \midrule

\qwenvlchat & 39.67 & 24.67 & 42.67 & 42.67 & 54.67 & 52.00 & 31.67 & 33.00 & 33.33 & 27.00 \\ \midrule

\llavas &42.00 & 40.67 & 40.00 & 45.67 & 48.67 & 49.67 & 31.00 & 39.00 & 28.00 & 34.33\\ \midrule

\llaval & 49.33 & 48.33 & 46.00 & 46.67 & 58.33 & 55.33 & 39.67 & 32.67 & 34.67 & 33.00 \\ \midrule

\glmfourv & 56.67 & 58.67 & 57.67 & 56.00 & 62.00 & 59.33 & 46.67 & 42.33 & 40.67 & 41.33 \\\midrule
\cogvlmtwo & 45.67 & 53.33 & 50.00 & 50.00 & 66.67 & 64.67 & 36.33 & 38.00 & 32.33 & 38.67 \\\midrule
\ideficstwo & 61.33 & 61.33 & 49.00 & 46.00 & 56.33 & 56.33 & 38.33 & 44.33 & 41.00 & 39.00 \\\midrule
\phivision & 60.00 & 63.33 & 50.00 & 51.33 & 66.67 & 69.67 & 47.00 & 49.33 & 44.33 & 44.33 \\\midrule
\paligemma & 48.67 & 47.33 & 48.67 & 53.00 & 59.00 & 64.00 & 49.33 & 54.00 & 40.67 & 44.00 \\\midrule

\internvlchat &57.67 & 60.67 & 62.00 & 55.00  & \textbf{75.33} & \textbf{72.33} & \textbf{51.33} & 49.33 & 46.67 & \textbf{49.67} \\ \midrule

\llavanext & \textbf{68.33} & \textbf{64.67} & \textbf{63.33} & \textbf{62.67} & \textbf{72.00} & 70.67 & \textbf{57.33} & \textbf{58.33} & \textbf{53.00} & \textbf{55.33} \\ \midrule\midrule

\geminipro & 33.33 & 34.33 & 32.67 & 38.00 & 32.33 & 33.00 & 26.67 & 28.33 & 28.00 & 23.67 \\ \midrule

\qwenvlmax & 58.33 & 45.00 & 57.00 & 59.67 & 71.33 & 68.33 & 48.33 & 47.33 & 42.33 & 46.33\\ \midrule

\gptfourv & 40.00  & \textbf{68.67} & 40.67 & 50.33 & 41.00 & 60.33 & 25.67 & 42.67 & 25.67 & 40.00 \\ \midrule

\gptfouro & 44.67 & 53.67 & 50.33 & \textbf{62.33} & 60.00 & 67.00 & 36.00 & 45.67 & 35.67 & 45.33 \\

\bottomrule 
\end{tabular}
   }
\label{tab:random-image-2d}
\end{table*}

\begin{table*}[!h]
  \centering
  \small
  \caption{random-\threed part 1}
  \scalebox{0.75}{
    \begin{tabular}{l| cc|cc|cc|cc|cc} 
    \toprule
     & \multicolumn{2}{c|}{\bf how many} & \multicolumn{2}{c|}{\bf what} & \multicolumn{2}{c|}{\bf what attribute} & \multicolumn{2}{c|}{\bf where} & \multicolumn{2}{c|}{\bf where attribute}\\ 

    \cmidrule(lr){2-3}\cmidrule(lr){4-5}\cmidrule(lr){6-7}\cmidrule(lr){8-9}\cmidrule(lr){10-11}

 & {\bf DP} &  {\bf SP} & {\bf DP} &  {\bf SP}  & {\bf DP} &  {\bf SP}  & {\bf DP} &  {\bf SP}  & {\bf DP} &  {\bf SP} \\\midrule\midrule

 \textbf{Human} & \multicolumn{2}{c|}{99.00} & \multicolumn{2}{c|}{100.00} & \multicolumn{2}{c|}{100.00} & \multicolumn{2}{c|}{99.00} & \multicolumn{2}{c|}{100.00} \\\midrule\midrule

\instructblips & 32.67 & 0.00 & 28.00 & 0.00 & 45.00 & 0.00 & 25.67 & 1.00 & 27.00 & 2.33\\ \midrule

\instructblipl & 32.00 & 32.33 & 22.67 & 23.33 & 42.67 & 0.00 & 28.67 & 25.33 & 23.00 & 24.67\\ \midrule

\qwenvl & 32.33 & 11.00 & 28.00 & 8.67 & 50.67 & 19.67 & 22.67 & 18.33 & 24.67 & 15.00 \\ \midrule

\qwenvlchat & 45.00 & 33.33 & 32.33 & 33.33 & 55.00 & 57.00 & 21.67 & 24.00 & 29.67 & 32.33 \\ \midrule

\llavas & 38.67 & 39.33 & 32.67 & 40.33 & 57.00 & 54.00 & 27.00 & 27.67 & 26.00 & 26.00 \\ \midrule

\llaval & 46.67 & 48.33 & 40.67 & 41.00 & 60.33 & 56.00 & 34.33 & 32.67 & 36.00 & 32.67 \\ \midrule

\glmfourv & \textbf{74.00} & \textbf{73.00} & 55.33 & 47.00 & 67.67 & 65.67 & 40.67 & 38.33 & 33.33 & 36.00 \\\midrule
\cogvlmtwo & 60.67 & 62.67 & 39.67 & 42.67 & 57.67 & 58.33 & 31.67 & 34.00 & 29.67 & 32.00 \\\midrule
\ideficstwo & 65.00 & 65.67 & 41.00 & 39.33 & 64.33 & 58.33 & 36.67 & 38.33 & 39.00 & 37.67 \\\midrule
\phivision & 59.33 & 60.33 & 37.67 & 40.33 & 62.00 & 62.00 & 34.33 & 36.67 & 36.67 & 38.00 \\\midrule
\paligemma & 49.00 & 45.33 & 40.67 & 43.67 & 63.67 & 68.00 & 34.67 & 39.67 & 35.67 & 35.67 \\\midrule\midrule

\internvlchat & \textbf{67.00} & \textbf{67.00} & \textbf{60.33} & \textbf{56.33} & 68.33 & 65.67 & \textbf{54.67} & \textbf{55.67} & \textbf{46.67} & \textbf{46.00}\\ \midrule

\llavanext & 63.67 & 63.33 & 49.67 & 50.67 & \textbf{71.33} & \textbf{71.33} & 48.33 & \textbf{51.00} & 40.33 & \textbf{49.00} \\ \midrule\midrule

\geminipro & 40.00 & 38.67 & 32.67 & 25.00 & 31.33 & 34.67 & 28.00 & 31.00 & 27.67 & 28.00 \\ \midrule

\qwenvlmax & 65.00 & 60.67 & 54.67 & 55.33 & 63.67 & 61.33 & 42.33 & 44.00 & 32.67 & 37.33 \\ \midrule

\gptfourv & 41.67 & 66.67 & 31.67 & 37.67 & 41.33 & 54.67 & 25.00 & 39.00 & 25.67 & 28.33 \\ \midrule

\gptfouro & 45.00 & 64.33 & 47.33 & \textbf{58.67} & 57.33 & \textbf{68.67} & 37.67 & 45.33 & 30.67 & 44.33 \\

\bottomrule 
\end{tabular}
}
\label{tab:random-image-3d1}
\end{table*}

\begin{table*}[!h]
  \centering
  \small
  \caption{random-\threed part 2}
  \scalebox{0.7}{
    \begin{tabular}{l| cc|cc|cc|cc|cc|cc} 
    \toprule
     & \multicolumn{2}{c|}{\bf what distance} & \multicolumn{2}{c|}{\bf where distance} & \multicolumn{2}{c|}{\bf what attribute distance} & \multicolumn{2}{c|}{\bf what size}  & \multicolumn{2}{c|}{\bf where size} & \multicolumn{2}{c|}{\bf what attribute size}\\ 

    \cmidrule(lr){2-3}\cmidrule(lr){4-5}\cmidrule(lr){6-7}\cmidrule(lr){8-9}\cmidrule(lr){10-11}\cmidrule(lr){12-13}

 & {\bf DP} &  {\bf SP} & {\bf DP} &  {\bf SP}  & {\bf DP} &  {\bf SP}  & {\bf DP} &  {\bf SP}  & {\bf DP} &  {\bf SP}  & {\bf DP} &  {\bf SP} \\\midrule\midrule

 \textbf{Human} & \multicolumn{2}{c|}{100.00} & \multicolumn{2}{c|}{99.00} & \multicolumn{2}{c|}{100.00} & \multicolumn{2}{c|}{100.00} & \multicolumn{2}{c|}{100.00} & \multicolumn{2}{c|}{100.00}\\\midrule\midrule

\instructblips  & 17.67 & 0.00 & 38.33 & 0.00 & 51.00 & 0.00 & 30.33 & 0.00 & 32.33 & 1.67 & 51.33 & 0.00 \\ \midrule

\instructblipl   & 23.67 & 24.33 & 29.33 & 29.00 & 48.00 & 1.67 & 35.67 & 37.00 & 25.33 & 24.00 & 53.33 & 50.33 \\ \midrule

\qwenvl   & 25.33 & 8.67 & 26.33 & 14.00 & 50.33 & 19.67 & 34.67 & 14.00 & 21.33 & 19.00 & 52.00 & 27.00 \\ \midrule

\qwenvlchat  & 25.00 & 24.00 & 25.67 & 28.33 & 56.67 & 56.00 & 43.00 & 48.67 & 31.00 & 30.67 & 62.67 & 65.33 \\ \midrule

\llavas  & 28.00 & 30.67 & 26.33 & 25.67 & 49.67 & 48.67 & 43.00 & 44.67 & 29.33 & 34.67 & 55.33 & 60.00 \\ \midrule

\llaval   & 33.67 & 29.33 & 26.00 & 23.67 & 57.67 & 55.33 & 48.33 & 48.33 & 34.67 & 35.67 & 65.33 & 63.33 \\ \midrule

\glmfourv & 45.67 & 46.00 & 18.33 & 25.00 & 53.33 & 57.67 & 72.67 & \textbf{73.67} & 49.67 & 45.33 & \textbf{82.33} & \textbf{83.00} \\\midrule
\cogvlmtwo & \textbf{53.00} & \textbf{55.67} & 26.33 & 35.67 & 63.00 & 63.67 & 70.67 & \textbf{76.33} & 37.67 & 40.33 & 79.33 & 81.67 \\\midrule
\ideficstwo & 35.67 & 36.00 & 26.00 & 30.67 & 65.00 & 61.00 & 62.00 & 63.33 & 45.33 & 44.67 & 77.33 & 75.00 \\\midrule
\phivision & 36.33 & 33.67 & 23.00 & 25.00 & 61.00 & 62.00 & 52.00 & 53.33 & 44.33 & 43.67 & 70.33 & 72.00 \\\midrule
\paligemma & 20.67 & 26.33 & 29.00 & 23.67 & 50.67 & 54.00 & 48.33 & 49.33 & 35.00 & 41.67 & 74.33 & 72.33 \\\midrule\midrule

\internvlchat & 52.33 & 36.00 & 39.00 & \textbf{47.00} & \textbf{69.67} & 68.67 & 73.33 & \textbf{73.67} & \textbf{57.67} & \textbf{57.67} & \textbf{82.67} & \textbf{82.33} \\ \midrule

\llavanext  & 48.00 & 45.33 & 34.33 & \textbf{40.67} & \textbf{75.00} & \textbf{74.00} & 62.33 & 62.00 & 49.00 & \textbf{52.67} & 77.67 & 78.67 \\ \midrule\midrule

\geminipro   & 39.33 & 31.00 & 25.33 & 24.33 & 38.33 & 36.00 & 34.33 & 29.67 & 26.67 & 26.67 & 39.67 & 37.00\\ \midrule

\qwenvlmax & 39.00 & \textbf{53.00} & 2.67 & 35.67 & 65.00 & 66.67 & 72.33 & 69.67 & 45.33 & 50.00 & 75.67 & 72.00 \\ \midrule

\gptfourv  & 39.33 & 46.67 & 21.67 & 19.00 & 43.33 & 64.33 & 46.00 & 54.00 & 22.33 & 37.67 & 66.00 & 75.00 \\ \midrule

\gptfouro & 44.67 & \textbf{62.33} & 24.00 & \textbf{41.67} & 58.33 & 65.33 & 57.67 & 73.00 & 32.33 & 44.67 & 71.00 & 76.33\\

\bottomrule 
\end{tabular}
}
\label{tab:random-image-3d2}
\end{table*}

\begin{table*}[!h]
  \centering
  \small
  \caption{random-Real images with \sg}
  \scalebox{0.95}{
    \begin{tabular}{l |cc|cc|cc} 
    \toprule

     &  \multicolumn{2}{c|}{\bf what attribute} & \multicolumn{2}{c|}{\bf what object} & \multicolumn{2}{c|}{\bf what relation}\\ 

     \cmidrule(lr){2-3}\cmidrule(lr){4-5}\cmidrule(lr){6-7}

 & {\bf DP} &  {\bf SP} & {\bf DP} &  {\bf SP}  & {\bf DP} &  {\bf SP}  \\\midrule\midrule

\textbf{Human} & \multicolumn{2}{c|}{96.00} & \multicolumn{2}{c|}{99.00} & \multicolumn{2}{c|}{97.00}\\\midrule\midrule

\instructblips  & 65.67 & 0.00 & 79.00 & 0.00 & 60.33 & 0.33 \\ \midrule

\instructblipl   & 66.33 & 68.67 & 84.33 & 80.00 & 45.00 & 49.67\\ \midrule

\qwenvl   & 64.00 & 4.33 & 83.33 & 8.67 & 59.00 & 24.67 \\ \midrule

\qwenvlchat  & 69.67 & 69.00 & 87.00 & 86.67 & 78.33 & 82.67 \\ \midrule

\llavas  & 70.00 & 65.33 & 85.00 & 84.33 & 31.00 & 76.00 \\ \midrule

\llaval   & 72.67 & 70.33 & 90.00 & 90.00 & 75.00 & \textbf{88.00}\\ \midrule

\glmfourv & 74.33 & 72.00 & 88.67 & 88.00 & 62.00 & 51.67 \\\midrule
\cogvlmtwo & 70.00 & 71.33 & 92.67 & 93.33 & 51.33 & 47.33 \\\midrule
\ideficstwo & 69.67 & 68.67 & 86.33 & 85.67 & 67.00 & 68.00 \\\midrule
\phivision & 77.67 & 76.00 & 92.00 & 93.67 & 59.00 & 62.00 \\\midrule
\paligemma & 75.33 & 76.00 & 94.00 & 93.33 & 70.67 & 74.33 \\\midrule\midrule

\internvlchat & \textbf{80.00} & 77.33 & \textbf{94.67} & 92.00 & 79.33 & 77.67 \\ \midrule

\llavanext  & \textbf{78.33} & 75.33 & 93.33 & \textbf{95.33} & 85.33 & 84.00 \\ \midrule\midrule

\geminipro   & 51.00 & 50.67 & 71.00 & 68.67 & 48.33 & 63.33\\ \midrule

\qwenvlmax & 76.67 & \textbf{81.33} & 93.67 & \textbf{96.00} & \textbf{86.67} & \textbf{90.67} \\ \midrule

\gptfourv  & 69.33 & 67.00 & 82.67 & 79.33 & 68.33 & 69.00 \\ \midrule

\gptfouro  & 68.00 & 67.67 & 83.00 & 81.67 & 78.00 & 82.67 \\

\bottomrule 
\end{tabular}
}
\label{tab:random-image-sg}
\end{table*}

\clearpage

\subsection{A breakdown of Table~\ref{tab:random-video}}
\label{app:random-video}

\begin{table*}[!h]
  \centering
  \small
  \caption{random-\threed}
  \scalebox{0.7}{
    \begin{tabular}{l| cc|cc|cc|cc|cc|cc} 
    \toprule
     &  \multicolumn{2}{c}{\bf what attribute move} & \multicolumn{2}{c}{\bf what attribute rotate} & \multicolumn{2}{c}{\bf what move} & \multicolumn{2}{c}{\bf what rotate} & \multicolumn{2}{c}{\bf where move} & \multicolumn{2}{c}{\bf where rotate}\\ 

\cmidrule(lr){2-3}\cmidrule(lr){4-5}\cmidrule(lr){6-7}\cmidrule(lr){8-9}\cmidrule(lr){10-11}\cmidrule(lr){12-13}

 & {\bf DP} &  {\bf SP} & {\bf DP} &  {\bf SP}  & {\bf DP} &  {\bf SP}   & {\bf DP} &  {\bf SP} & {\bf DP} &  {\bf SP}  & {\bf DP} &  {\bf SP} \\\midrule\midrule

 \textbf{Human} & \multicolumn{2}{c|}{100.00} & \multicolumn{2}{c|}{100.00} & \multicolumn{2}{c|}{98.00} & \multicolumn{2}{c|}{92.00} & \multicolumn{2}{c|}{100.00} & \multicolumn{2}{c|}{100.00}\\\midrule\midrule     

\videochatgpts &  27.00 & 24.33 & 27.00 & 28.33 & 18.33 & 19.00 & 15.67 & 18.67 & 27.33 & 26.33 & 13.33 & 11.67 \\ \midrule

\videollavas &  28.33 & 54.00 & 25.00 & 49.33 & 26.00 & \textbf{34.00} & 26.67 & \textbf{35.33}& 25.00 & 31.33 & 25.00 & 28.67 \\ \midrule

\videochattwos  & 46.67 & 48.33 & 41.33 & 47.67 & 29.00 & 22.33 & 27.67 & 19.67 & 17.00 & 14.00 & 22.00 & 19.33\\ \midrule

\videollamatwos &  28.67 & 24.00 & 27.67 & 25.00 & 22.33 & 19.00 & 23.33 & 16.00 & 20.00 & 7.33 & 20.67 & 6.67 \\ \midrule

\videollamatwol  & 29.67 & 26.67 & 32.33 & 32.00 & 18.33 & 17.67 & 19.33 & 17.67 & 17.67 & 14.67 & 18.67 & 12.67 \\ \midrule

\chatunivis & 36.67 & 27.67 & 35.33 & 39.67 & 27.67 & 20.33 & 28.33 & 24.00 & 25.67 & 24.00 & 24.67 & 20.00\\ \midrule

\chatunivil & 33.67 & 31.33 & 33.67 & 37.00 & 24.33 & 22.67 & 29.33 & 28.00 & 25.33 & 16.33 & 22.67 & 18.67\\ \midrule\midrule

\internvlchat & 52.33 & 43.00 & 56.00 & 49.33 & 26.67 & 21.00 & 31.33 & 22.67 & 31.67 & 28.00 & 32.00 & 26.00\\ \midrule

\llavanext & \textbf{57.67} & \textbf{56.67} & 59.00 & \textbf{62.67} & 28.00 & 29.33 & 30.67 & 29.67 & \textbf{32.33}& \textbf{32.33}& \textbf{32.67}& \textbf{36.33}\\ \midrule\midrule

\geminipro & 39.33 & 38.67 & 40.33 & 37.67 & \textbf{30.67}& 28.67 & 27.33 & 25.33 & 27.67 & 29.67 & 25.33 & 20.67\\ \midrule

 \qwenvlmax &  \textbf{56.33} & 52.67 & \textbf{67.33} & \textbf{67.00} & 29.00 & 30.00 & 34.00 & \textbf{35.33}& 26.00 & 25.00 & 20.67 & 26.33\\ \midrule

 \gptfourv & 43.67 & 51.00 & 46.67 & 57.33 & 28.00 & 29.33 & 29.67 & 32.00& 22.00 & 26.00 & 15.67 & 25.33\\ \midrule

  \gptfouro & 47.67 & 46.00 & 54.67 & \textbf{62.67} & 27.33 & \textbf{31.00} & 34.33 & \textbf{38.67}& 27.00 & \textbf{36.33}& 23.00 & \textbf{35.67}\\ 

\bottomrule 
\end{tabular}
}
  
  \label{tab:random-video-3d}
\end{table*}

\begin{table*}[h]
  \centering
  \small
  \caption{random-Real videos with \sg}
  \scalebox{0.95}{
    \begin{tabular}{l| cc|cc|cc} 
    \toprule
     &  \multicolumn{2}{c}{\bf what action} & \multicolumn{2}{c}{\bf what object} & \multicolumn{2}{c}{\bf what relation}\\ 

     \cmidrule(lr){2-3}\cmidrule(lr){4-5}\cmidrule(lr){6-7}

 & {\bf DP} &  {\bf SP} & {\bf DP} &  {\bf SP}  & {\bf DP} &  {\bf SP}  \\\midrule\midrule

 \textbf{Human} & \multicolumn{2}{c|}{100.00} & \multicolumn{2}{c|}{98.00} & \multicolumn{2}{c|}{100.00} \\\midrule\midrule

\videochatgpts &  19.67 & 16.33 & 37.00 & 29.67 & 34.67 & 31.00 \\ \midrule

\videollavas & 29.67 & 58.33 & 31.33 & 62.67 & 35.33 & 49.00 \\ \midrule

\videochattwos  & 36.33 & 26.33 & 44.33 & 42.67 & 33.00 & 29.67 \\ \midrule

\videollamatwos & 33.67 & 21.33 & 37.67 & 40.00 & 37.67 & 33.67 \\ \midrule

\videollamatwol  & 30.33 & 23.67 & 39.00 & 36.00 & 23.00 & 25.67 \\ \midrule

\chatunivis & 44.67 & 37.67 & 57.33 & 47.67 & 48.33 & 49.67 \\ \midrule

\chatunivil & 38.33 & 25.00 & 58.67 & 52.00 & 38.67 & 42.67 \\ \midrule\midrule

\internvlchat &\textbf{72.33} & 52.33 & \textbf{73.00} & 54.33 & 59.00 & 62.33\\ \midrule

\llavanext & 67.00 & 60.00 & 67.33 & 65.33 & 68.33 & 65.00 \\ \midrule\midrule

\geminipro & 54.33 & 39.67 & 55.00 & 53.00 & 40.67 & 44.67 \\ \midrule

\qwenvlmax & \textbf{67.33} & \textbf{68.67} & \textbf{69.67} & \textbf{68.00} & 70.33 & 63.67 \\ \midrule 

\gptfourv & 53.67 & 56.67 & 57.67 & 58.67 & 66.00 & \textbf{72.67} \\ \midrule 

\gptfouro & 64.67 & 62.33 & 66.00 & 60.00 & \textbf{78.00} & \textbf{76.33} \\ 

\bottomrule 
\end{tabular}
}
  
  \label{tab:random-video-sg}

\end{table*}

\clearpage

\section{Details of Model Performance on TaskMeAnything 2024 benchmark}

In this section, we present the full results of our evaluation on \benchname with 18 MLMs.

\subsection{Raw results of Figure~\ref{fig:2024-result}}

\begin{table*}[!h]
  \centering
  \small
  \caption{\textbf{2024-ImageQA}. The model performance on 2024 subsets of ImageQA tasks using both the detailed prompt and the succinct prompt. Numbers in parentheses are the number of task instances for each set.}
  \scalebox{0.65}{
    \begin{tabular}{l| cc|cc|cc} 
    \toprule
     & \multicolumn{2}{c|}{\bf \twod} &  \multicolumn{2}{c|}{\bf \threed} & \multicolumn{2}{c}{\bf \sg}\\ 
     
     & \multicolumn{2}{c|}{(3,279)} & \multicolumn{2}{c|}{(7,095)} & \multicolumn{2}{c}{(1,896)}\\ \cmidrule(lr){2-3}\cmidrule(lr){4-5}\cmidrule(lr){6-7}

 & {\bf Detailed prompt} &  {\bf Succinct prompt} & {\bf Detailed prompt} &  {\bf Succinct prompt}  & {\bf Detailed prompt} &  {\bf Succinct prompt}\\\midrule\midrule

\instructblips & 22.92 & 1.12 & 24.94 & 0.42 & 43.23 & 0.38 \\ \midrule
\instructblipl & 22.20 & 0.72 & 21.55 & 0.45 & 45.35 & 0.86 \\ \midrule
\qwenvl & 23.48 & 12.02 & 23.77 & 12.36 & 42.68 & 11.26 \\ \midrule
\qwenvlchat & 25.53 & 24.55 & 24.21 & 25.72 & 53.04 & 52.38 \\ \midrule
\llavas & 25.83 & 26.09 & 23.64 & 23.01 & 42.67 & 46.75 \\ \midrule
\llaval & 27.03 & 24.85 & 26.25 & 25.78 & 51.01 & 49.10 \\ \midrule
\glmfourv & 32.76 & 33.80 & 37.09 & 37.33 & 47.80 & 44.22 \\ \midrule
\cogvlmtwo & 29.99 & 31.11 & 37.29 & 39.66 & 49.46 & 48.25 \\ \midrule
\ideficstwo & 30.50 & 32.20 & 35.32 & 34.80 & 44.40 & 46.61 \\ \midrule
\phivision & 30.44 & 32.95 & 27.88 & 29.52 & 46.57 & 49.46 \\ \midrule
\paligemma & 32.14 & 33.77 & 28.95 & 29.42 & 58.45 & 58.71 \\ \midrule\midrule
\internvlchat & 36.78 & 37.08 & 43.53 & 42.55 & 57.08 & 52.01 \\ \midrule\midrule
\gptfouro & 31.46 & 40.87 & 31.86 & 47.46 & 52.03 & 53.87 \\
\bottomrule 
\end{tabular}
}
  
  \label{tab:2024-image}

\end{table*}
\begin{table*}[h]
  \centering
  \small
  \caption{\textbf{2024-VideoQA}. The model performance on 2024 subsets of VideoQA tasks using both the detailed prompt and the succinct prompt. Numbers in parentheses are the number of task instances for each set.}
  \scalebox{0.8}{
    \begin{tabular}{l| cc|cc} 
    \toprule
     &  \multicolumn{2}{c|}{\bf \threed} & \multicolumn{2}{c}{\bf \sg}\\ 
     
     &  \multicolumn{2}{c|}{(2,394)} & \multicolumn{2}{c}{(1,173)}\\ \cmidrule(lr){2-3}\cmidrule(lr){4-5}

 & {\bf Detailed prompt} &  {\bf Succinct prompt} & {\bf Detailed prompt} &  {\bf Succinct prompt}  \\\midrule\midrule

\videochatgpts & 12.86 & 10.87 & 15.56 & 13.87 \\ \midrule
\videollavas & 19.59 & 22.14 & 21.68 & 34.54 \\ \midrule
\videochattwos & 21.69 & 17.42 & 30.31 & 19.01 \\ \midrule
\videollamatwos & 18.79 & 10.59 & 27.32 & 20.36 \\ \midrule
\videollamatwol & 17.31 & 12.34 & 23.17 & 15.10 \\ \midrule
\chatunivis & 16.48 & 14.47 & 36.44 & 26.66 \\ \midrule
\chatunivil & 17.84 & 15.36 & 27.30 & 20.60 \\ \midrule\midrule
\internvlchat & 23.67 & 23.58 & 54.04 & 38.02 \\ \midrule
\gptfouro & 26.96 & 34.53 & 57.88 & 58.23 \\
\bottomrule 
\end{tabular}
}
  
  \label{tab:2024-video}

\end{table*}

\clearpage

\subsection{A breakdown of Table~\ref{tab:2024-image}}
\label{app:2024-image}
\begin{table*}[!h]
  \centering
  \small
  \caption{2024-\twod}
  \scalebox{0.75}{
    \begin{tabular}{l| cc|cc|cc|cc|cc} 
    \toprule
     & \multicolumn{2}{c|}{\bf how many} & \multicolumn{2}{c|}{\bf what} & \multicolumn{2}{c|}{\bf what attribute} & \multicolumn{2}{c|}{\bf where} & \multicolumn{2}{c}{\bf where attribute}\\ 
    \cmidrule(lr){2-3}\cmidrule(lr){4-5}\cmidrule(lr){6-7}\cmidrule(lr){8-9}\cmidrule(lr){10-11}
 & {\bf DP} &  {\bf SP} & {\bf DP} &  {\bf SP}  & {\bf DP} &  {\bf SP}  & {\bf DP} &  {\bf SP}  & {\bf DP} &  {\bf SP} \\\midrule\midrule
\instructblips & 25.25 & 0.00 & 16.74 & 0.00 & 28.77 & 0.00 & 22.75 & 2.15 & 21.07 & 3.46 \\\midrule
\instructblipl & 22.11 & 0.00 & 14.83 & 0.00 & 33.33 & 0.00 & 20.60 & 1.57 & 20.13 & 2.04 \\\midrule
\qwenvl & 23.27 & 4.13 & 21.15 & 7.64 & 35.16 & 11.42 & 19.74 & 18.03 & 18.08 & 18.87 \\\midrule
\qwenvlchat & 28.88 & 21.12 & 23.49 & 24.52 & 36.53 & 35.31 & 21.32 & 24.03 & 17.45 & 17.77 \\\midrule
\llavas & 29.04 & 31.19 & 19.53 & 23.20 & 31.20 & 29.07 & 26.90 & 25.75 & 22.48 & 21.23 \\\midrule
\llaval & 30.86 & 34.32 & 21.73 & 21.44 & 40.03 & 34.40 & 24.46 & 19.46 & 18.08 & 14.62 \\\midrule
\glmfourv & 43.40 & 45.05 & 29.37 & 31.28 & 42.31 & 44.44 & 27.18 & 27.18 & 21.54 & 21.07 \\\midrule
\cogvlmtwo & 41.09 & 43.23 & 26.87 & 27.61 & 42.16 & 44.75 & 23.03 & 23.61 & 16.82 & 16.35 \\\midrule
\ideficstwo & 42.90 & 41.91 & 24.96 & 26.28 & 41.70 & 39.88 & 25.04 & 29.33 & 17.92 & 23.58 \\\midrule
\phivision & 38.94 & 40.26 & 27.61 & 26.28 & 40.33 & 44.14 & 23.32 & 27.18 & 22.01 & 26.89 \\\midrule
\paligemma & 33.66 & 33.83 & 22.61 & 24.96 & 38.81 & 45.36 & 37.34 & 37.48 & 28.30 & 27.20 \\\midrule
\internvlchat & 38.78 & 47.85 & 34.95 & 30.40 & 48.86 & 40.33 & 33.48 & 34.91 & 27.83 & 31.92 \\\midrule
\gptfouro & 36.80 & 48.35 & 31.57 & 41.56 & 37.29 & 48.25 & 28.04 & 36.77 & 23.58 & 29.40 \\
\bottomrule 
    \end{tabular}
   }
\label{tab:2024-image-2d}
\end{table*}

\begin{table*}[!h]
  \centering
  \small
  \caption{2024-\threed part 1}
  \scalebox{0.75}{
    \begin{tabular}{l| cc|cc|cc|cc|cc} 
    \toprule
     & \multicolumn{2}{c|}{\bf how many} & \multicolumn{2}{c|}{\bf what} & \multicolumn{2}{c|}{\bf what attribute} & \multicolumn{2}{c|}{\bf where} & \multicolumn{2}{c}{\bf where attribute}\\ 
    \cmidrule(lr){2-3}\cmidrule(lr){4-5}\cmidrule(lr){6-7}\cmidrule(lr){8-9}\cmidrule(lr){10-11}
 & {\bf DP} &  {\bf SP} & {\bf DP} &  {\bf SP}  & {\bf DP} &  {\bf SP}  & {\bf DP} &  {\bf SP}  & {\bf DP} &  {\bf SP} \\\midrule\midrule
\instructblips & 26.30 & 0.00 & 17.36 & 0.00 & 28.79 & 0.00 & 26.31 & 1.05 & 25.45 & 1.81 \\\midrule
\instructblipl & 20.49 & 0.00 & 15.35 & 0.16 & 29.89 & 0.00 & 21.23 & 0.75 & 21.35 & 1.15 \\\midrule
\qwenvl & 30.28 & 6.88 & 20.00 & 5.74 & 33.49 & 14.55 & 20.33 & 14.05 & 20.85 & 11.82 \\\midrule
\qwenvlchat & 34.86 & 32.72 & 19.84 & 20.47 & 36.46 & 35.84 & 18.24 & 18.68 & 16.09 & 20.53 \\\midrule
\llavas & 21.87 & 21.87 & 21.24 & 24.65 & 29.58 & 28.17 & 24.96 & 20.48 & 24.96 & 19.87 \\\midrule
\llaval & 29.66 & 34.25 & 23.88 & 24.50 & 34.90 & 32.55 & 22.87 & 19.43 & 21.02 & 19.54 \\\midrule
\glmfourv & 48.93 & 48.78 & 39.69 & 38.45 & 42.88 & 43.51 & 24.51 & 22.12 & 19.54 & 19.38 \\\midrule
\cogvlmtwo & 37.77 & 44.50 & 29.77 & 30.23 & 44.29 & 46.17 & 21.23 & 24.07 & 21.02 & 20.36 \\\midrule
\ideficstwo & 42.05 & 41.13 & 25.89 & 24.50 & 40.85 & 38.97 & 29.75 & 28.85 & 27.09 & 29.06 \\\midrule
\phivision & 40.98 & 42.05 & 25.89 & 24.65 & 37.25 & 38.50 & 20.03 & 20.63 & 18.88 & 23.65 \\\midrule
\paligemma & 37.46 & 30.58 & 22.95 & 26.36 & 38.03 & 42.25 & 33.78 & 32.14 & 26.27 & 26.44 \\\midrule
\internvlchat & 43.12 & 49.08 & 39.84 & 38.14 & 50.23 & 45.54 & 34.08 & 33.03 & 33.17 & 26.93 \\\midrule
\gptfouro & 32.26 & 54.43 & 27.29 & 40.93 & 36.62 & 52.27 & 22.12 & 32.59 & 19.38 & 26.93 \\
\bottomrule 
    \end{tabular}
   }
\label{tab:2024-image-3d1}
\end{table*}

\begin{table*}[!h]
  \centering
  \small
  \caption{2024-\threed part 2}
  \scalebox{0.7}{
    \begin{tabular}{l| cc|cc|cc|cc|cc|cc} 
    \toprule
     & \multicolumn{2}{c|}{\bf what distance} & \multicolumn{2}{c|}{\bf where distance} & \multicolumn{2}{c|}{\bf what attribute distance} & \multicolumn{2}{c|}{\bf what size}  & \multicolumn{2}{c|}{\bf where size} & \multicolumn{2}{c}{\bf what attribute size}\\ 
    \cmidrule(lr){2-3}\cmidrule(lr){4-5}\cmidrule(lr){6-7}\cmidrule(lr){8-9}\cmidrule(lr){10-11}\cmidrule(lr){12-13}
 & {\bf DP} &  {\bf SP} & {\bf DP} &  {\bf SP}  & {\bf DP} &  {\bf SP}  & {\bf DP} &  {\bf SP}  & {\bf DP} &  {\bf SP}  & {\bf DP} &  {\bf SP} \\\midrule\midrule
\instructblips & 17.26 & 0.00 & 31.94 & 0.00 & 31.71 & 0.00 & 13.04 & 0.00 & 27.99 & 1.78 & 28.17 & 0.00 \\\midrule
\instructblipl & 16.92 & 0.17 & 24.50 & 0.62 & 29.20 & 0.00 & 9.33 & 0.15 & 20.06 & 1.94 & 28.76 & 0.00 \\\midrule
\qwenvl & 17.78 & 8.03 & 23.88 & 16.59 & 28.17 & 11.50 & 16.59 & 11.41 & 21.68 & 17.80 & 28.47 & 17.55 \\\midrule
\qwenvlchat & 19.66 & 18.63 & 17.36 & 21.09 & 31.86 & 31.12 & 20.30 & 22.22 & 19.90 & 26.21 & 31.71 & 35.40 \\\midrule
\llavas & 22.91 & 23.42 & 22.17 & 22.17 & 26.11 & 25.22 & 17.04 & 19.85 & 22.98 & 24.60 & 26.25 & 22.86 \\\midrule
\llaval & 23.93 & 25.64 & 19.38 & 17.05 & 32.60 & 30.24 & 23.85 & 24.59 & 21.84 & 25.24 & 34.81 & 30.53 \\\midrule

\glmfourv & 35.90 & 36.75 & 25.89 & 25.27 & 34.96 & 40.27 & 55.26 & 55.26 & 32.36 & 31.39 & 48.08 & 49.41 \\\midrule
\cogvlmtwo & 42.39 & 44.79 & 26.51 & 32.25 & 50.29 & 52.51 & 54.52 & 54.07 & 29.77 & 32.20 & 52.65 & 55.16 \\\midrule
\ideficstwo & 31.79 & 30.26 & 22.79 & 24.50 & 42.18 & 43.81 & 34.96 & 33.48 & 37.22 & 36.57 & 53.98 & 51.62 \\\midrule
\phivision & 30.43 & 28.72 & 15.66 & 19.38 & 31.27 & 34.22 & 28.59 & 28.74 & 27.35 & 30.10 & 30.38 & 34.07 \\\midrule
\paligemma & 17.44 & 21.37 & 25.27 & 18.29 & 30.38 & 31.56 & 16.00 & 19.70 & 31.88 & 33.33 & 38.94 & 41.59 \\\midrule
\internvlchat & 44.96 & 26.84 & 33.95 & 41.86 & 46.76 & 48.53 & 50.37 & 51.85 & 43.85 & 43.53 & 58.55 & 62.68 \\\midrule
\gptfouro & 38.97 & 57.61 & 19.84 & 38.91 & 42.04 & 54.72 & 32.59 & 54.07 & 28.80 & 43.04 & 50.59 & 66.52 \\
\bottomrule 
    \end{tabular}
   }
\label{tab:2024-image-3d2}
\end{table*}

\begin{table*}[!h]
  \centering
  \small
  \caption{2024-Real images with \sg}
  \scalebox{0.95}{
    \begin{tabular}{l |cc|cc|cc} 
    \toprule
     &  \multicolumn{2}{c|}{\bf what attribute} & \multicolumn{2}{c|}{\bf what object} & \multicolumn{2}{c}{\bf what relation}\\ 
    \cmidrule(lr){2-3}\cmidrule(lr){4-5}\cmidrule(lr){6-7}
 & {\bf DP} &  {\bf SP} & {\bf DP} &  {\bf SP}  & {\bf DP} &  {\bf SP}  \\\midrule\midrule
\instructblips & 42.48 & 0.00 & 39.97 & 0.32 & 47.25 & 0.81 \\\midrule
\instructblipl & 46.98 & 0.47 & 50.55 & 0.16 & 38.51 & 1.94 \\\midrule
\qwenvl & 43.10 & 1.24 & 46.76 & 6.48 & 38.19 & 26.05 \\\midrule
\qwenvlchat & 50.54 & 45.27 & 47.08 & 50.87 & 61.49 & 61.00 \\\midrule
\llavas & 47.44 & 36.59 & 54.03 & 48.97 & 26.54 & 54.69 \\\midrule
\llaval & 41.71 & 34.26 & 54.19 & 49.61 & 57.12 & 63.43 \\\midrule
\glmfourv & 42.17 & 40.78 & 56.56 & 54.03 & 44.66 & 37.86 \\\midrule
\cogvlmtwo & 45.12 & 46.51 & 65.72 & 65.40 & 37.54 & 32.85 \\\midrule
\ideficstwo & 38.29 & 36.12 & 48.97 & 53.55 & 45.95 & 50.16 \\\midrule
\phivision & 45.58 & 44.03 & 54.98 & 60.98 & 39.16 & 43.37 \\\midrule
\paligemma & 48.84 & 49.15 & 74.57 & 74.88 & 51.94 & 52.10 \\\midrule
\internvlchat & 47.75 & 42.64 & 65.24 & 55.29 & 58.25 & 58.09 \\\midrule
\gptfouro & 39.69 & 41.86 & 51.50 & 52.76 & 64.89 & 66.99  \\
\bottomrule 
    \end{tabular}
   }
\label{tab:2024-image-sg}
\end{table*}
\clearpage

\subsection{A breakdown of Table~\ref{tab:2024-video}}
\label{app:2024-video}
\begin{table*}[!h]
  \centering
  \small
  \caption{2024-\threed}
  \scalebox{0.7}{
    \begin{tabular}{l| cc|cc|cc|cc|cc|cc} 
    \toprule
     &  \multicolumn{2}{c}{\bf what attribute move} & \multicolumn{2}{c|}{\bf what attribute rotate} & \multicolumn{2}{c|}{\bf what move} & \multicolumn{2}{c|}{\bf what rotate} & \multicolumn{2}{c|}{\bf where move} & \multicolumn{2}{c}{\bf where rotate}\\ 
    \cmidrule(lr){2-3}\cmidrule(lr){4-5}\cmidrule(lr){6-7}\cmidrule(lr){8-9}\cmidrule(lr){10-11}\cmidrule(lr){12-13}
 & {\bf DP} &  {\bf SP} & {\bf DP} &  {\bf SP}  & {\bf DP} &  {\bf SP}   & {\bf DP} &  {\bf SP} & {\bf DP} &  {\bf SP}  & {\bf DP} &  {\bf SP} \\\midrule\midrule
\videochatgpts & 21.54 & 13.33 & 10.12 & 7.65 & 5.88 & 7.11 & 12.41 & 12.17 & 21.39 & 19.40 & 5.82 & 5.56 \\\midrule
\videollavas & 24.36 & 32.05 & 20.49 & 22.22 & 17.65 & 18.38 & 18.25 & 19.22 & 16.92 & 17.66 & 19.84 & 23.28 \\\midrule
\videollamatwos & 22.31 & 15.38 & 24.20 & 12.84 & 13.73 & 11.52 & 13.14 & 9.25 & 17.41 & 8.46 & 21.96 & 6.08 \\\midrule
\videollamatwol & 26.41 & 15.90 & 21.98 & 15.80 & 13.97 & 9.56 & 13.14 & 12.17 & 11.69 & 8.46 & 16.67 & 12.17 \\\midrule
\chatunivis & 16.67 & 14.62 & 16.54 & 14.07 & 13.73 & 15.93 & 17.03 & 15.33 & 16.42 & 14.68 & 18.52 & 12.17 \\\midrule
\chatunivil & 24.62 & 23.08 & 19.51 & 15.56 & 12.75 & 9.80 & 14.84 & 17.52 & 18.41 & 7.71 & 16.93 & 18.52 \\\midrule\midrule
\internvlchat & 35.38 & 35.90 & 36.30 & 36.79 & 16.67 & 12.99 & 18.00 & 15.82 & 13.43 & 16.17 & 22.22 & 23.81 \\\midrule
\gptfouro & 31.54 & 37.69 & 39.01 & 53.33 & 22.06 & 17.89 & 26.52 & 33.82 & 15.92 & 26.62 & 26.72 & 37.83 \\
\bottomrule 
    \end{tabular}
   }
\label{tab:2024-video-3d}
\end{table*}

\begin{table*}[!h]
  \centering
  \small
  \caption{2024-Real videos with \sg}
  \scalebox{0.95}{
    \begin{tabular}{l| cc|cc|cc} 
    \toprule
     &  \multicolumn{2}{c|}{\bf what action} & \multicolumn{2}{c|}{\bf what object} & \multicolumn{2}{c}{\bf what relation}\\ 
    \cmidrule(lr){2-3}\cmidrule(lr){4-5}\cmidrule(lr){6-7}
 & {\bf DP} &  {\bf SP} & {\bf DP} &  {\bf SP}  & {\bf DP} &  {\bf SP}  \\\midrule\midrule
\videochatgpts & 18.40 & 19.47 & 13.44 & 12.66 & 14.84 & 9.49 \\\midrule
\videollavas & 23.47 & 37.33 & 16.28 & 27.13 & 25.30 & 39.17 \\\midrule
\videollamatwos & 21.87 & 14.13 & 28.94 & 24.81 & 31.14 & 22.14 \\\midrule
\videollamatwol & 25.60 & 16.27 & 26.87 & 17.83 & 17.03 & 11.19 \\\midrule
\chatunivis & 40.00 & 28.53 & 33.07 & 18.60 & 36.25 & 32.85 \\\midrule
\chatunivil & 30.67 & 20.00 & 27.13 & 19.90 & 24.09 & 21.90 \\\midrule\midrule
\internvlchat & 63.73 & 41.07 & 47.29 & 28.94 & 51.09 & 44.04 \\\midrule
\gptfouro & 57.33 & 59.73 & 41.86 & 39.79 & 74.45 & 75.18 \\
\bottomrule 
    \end{tabular}
   }
\label{tab:2024-video-sg}
\end{table*}
\clearpage

\section{Details of Experiments on Query Results Approximation Algorithms}
\label{app:query-approximation}

To experiment with different query results approximation approaches, we first conduct extensive experiments to evaluate a set of representative models against a subset of tasks for each task generator.
Then, we build an Oracle database with the obtained evaluation results, referred to as \dbname, and study different query results approximation methods with this Oracle database to verify their effectiveness.
We will release the \dbname for future studies of query results approximation or model performance prediction.

\subsection{Experiment details}

\paragraph{Setup.} 
For image question answering tasks, We select 6 representative open-sourced large multimodal language models (MLMs) from 3 model families: \instructblips and \instructblipl from \instructblip~\cite{dai2024instructblip}, 
\qwenvl and \qwenvlchat from \qwenvl~\cite{bai2023qwen}, and
\llavas and \llaval from \llava~\cite{liu2024visual}.
For video question answering tasks, We select 7 representative open-sourced Large Video Language Models from 5 model families: \videollamatwos and \videollamatwol from \videollamatwo~\cite{damonlpsg2023videollama},
\videochatgpts from \videochatgpt~\cite{Maaz2023VideoChatGPT}, \chatunivis and \chatunivil from \chatunivi~\cite{jin2023chatunivi}, \videollavas from \videollava~\cite{lin2023video}, and \videochattwos from \videochattwo~\cite{li2023mvbench}.
We evaluate the models against a subset of tasks whose statistics can be found in Table~\ref{tab:subset-stats}.
Since we generate 15 task instances for each task and involve multiple models, these lead to a total number of 24,240,780 <model, task instance> pairs in evaluation.
We evaluate the query results approximation methods on a series of query instances for each type of query.
These query instances cover all the subsets of tasks and models we evaluate, leading to a set of 1137 query instances in total (741 for ImageQA and 396 for VideoQA).
We set the budget to 2,000 task evaluations.
\begin{table*}[h]
  \centering
  \small
  \caption{\textbf{Statistics of evaluated tasks.} For each task, we generate 15 task instances for evaluation.}
  \resizebox{0.7\linewidth}{!}{
    \begin{tabular}{lllr} 
    \toprule
     &  {\bf Scenerio} & {\bf Task generator} & {\bf \# of tasks} \\ 
     \midrule

\multirow{19}{*}{\bf ImageQA} & \multirow{5}{*}{\bf \twod}   
& how many &  17,238\\ 
& & what & 12,740\\ 
& & where & 12,740\\ 
& & what attribute & 12,740\\ 
& & where attribute & 12,740\\  
\cmidrule{2-4}

& \multirow{11}{*}{\bf \threed} 
&  how many & 17,238\\ 
&  &  what & 12,740\\ 
&  &  where & 12,740\\ 
&  &  what attribute & 12,740\\ 
&  &  where attribute & 12,740\\ 
&  &  what size & 10,304\\ 
&  &  what attribute size & 7,840\\ 
&  &  where size & 10,304\\ 
&  &  what distance & 6,160\\ 
&  &  what attribute distance & 6,000\\ 
&  &  where distance & 6,160\\  
\cmidrule{2-4}

& \multirow{3}{*}{\bf real image w \sg} 
& what object & 10,000\\ 
&  & what attribute & 10,000\\ 
&  & what relation & 10,000\\  
\midrule

\multicolumn{4}{c}{\bf Total number of tasks: 144,966} \\ \midrule
   
\multirow{9}{*}{\bf VideoQA} & \multirow{6}{*}{\bf \threed} 
& what rotate video & 2,464\\ 
&  & what attribute rotate video & 7,840\\ 
&  & where rotate video & 2,464\\ 
&  & what distance video & 4,928\\ 
&  & what attribute distance video & 15,680\\
&  & where distance video & 4,928\\  
\cmidrule{2-4}

& \multirow{3}{*}{\bf Real video w \sg} 
& what object video & 10,000\\ 
&  & what action video & 10,000\\ 
&  & what relation video &  10,000\\ \midrule

\multicolumn{4}{c}{\bf Total number of tasks: 106,608} \\ 

\bottomrule 
\end{tabular}
}
  
  \label{tab:subset-stats}

\end{table*}

\clearpage

\paragraph{Evaluation metrics.}
To evaluate the query results approximation methods, we adopt different evaluation metrics for different types of queries. 
For Top-K queries, we report the Mean Rank and the Hit Rate: Mean Rank is the average of the ground truth rank of the K items returned by the query results approximation method, so a lower Mean Rank indicates the returned items are actually ranked higher and the query results approximation method is better; Hit Rate measures the percentage of the K returned items are actual Top-K items, so the higher is the better.
For the Threshold query and its variants (Model Comparison and Model Debugging query), we can treat them as a binary classification problem and adopt the Prediction, Recall, and F1-score as evaluation metrics.

\subsection{Experiments on approximations under different budgets.}

To evaluate the performance of approximation algorithms under different budgets, we conducted an experiment using \qwenvlchat as the target model on 2D how-many tasks. We tested three query approximation algorithms on four types of queries: Top-K query, Threshold query, Model comparison query, and Model debugging query. The experiments were performed under budgets of $1,000$, $2,000$, and $3,000$. The results of the experiment can be found in Table~\ref{tab:topk-query},~\ref{tab:threshold-query},~\ref{tab:model-comparison-query}, and~\ref{tab:model-debugging-query}.

The results demonstrate that the \emph{Active} approximation algorithm consistently outperforms the \emph{Random} and \emph{Fitting} algorithms across all query types and budget levels. 
In particular, for the Model Compare query, \emph{Active} achieves better results with a 2,000 budget than baselines with larger budgets. 
Also, we can see the performance increase rapidly with more budget, indicating that users could have more accurate results when using a larger budget

\begin{table*}[h]
  \centering
  \small
  \caption{The performance of Top-K query results approximation algorithms with different budgets. }
  \scalebox{0.7}{
    \begin{tabular}{l|cc|cc|cc} 
      \toprule
      \multirow{2}{*}{\bf Budget} & \multicolumn{2}{c|}{\bf Random} & \multicolumn{2}{c|}{\bf Fitting} & \multicolumn{2}{c}{\bf Active} \\ \cmidrule(lr){2-3}\cmidrule(lr){4-5}\cmidrule(lr){6-7}
      & {\bf MR} & {\bf HR (\%)} & {\bf MR} & {\bf HR (\%)} & {\bf MR} & {\bf HR (\%)} \\\midrule
      1,000 & 137.1 & 0.0 & 143.3 & 10.0 & 44.3 & 20.0 \\
      2,000 & 116.6 & 0.0 & 121.8 & 0.0 & 32.2 & 20.0 \\
      3,000 & 110.3 & 10.0 & 121.4 & 10.0 & 21.4 & 20.0 \\
      \bottomrule 
    \end{tabular}
  }
  \label{tab:topk-query}
\end{table*}

\begin{table*}[h]
  \centering
  \small
  \caption{The performance of Threshold query results approximation algorithms with different budgets. }
  \scalebox{0.75}{
    \begin{tabular}{l|ccc|ccc|ccc}
      \toprule
      \multirow{2}{*}{\bf Budget} & \multicolumn{3}{c|}{\bf Random} & \multicolumn{3}{c|}{\bf Fitting} & \multicolumn{3}{c}{\bf Active} \\ \cmidrule(lr){2-4}\cmidrule(lr){5-7}\cmidrule(lr){8-10}
      & {\bf P (\%)} & {\bf R (\%)} & {\bf F1 (\%)} & {\bf P (\%)} & {\bf R (\%)} & {\bf F1 (\%)} & {\bf P (\%)} & {\bf R (\%)} & {\bf F1 (\%)} \\\midrule
      1,000 & 42.61 & 31.82 & 36.43 & 48.48 & 10.39 & 17.11 & 45.0 & 11.69 & 18.56 \\
      2,000 & 43.90 & 35.06 & 38.99 & 43.44 & 34.42 & 38.41 & 43.44 & 34.42 & 38.41 \\
      3,000 & 45.38 & 38.31 & 41.55 & 45.89 & 43.51 & 44.67 & 50.93 & 71.43 & 59.46 \\
      \bottomrule
    \end{tabular}
  }
  \label{tab:threshold-query}
\end{table*}

\begin{table*}[h]
  \centering
  \small
  \caption{The performance of Model comparison query results approximation algorithms with different budgets.}
  \scalebox{0.75}{
    \begin{tabular}{l|ccc|ccc|ccc}
      \toprule
      \multirow{2}{*}{\bf Budget} & \multicolumn{3}{c|}{\bf Random} & \multicolumn{3}{c|}{\bf Fitting} & \multicolumn{3}{c}{\bf Active} \\ \cmidrule(lr){2-4}\cmidrule(lr){5-7}\cmidrule(lr){8-10}
      & {\bf P (\%)} & {\bf R (\%)} & {\bf F1 (\%)} & {\bf P (\%)} & {\bf R (\%)} & {\bf F1 (\%)} & {\bf P (\%)} & {\bf R (\%)} & {\bf F1 (\%)} \\\midrule
      1,000 & 100.0 & 5.86 & 11.08 & 88.34 & 6.73 & 12.51 & 61.22 & 28.71 & 39.09 \\
      2,000 & 100.0 & 11.37 & 20.42 & 62.88 & 31.82 & 42.26 & 75.18 & 41.44 & 53.43 \\
      3,000 & 100.0 & 17.41 & 29.66 & 69.74 & 43.19 & 53.35 & 82.81 & 52.30 & 64.11 \\
      \bottomrule
    \end{tabular}
  }
  \label{tab:model-comparison-query}
\end{table*}

\begin{table*}[!h]
  \centering
  \small
  \caption{The performance of Model debugging query results approximation algorithms with different budgets.}
  \scalebox{0.75}{
    \begin{tabular}{l|ccc|ccc|ccc}
      \toprule
      \multirow{2}{*}{\bf Budget} & \multicolumn{3}{c|}{\bf Random} & \multicolumn{3}{c|}{\bf Fitting} & \multicolumn{3}{c}{\bf Active} \\ \cmidrule(lr){2-4}\cmidrule(lr){5-7}\cmidrule(lr){8-10}
      & {\bf P (\%)} & {\bf R (\%)} & {\bf F1 (\%)} & {\bf P (\%)} & {\bf R (\%)} & {\bf F1 (\%)} & {\bf P (\%)} & {\bf R (\%)} & {\bf F1 (\%)} \\\midrule
      1,000 & 100.0 & 6.34 & 11.92 & 100.0 & 6.34 & 11.92 & 100.0 & 6.93 & 12.96 \\
      2,000 & 100.0 & 13.50 & 23.79 & 97.18 & 13.58 & 23.83 & 100.0 & 15.0 & 26.09 \\
      3,000 & 100.0 & 18.82 & 31.68 & 95.29 & 19.13 & 31.87 & 100.0 & 22.01 & 36.08 \\
      \bottomrule
    \end{tabular}
  }
  \label{tab:model-debugging-query}
\vspace{-5mm}
\end{table*}

\clearpage

\subsection{Query results approximation experiments in ImageQA}

\begin{figure}[!h]
  \centering
\includegraphics[width=0.9\linewidth]{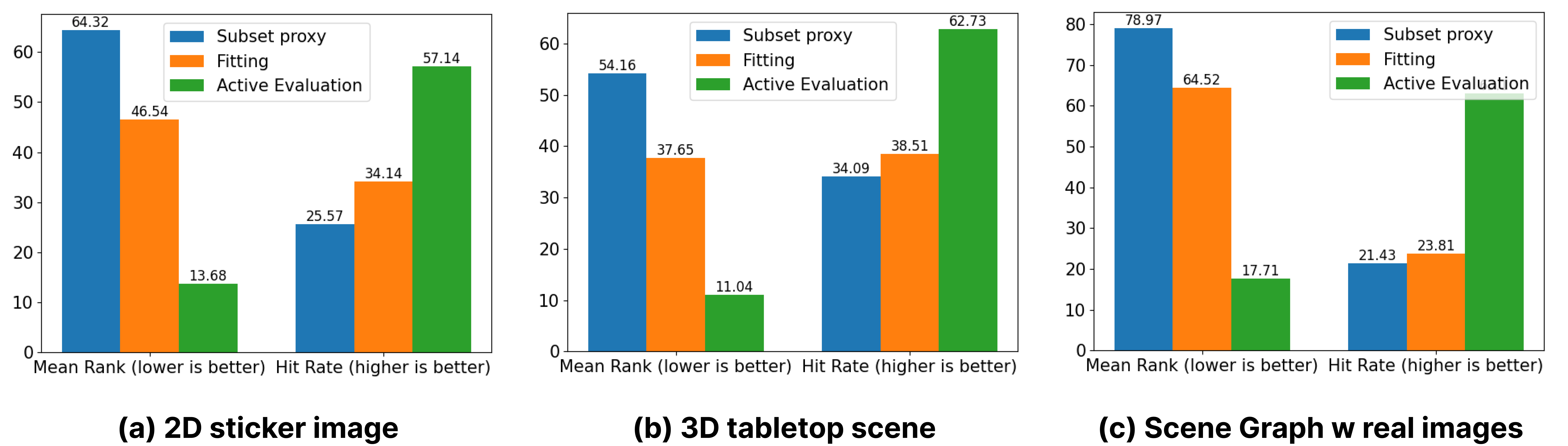}
  \caption{\textbf{Top-K Query.} These three bar graphs display the performance of three query approximation methods in Top-K Query, measured by Mean Rank and Hit Rate.} 
  \label{fig:top-k-query}
\end{figure}

\begin{figure}[!h]
  \centering
\includegraphics[width=0.9\linewidth]{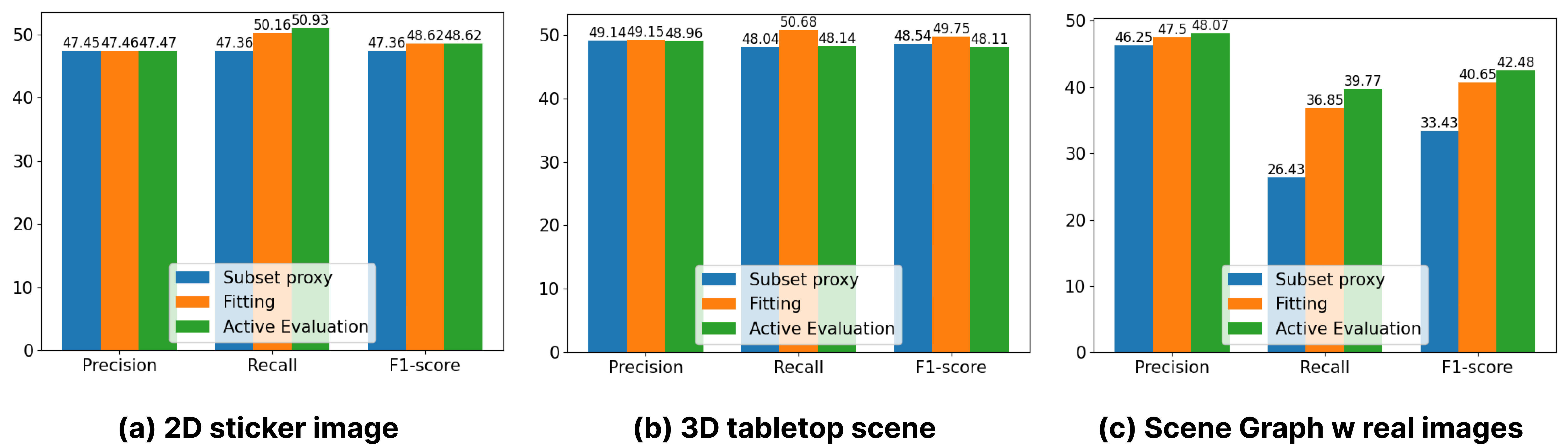}
  \caption{\textbf{Threshold Query.} These three bar graphs display the performance of three query approximation methods in Threshold Query, measured by Precision, Recall, and F1-score.}
  \label{fig:threshold-query}
\end{figure}

\begin{figure}[!h]
  \centering
\includegraphics[width=0.9\linewidth]{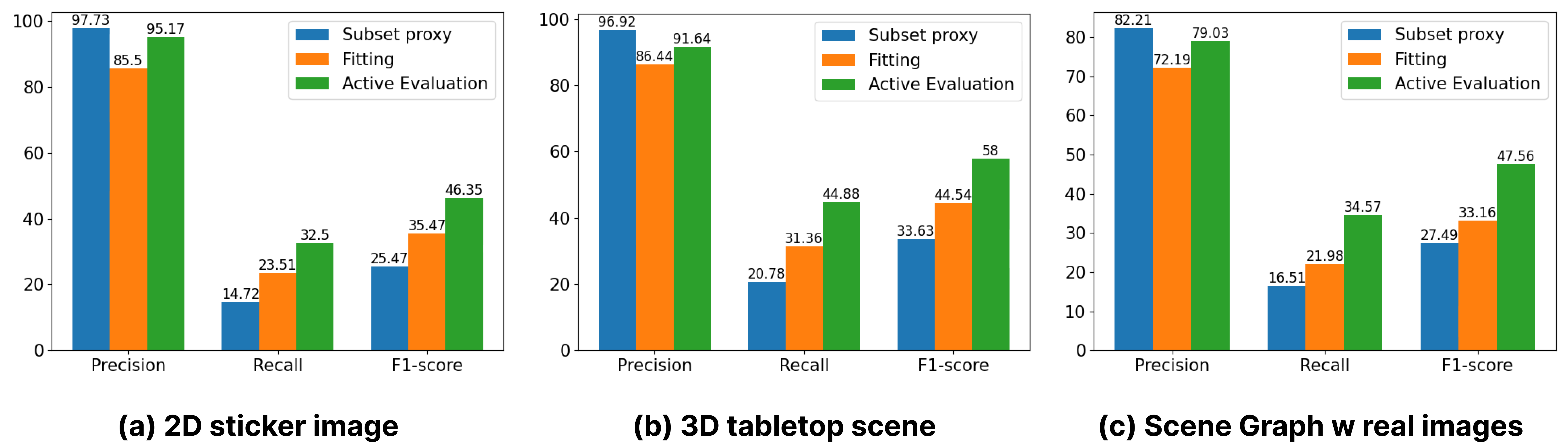}
  \caption{\textbf{Model Debugging Query.} These three bar graphs display the performance of three query approximation methods in Model Debugging Query, measured by Precision, Recall, and F1-score.} 
  \label{fig:debug-query}
\end{figure}

\begin{figure}[!h]
  \centering
\includegraphics[width=0.9\linewidth]{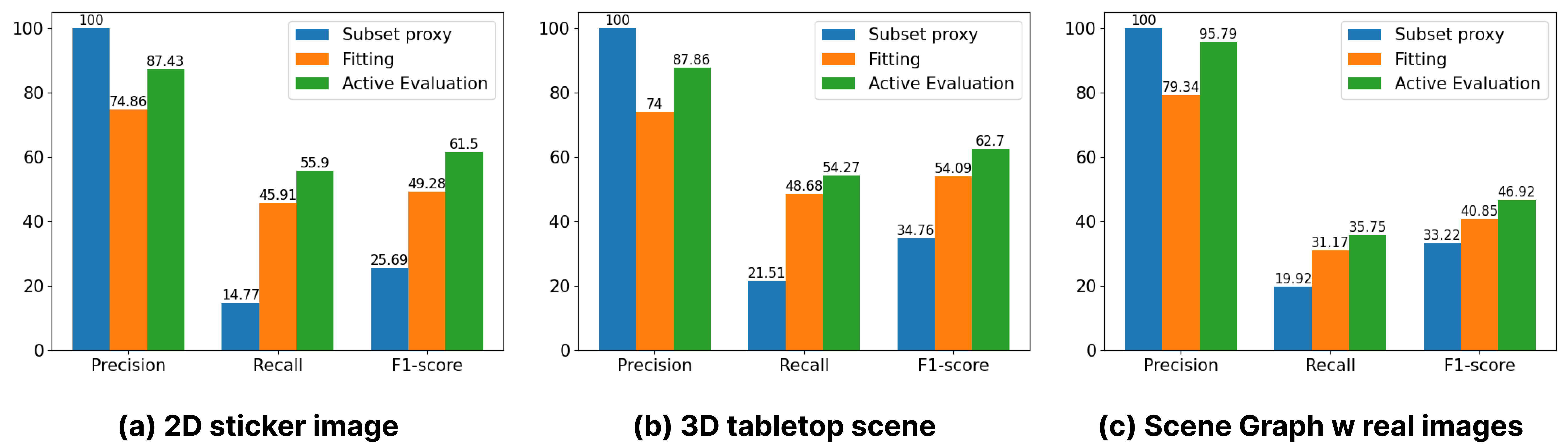}
  \caption{\textbf{Model Comparison Query.} These three bar graphs display the performance of three query approximation methods in Model Comparison Query, measured by Precision, Recall, and F1-score.} 
  \label{fig:compare-query}
\end{figure}

\clearpage

\subsection{Query results approximation experiments in VideoQA}

\begin{figure}[!h]
  \centering
\includegraphics[width=0.6\linewidth]{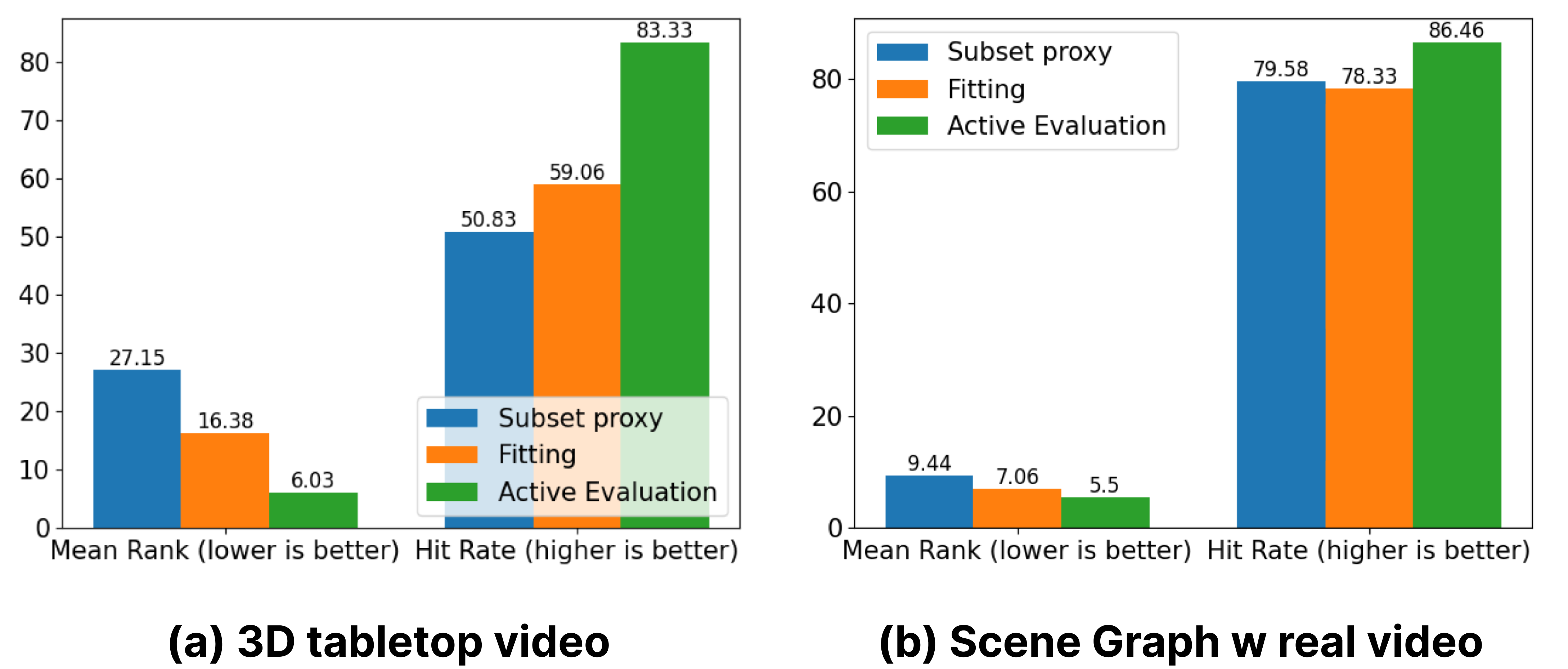}
  \caption{\textbf{Top-K Query in VideoQA.} These three bar graphs display the performance of three query approximation methods in Top-K Query, measured by Mean Rank and Hit Rate.} 
  \label{fig:top-k-query-video}
\end{figure}

\begin{figure}[!h]
  \centering
\includegraphics[width=0.6\linewidth]{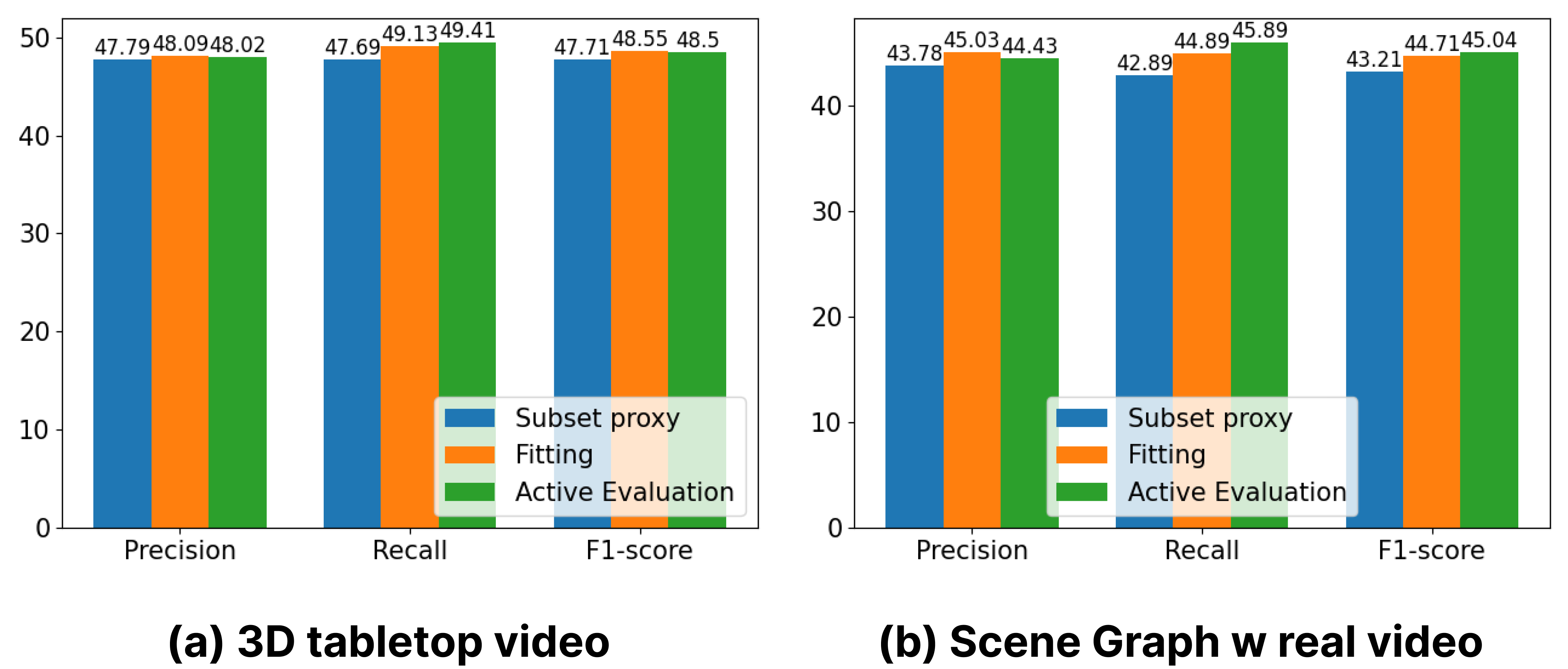}
  \caption{\textbf{Threshold Query in VideoQA.} These three bar graphs display the performance of three query approximation methods in Threshold Query, measured by Precision, Recall, and F1-score.} 
  \label{fig:threshold-query-video}
\end{figure}

\begin{figure}[!h]
  \centering
\includegraphics[width=0.6\linewidth]{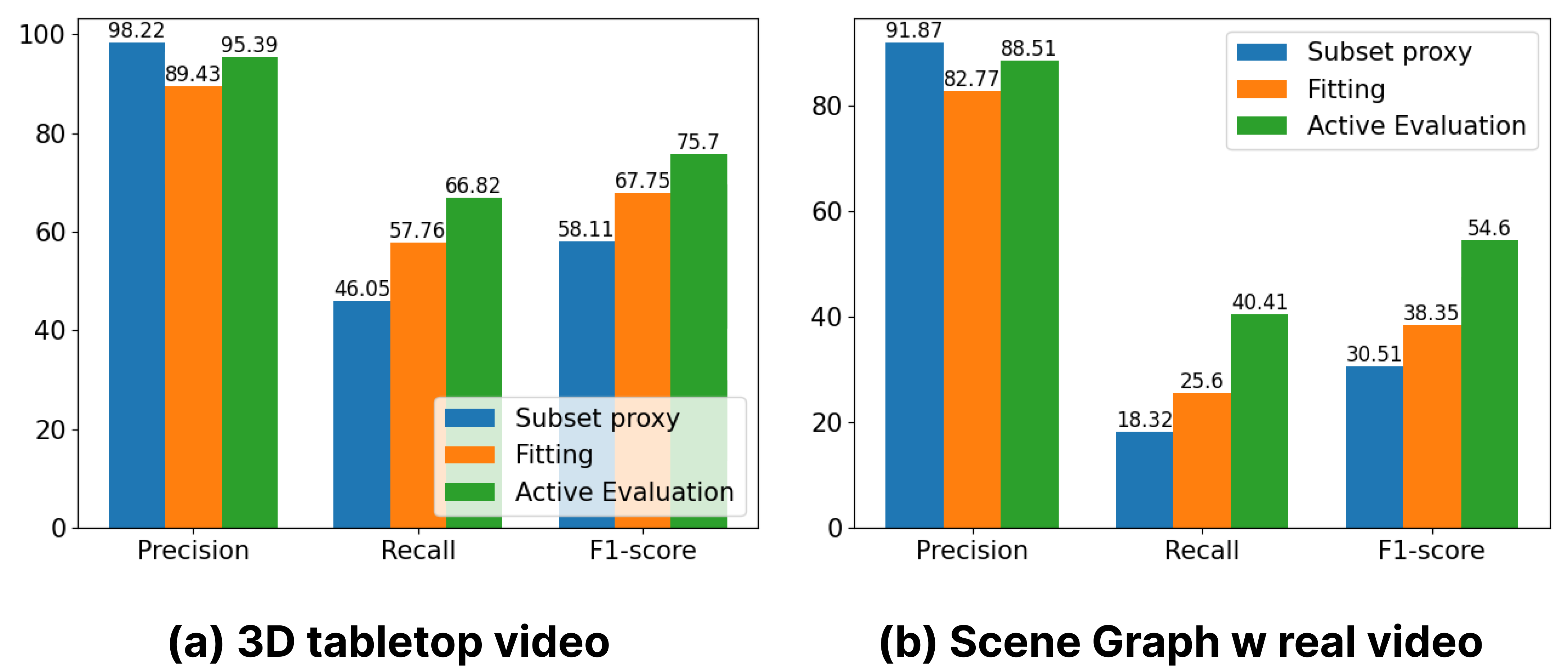}
  \caption{\textbf{Model Debugging Query in VideoQA.} These three bar graphs display the performance of three query approximation methods in Model Debugging Query, measured by Precision, Recall, and F1-score.} 
  \label{fig:debug-query-video}
\end{figure}

\begin{figure}[!h]
  \centering
\includegraphics[width=0.6\linewidth]{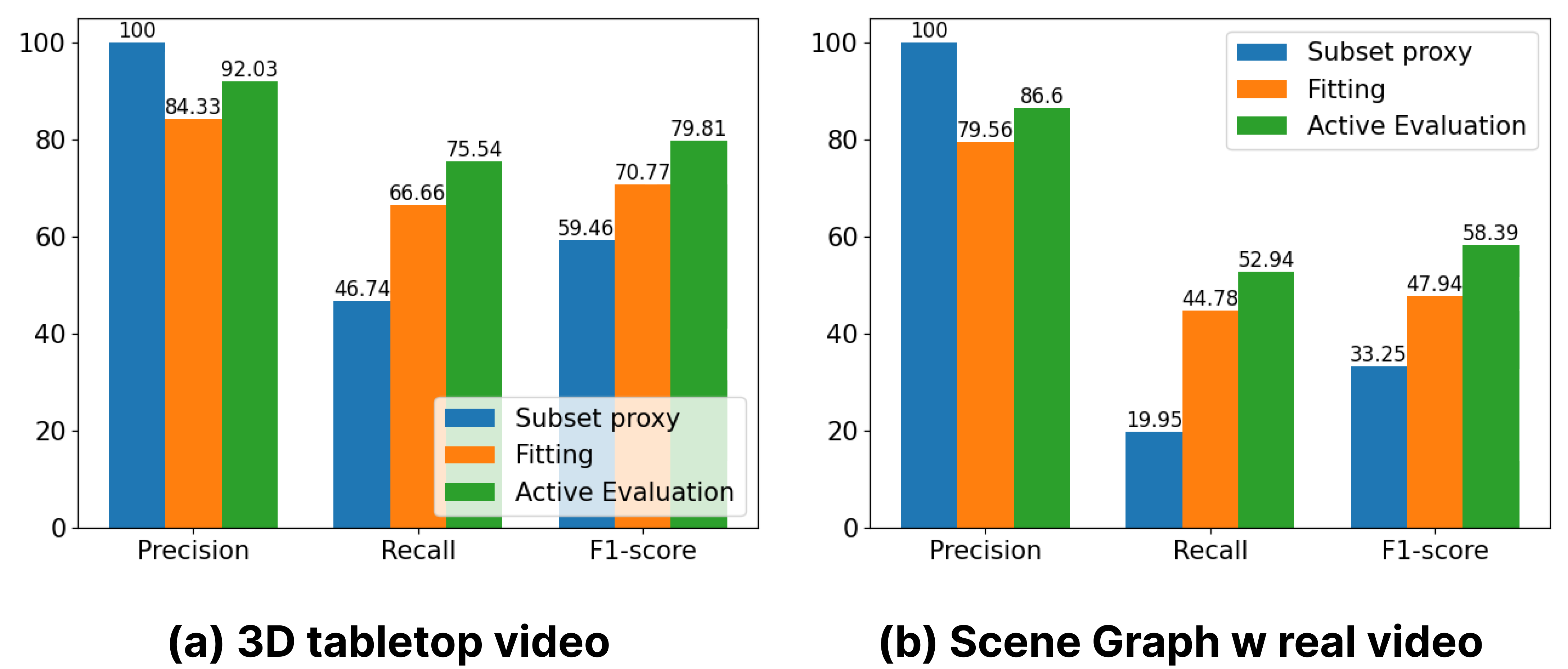}
  \caption{\textbf{Model Comparison Query in VideoQA.} These three bar graphs display the performance of three query approximation methods in Model Comparison Query, measured by Precision, Recall, and F1-score.} 
  \label{fig:compare-query-video}
\end{figure}

\clearpage
\section{Details of Analysis and Case Study}
\label{app:analysis}
\subsection{What task metadata are models good or bad at?}
\begin{figure}[!h]
  \centering
  \includegraphics[width=\linewidth]{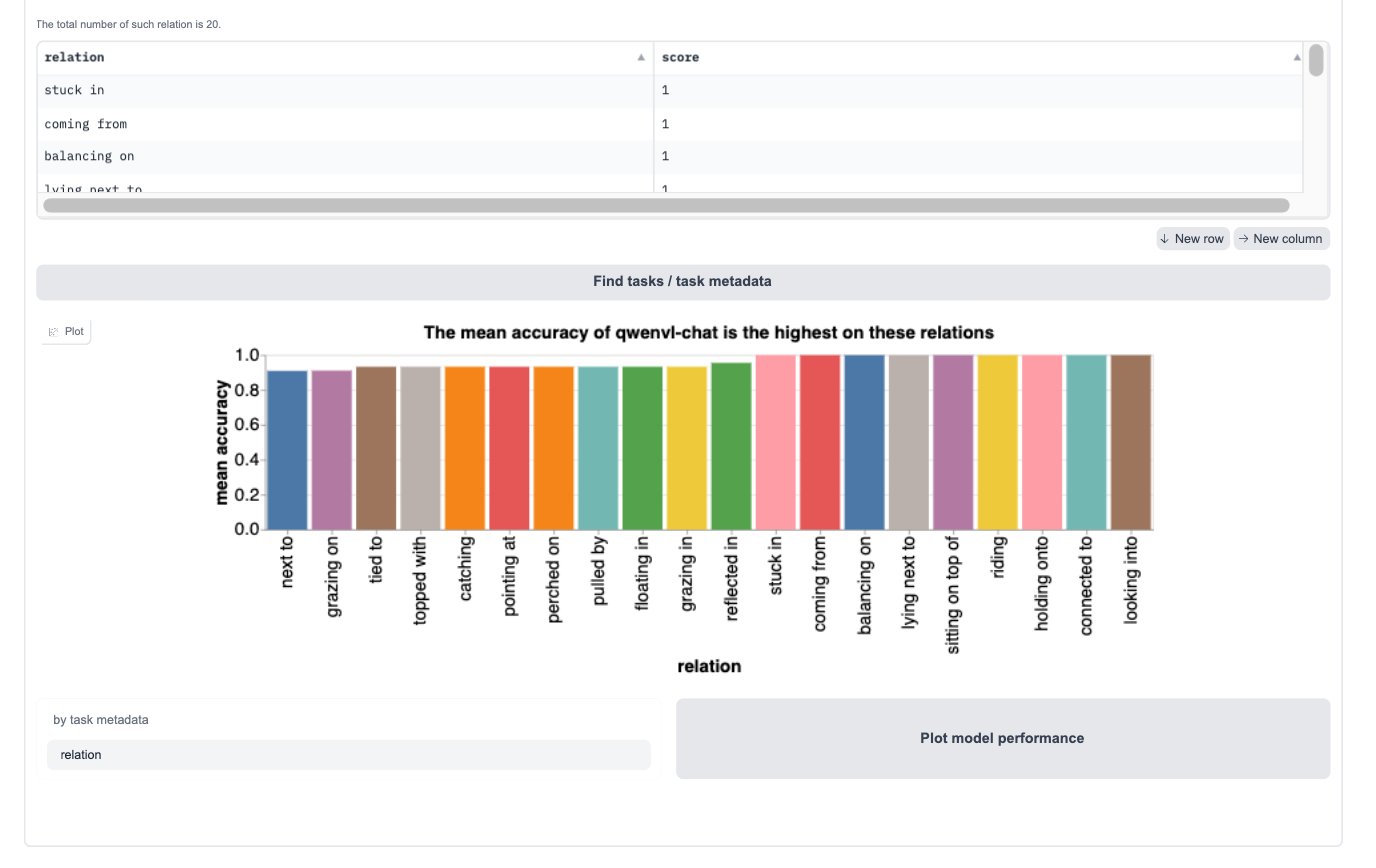}
  \caption{ImageQA: Best relations}
  \label{fig:image-best-relations}
\end{figure}

\begin{figure}[!h]
  \centering
  \includegraphics[width=\linewidth]{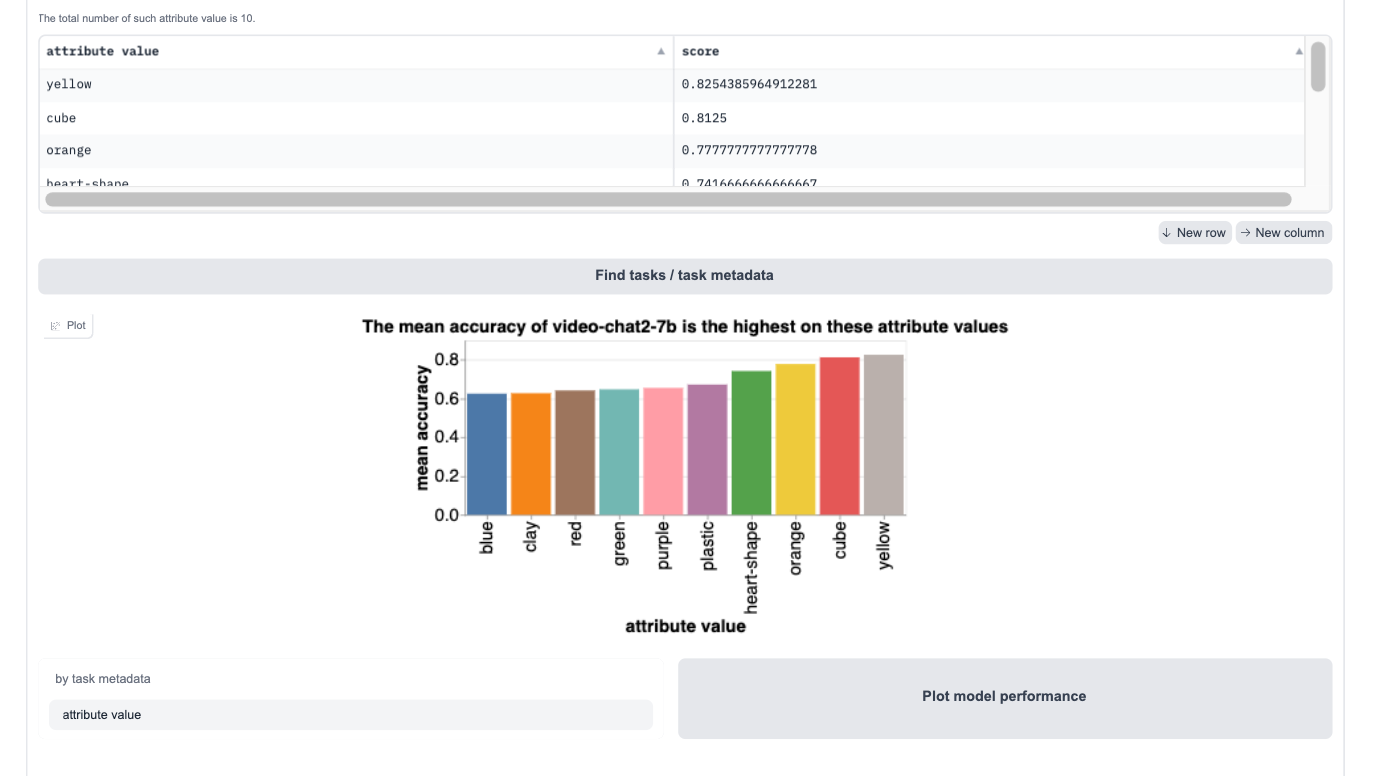}
  \caption{VideoQA: Best attributes}
  \label{fig:video-best-attributes}
\end{figure}

\begin{figure}[!h]
  \centering
  \includegraphics[width=\linewidth]{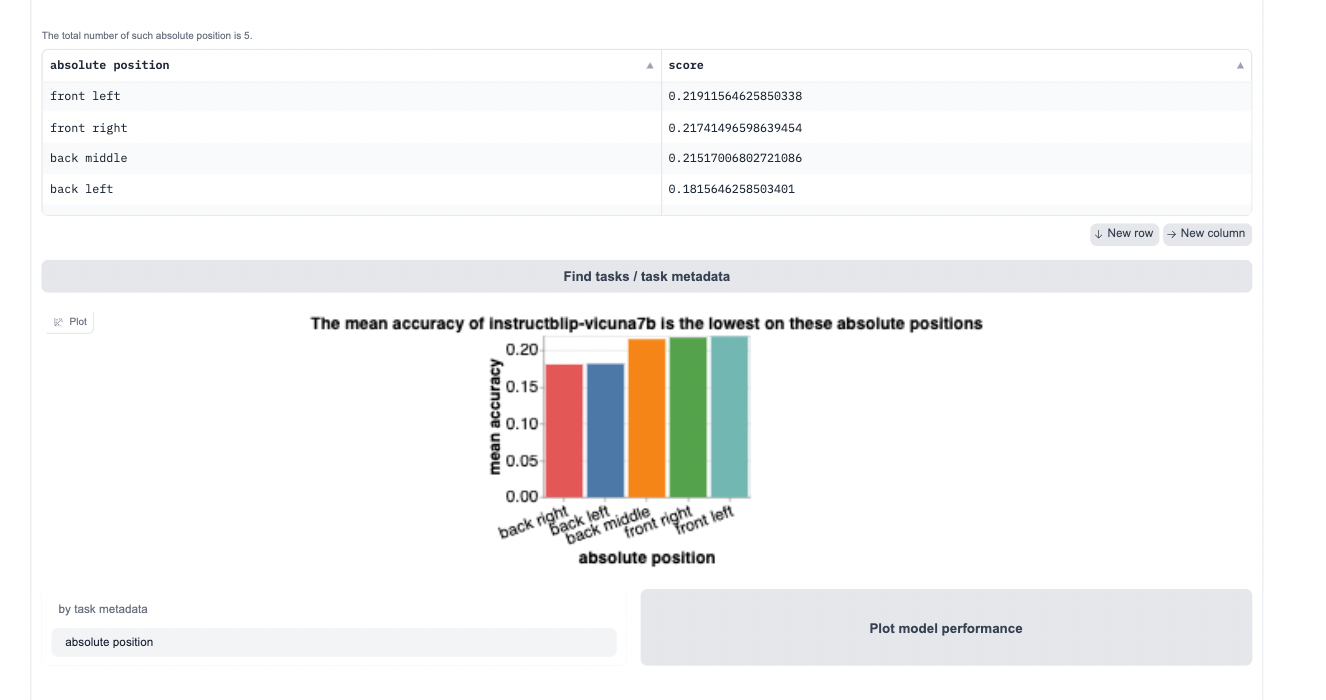}
  \caption{ImageQA: Worst positions}
  \label{fig:image-worst-positions}
\end{figure}

\begin{figure}[!h]
  \centering
  \includegraphics[width=\linewidth]{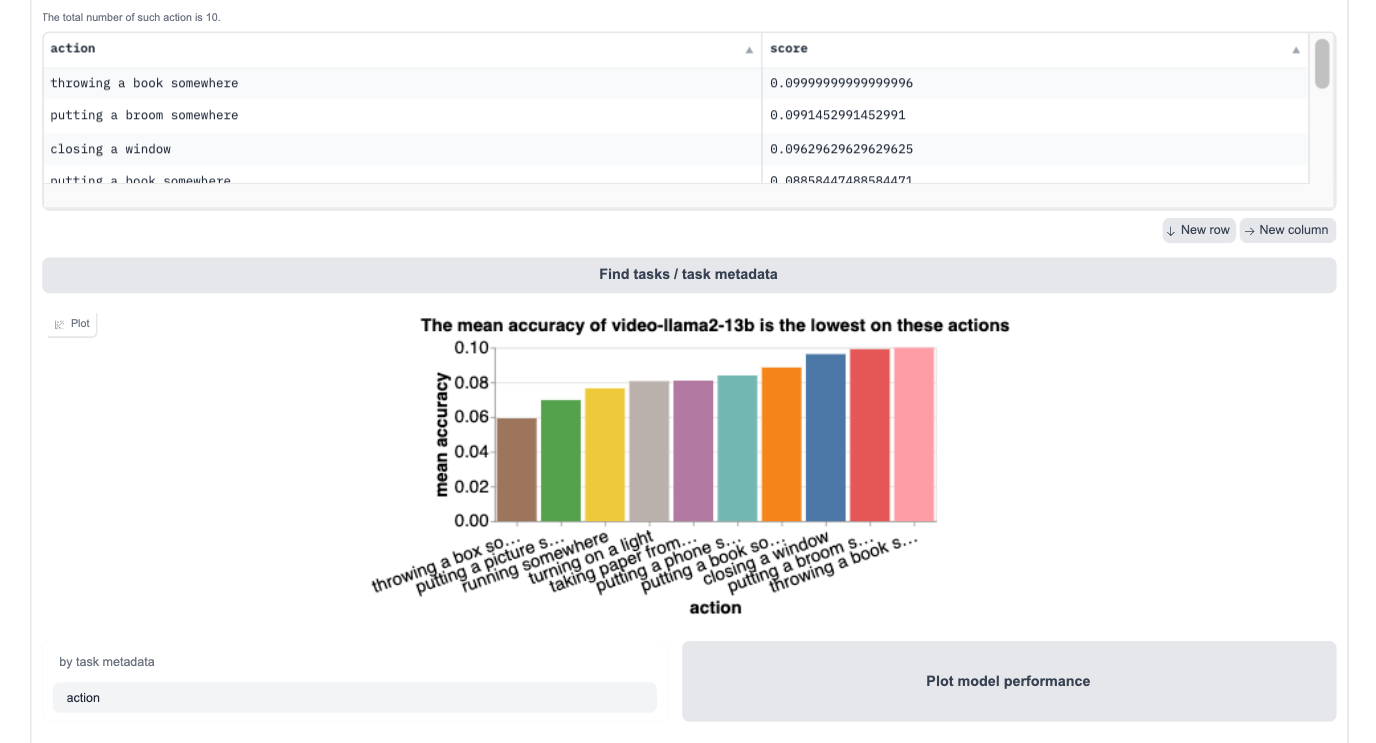}
  \caption{VideoQA: Worst actions}
  \label{fig:video-worst-actions}
\end{figure}

To obtain a more finegrained understanding of models’ skill sets, we also leverage our interface to examine the top and bottom task metadata related to models’ best and worst skills. For example, as \qwenvlchat performs the best on relation understanding across models and skills, we identify the top 20 relations where \qwenvlchat achieves the highest accuracies (Figure \ref{fig:image-best-relations}) and find that they are mostly actions. Similarly, on VideoQA tasks related to attribute understanding, we are also able to find the attribute values \videochattwos is the best at and learn that they are mostly associated with color instead of shape or material (Figure \ref{fig:video-best-attributes}). On the other hand, we learn that \instructblipl does terribly on spatial understanding especially when the object’s absolute position is in the back, followed by front right or left (Figure \ref{fig:image-worst-positions}); and among the actions \videollamatwol performs the worst on, most involve “putting” or “throwing” something (Figure \ref{fig:video-worst-actions}). 

\clearpage

\subsection{How do small models compare against large models? (continued) }
As discussed in the main paper, we observe that large multi-modal models collectively perform better than smaller models on ImageQA tasks (Figure \ref{fig:image-small-vs-large}). Nevertheless, this finding might not always hold for individual models. Through t-tests with pairs of small and large models from the same source, we find one exception: \instructblips ($\mu$ = 0.63) significantly outperforms \instructblipl  ($\mu$ = 0.49) on relation understanding (with p-value = 0) (Figure \ref{fig:instructblip-small-vs-large}). 

\begin{figure}[!h]
  \centering
  \includegraphics[width=\linewidth]{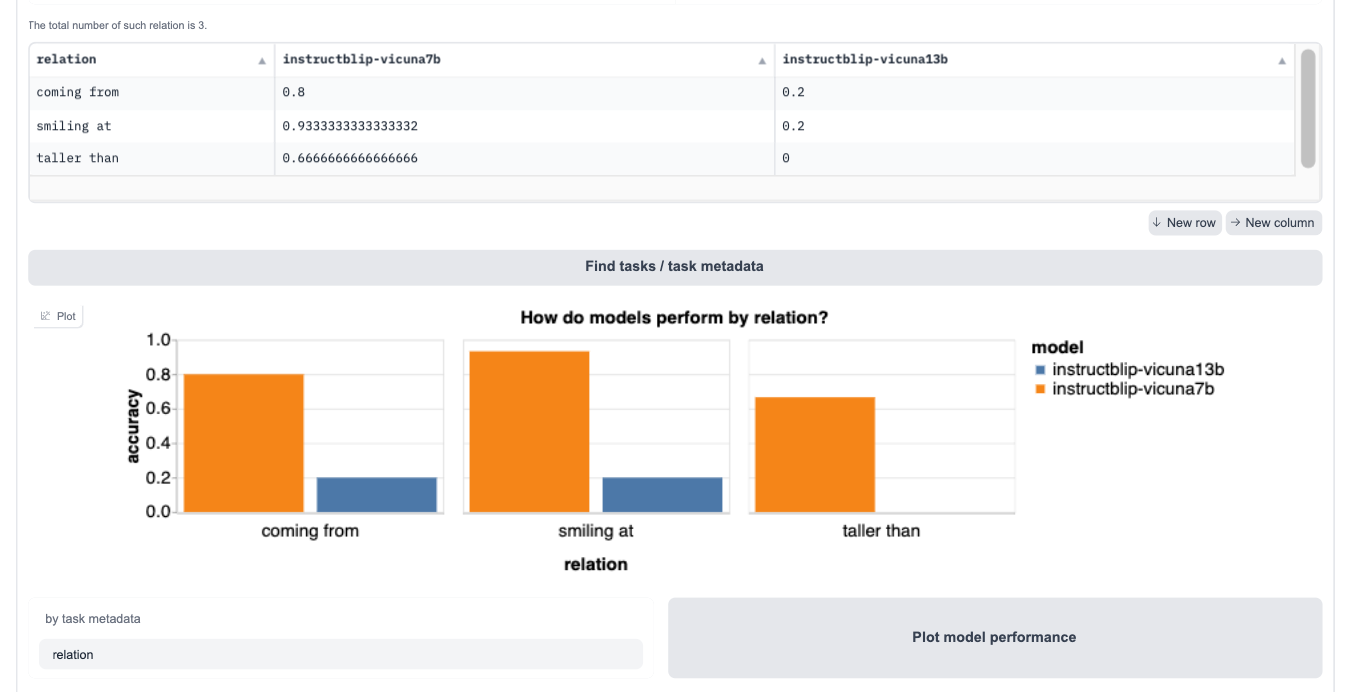}
  \caption{\instructblips vs. \instructblipl relations}
  \label{fig:instructblip-small-vs-large-relations}
\end{figure}

\begin{figure}[!h]
  \centering
  \includegraphics[width=\linewidth]{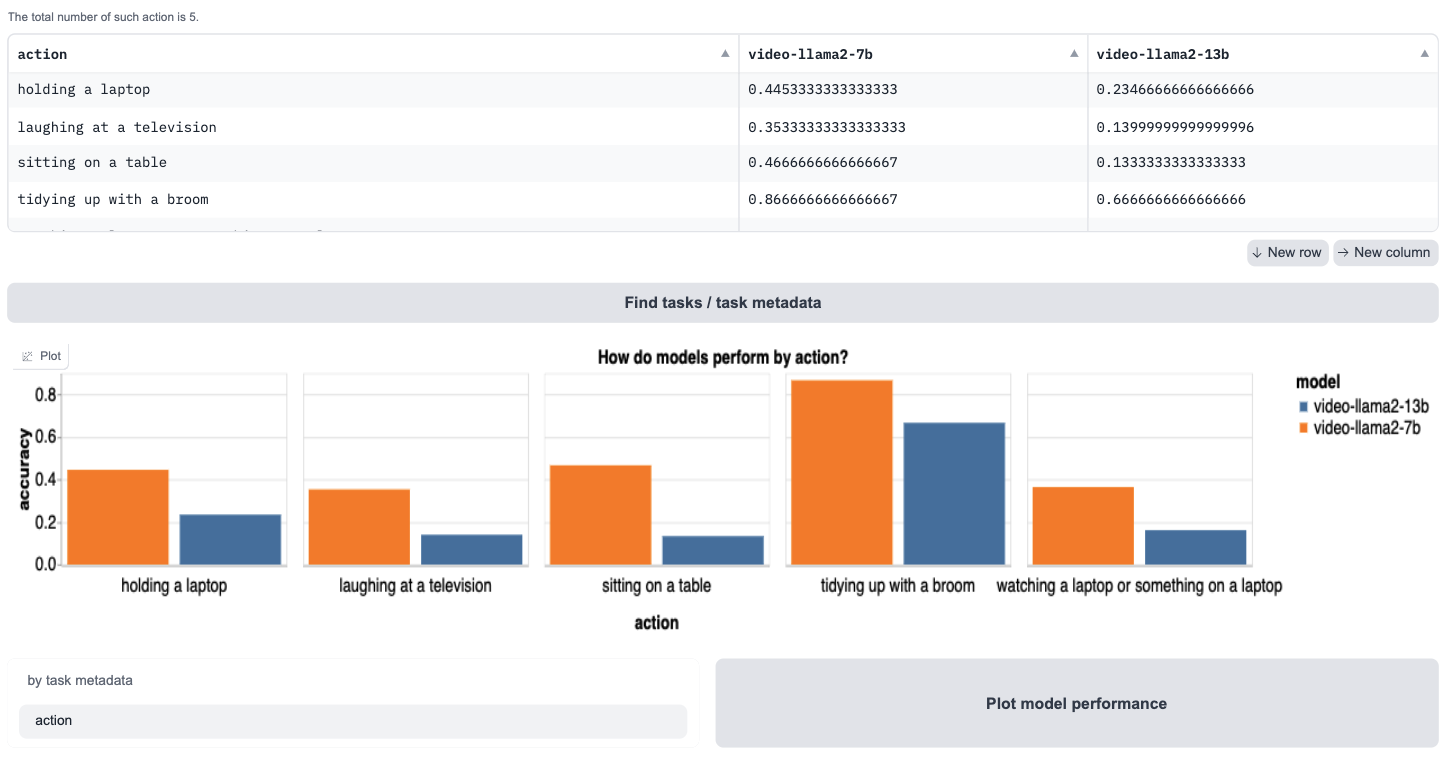}
  \caption{\videollamatwos vs. \videollamatwol actions}
  \label{fig:fig:videollama-small-vs-large-actions}
\end{figure}

\begin{figure}[!h]
  \centering
  \includegraphics[width=\linewidth]{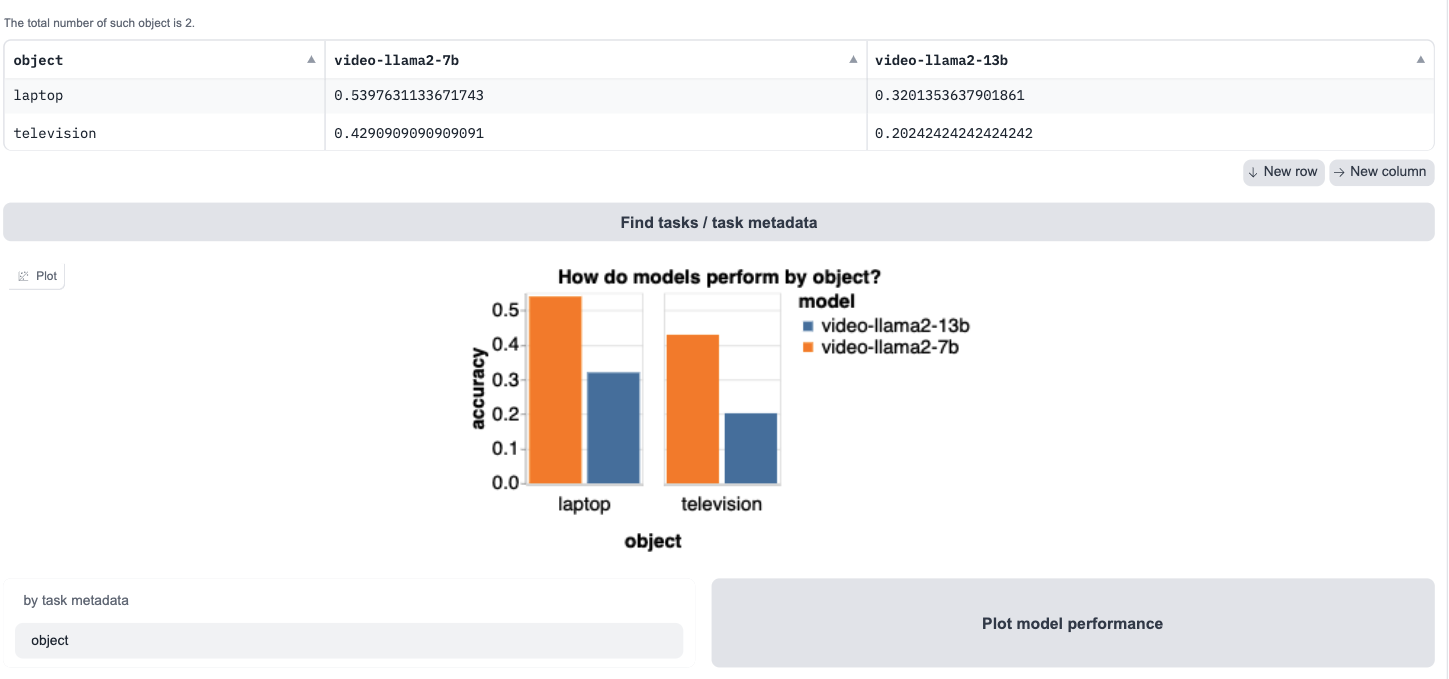}
  \caption{\videollamatwos vs. \videollamatwol objects}
  \label{fig:fig:videollama-small-vs-large-objects}
\end{figure}

Further, upon a closer look with our interface, we identify a few relations where \instructblips outperforms \instructblipl by a large margin e.g. 50\% (Figure \ref{fig:instructblip-small-vs-large-relations}). Similarly, we also retrieve a few actions and objects where \videollamatwos performs much better e.g. by 20\% than \videollamatwol (Figures \ref{fig:fig:videollama-small-vs-large-actions} and \ref{fig:fig:videollama-small-vs-large-objects}).

\clearpage

\subsection{Do \name yield results similar to existing benchmarks?}
\label{app:tallyqa}
To check whether our \name reflects model performance similarly to an existing benchmark, we conducted a case study testing six open-source models on both the well-known TallyQA Counting benchmark~\cite{Acharya2018TallyQAAC} (we selected 10,000 simple questions and 10,000 complex from the whole set) and 2D how-many and 3D how-many tasks in \randomname. (Table \ref{tab:tallyqa-comparison}), the results demonstrate a notable correlation. For instance, the \llaval is the best-performing model in both TallyQA and how-many tasks in \randomname.
The Spearman ranking coefficient for the correlation between the 2D how-many tasks and TallyQA is 0.714 (p-value = 0.111), while for the 3D how-many tasks, it is 0.543 (p-value = 0.266). These results indicate positive correlations of model performance between our tasks and existing ones, validating that \name can effectively reflect model performance in a manner similar to existing benchmark. 

\begin{table*}[!h]
  \centering
  \small
  \caption{Models performance on TallyQA Counting benchmark and 2D how-many and 3D how-many in our \randomname}
  \scalebox{0.95}{
    \begin{tabular}{l|c|c|c} 
    \toprule
    \textbf{Model} & \textbf{TallyQA} & \textbf{2D How Many} & \textbf{3D How Many} \\ 
    \midrule
    \llavas & 35.90 & 42.00 & 38.67 \\ 
    \midrule
    \llaval & \textbf{38.33} & \textbf{49.33} & \textbf{46.67} \\ 
    \midrule
    \qwenvl & 18.79 & 30.67 & 32.33 \\ 
    \midrule
    \qwenvlchat & 32.07 & 39.67 & 45.00 \\ 
    \midrule
    \instructblips & 29.92 & 23.67 & 32.67 \\ 
    \midrule
    \instructblipl & 33.22 & 26.67 & 32.00 \\ 
    \bottomrule 
    \end{tabular}
  }
  \label{tab:tallyqa-comparison}
\end{table*}

\clearpage

\section{Datasheet for \randomname}
\subsection{Motivation}
\begin{enumerate}
\item {\bf For what purpose was the dataset created?} \\ 
\randomname is created as a randomly selected subset of \nameone to provide an overview of \name.
\item {\bf Who created the dataset and on behalf of which entity?} \\
It was created by the authors of this paper.
\item {\bf Who funded the creation of the dataset?} \\
The creation of the dataset was funded by the institute to which the authors belong.
\end{enumerate}
\subsection{Composition}
\begin{enumerate}
\item {\bf What do the instances that comprise the dataset represent (e.g., documents, photos, people, countries?)} \\
The dataset consists of 2D and 3D synthetic images, videos, and real images and videos, each accompanied by corresponding task plans, questions, options, and ground truths.
\item {\bf How many instances are there in total (of each type, if appropriate)?} \\
ImageQA: 5,700 instances (19 types of task generators, each with 300 instances per split per generator type).
VideoQA: 2,700 instances (9 types of task generators, each with 300 instances per split per generator type).
\item {\bf Does the dataset contain all possible instances, or is it a sample of instances from a larger set? } \\ 
This dataset is a randomly selected subset from the \nameone task space. Additional tasks can be generated by users based on their needs.
\item {\bf Is there a label or target associated with each instance?} \\
Yes, each instance includes both input and targets.
\item {\bf Is any information missing from individual instances?} \\
No.
\item {\bf Are there recommended data splits (e.g., training, development/validation, testing)?} \\
For ImageQA, there are 19 splits, each containing 300 instances from a specific task type. For VideoQA, there are 9 splits, each also containing 300 instances from a specific task type.
\item {\bf Are there any errors, sources of noise, or redundancies in the dataset?} \\
For real images and videos, the scene graphs may contain a small amount of noise due to human annotation bias. However, this does not have a significant impact on the research.
\item {\bf Is the dataset self-contained, or does it link to or otherwise rely on external resources (e.g., websites, tweets, other datasets)?} \\
The 3D objects used in the 2D sticker and 3D table scenarios are sourced from Objaverse. The real image scenarios are derived from the GQA versions of Visual Genome (VG), and the real videos are obtained from AGQA.
\item {\bf Does the dataset contain data that might be considered confidential?} \\ 
No.
\item {\bf Does the dataset contain data that, if viewed directly, might be offensive, insulting, threatening, or might otherwise cause anxiety?} \\
No.
\end{enumerate}

\subsection{Collection Process}
\begin{enumerate}
\item {\bf How was the data associated with each instance acquired?} \\
The 3D objects used in the 2D sticker and 3D table scenarios are sourced from Objaverse. The real image scenarios are derived from the GQA versions of VG, while the real videos are from AGQAs. References are provided in Section 3 of the main text.
\item {\bf What mechanisms or procedures were used to collect the data (e.g., hardware apparatus or sensor, manual human curation, software program, software API)?} \\
We used multiple NVIDIA A6000 and A100 GPUs to run Blender for rendering the synthetic scenes. Questions, options, and ground truth were generated by task generators (Python code).
\item {\bf Who was involved in the data collection process (e.g., students, crowdworkers, contractors) and how were they compensated (e.g., how much were crowdworkers paid)?} \\
The authors of this paper were directly involved in the data collection process, annotating the attributes of 3d objects and build the taxonomy themselves.
\item {\bf Over what timeframe was the data collected?} \\
The final version of the dataset was generated in June, 2024.
\end{enumerate}

\subsection{Uses}
\begin{enumerate}
\item {\bf Has the dataset been used for any tasks already?}\\ 
No, this dataset has not been used for any tasks yet.
\item {\bf What (other) tasks could the dataset be used for?}\\ 
This data can also be used in various computer vision tasks, such as localization, object detection, etc.
\item {\bf Is there anything about the composition of the dataset or the way it was collected and preprocessed/cleaned/labeled that might impact future uses?}\\ 
No.
\item {\bf Are there tasks for which the dataset should not be used?}\\ 
No.
\end{enumerate}

\subsection{Distribution}
\begin{enumerate}
\item {\bf Will the dataset be distributed to third parties outside of the entity (e.g., company, institution, organization) on behalf of which the dataset was created?}\\
Yes, the dataset is open to the public.
\item {\bf How will the dataset be distributed (e.g., tarball on website, API, GitHub)?}\\
You can access our dataset via the links below:

Dataset (ImageQA): \url{https://huggingface.co/datasets/weikaih/TaskMeAnything-v1-videoqa-random}

Dataset (VideoQA): \url{https://huggingface.co/datasets/weikaih/TaskMeAnything-v1-videoqa-random}

Code: \url{https://github.com/JieyuZ2/TaskMeAnything}

\item {\bf Have any third parties imposed IP-based or other restrictions on the data associated with the instances?}\\
No.
\item {\bf Do any export controls or other regulatory restrictions apply to the dataset or to individual instances?}\\
No.
\end{enumerate}

\subsection{Maintenance}
\begin{enumerate}
\item {\bf Who will be supporting/hosting/maintaining the dataset?} \\ 
The authors of this paper will support, host, and maintain the dataset.
\item {\bf How can the owner/curator/manager of the dataset be contacted (e.g., email address)?} \\ 
The owner/curator/manager(s) of the dataset can be contacted through the following email: Jieyu Zhang (jieyuz2@cs.washington.edu)
\item {\bf Is there an erratum?} \\ 
No. If errors are found in the future, we will release errata on the GitHub repo for the dataset: (https://github.com/JieyuZ2/TaskMeAnything).
\item {\bf Will the dataset be updated (e.g., to correct labeling errors, add new instances, delete instances)?} \\ 
Yes, the datasets will be updated whenever necessary to ensure accuracy, and announcements will be made accordingly. These updates will be posted on the GitHub repo for the dataset: (https://github.com/JieyuZ2/TaskMeAnything).
\item {\bf If the dataset relates to people, are there applicable limits on the retention of the data associated with the instances (e.g., were the individuals in question told that their data would be retained for a fixed period of time and then deleted?) } \\ 
N/A
\item {\bf Will older versions of the dataset continue to be supported/hosted/maintained?} \\ 
Yes. Older versions of the dataset will continue to be maintained and hosted.
\item {\bf If others want to extend/augment/build on/contribute to the dataset, is there a mechanism for them to do so?} \\ 
Yes, one can extend the dataset by simply adding more source data and task generators, or by generating more instances from the existing task space.
\end{enumerate}

\section{Task Generator Cards}
\label{app:cards}

\begin{figure}
\begin{framed}
 \centering
{\Large {\bf WhatGridTaskGenerator}}
\begin{itemize}[leftmargin=*]

\item {\bf Basic Information}. 
\begin{itemize}
\item {\bf Task Type}.  ImageQA
\item {\bf Question Type}.  what object
\item {\bf Answer Type}.  object category
\item {\bf Image Type}. 2D sticker image
\item {\bf The model capability to evaluate}. object recognition with / without reference
\end{itemize}

\item {\bf Source Data}. 
\begin{itemize}
\item rendering images of objects from Objaverse
\item Annotations regarding object category, attribute, and shape
\end{itemize}

\item {\bf Task Plan Schema}.  
\begin{itemize}
\item \textbf{question type}: \texttt{string}. The question type of these tasks will be "what".
\item \textbf{grid number}: \texttt{integer}. The number of diagonal grids of the image, $N$ indicates there are $N\times N$ grids in the image. Support \{2, 3\}.
\item \textbf{target category}: \texttt{string}. The category name of the target object.
\item \textbf{absolute position}: \texttt{string}. The absolute position of the target object in the grid. It is a number ranging from 0 to 3 (grid number = 2) or 0 to 8 (grid number = 3).
\item \textbf{reference category}: \texttt{string}. The category name of the object that is used to reference the target object. 
\item \textbf{reference position}: \texttt{string}. The relative position of the target object from the reference object.
\item \textbf{attribute type}: \texttt{string}. The type of attributes of the target object, currently include: \texttt{color}, \texttt{material}, and \texttt{shape}.
\item \textbf{attribute value}: \texttt{string}. The value of the attributes of the target object.
\end{itemize}

\item {\bf Partitions}.

\begin{itemize}
\item {\bf Partition 1}.
\begin{itemize}
\item {\bf Template}
\begin{itemize}
\item {\bf Q}: What is the object in the <absolute pos> part of the image?
\item {\bf A}: <target category>
\end{itemize}
\item {\bf Example}
\begin{itemize}
\item {\bf Q}: What is the object in the bottom middle part of the image?
\item {\bf A}: folding chair
\end{itemize}
\end{itemize}

\item {\bf Partition 2}.
\begin{itemize}
\item {\bf Template}.
\begin{itemize}
\item {\bf Q}: What is the object <reference pos> the <reference category>?
\item {\bf A}: <target category>
\end{itemize}
\item {\bf Example}
\begin{itemize}
\item {\bf Q}: What is the object to the left of the telephone?
\item {\bf A}: table lamp
\end{itemize}
\end{itemize}

\end{itemize}

\item {\bf Limitations}: The current setup is primarily designed for stationary objects and may not effectively assess dynamic scenarios or human actions, such as interactions with objects or motion-based tasks.

\item {\bf Recommendations}: A task generator includes compositional and contextual challenges that require deeper reasoning about object relation and recognition.
\vspace{-.25em}
\end{itemize}
\end{framed}
\end{figure}

\begin{figure}
\begin{framed}
 \centering
{\Large {\bf WhereGridTaskGenerator}}
\begin{itemize}[leftmargin=*]

\item {\bf Basic Information}. 
\begin{itemize}
\item {\bf Task Type}.  ImageQA
\item {\bf Question Type}.  what object
\item {\bf Answer Type}.  object category
\item {\bf Image Type}. 2D sticker image
\item {\bf The model capability to evaluate}. object recognition with / without reference
\end{itemize}

\item {\bf Source Data}. 
\begin{itemize}
\item rendering images of objects from Objaverse
\item Annotations regarding object category, attribute, and shape
\end{itemize}

\item {\bf Task Plan Schema}.  
\begin{itemize}
\item \textbf{question type}: \texttt{string}. The question type of these tasks will be "what".
\item \textbf{grid number}: \texttt{integer}. The number of diagonal grids of the image, $N$ indicates there are $N\times N$ grids in the image. Support \{2, 3\}.
\item \textbf{target category}: \texttt{string}. The category name of the target object.
\item \textbf{absolute position}: \texttt{string}. The absolute position of the target object in the grid. It is a number ranging from 0 to 3 (grid number = 2) or 0 to 8 (grid number = 3).
\item \textbf{reference category}: \texttt{string}. The category name of the object that is used to reference the target object. 
\item \textbf{reference position}: \texttt{string}. The relative position of the target object from the reference object.
\item \textbf{attribute type}: \texttt{string}. The type of attributes of the target object, currently include: \texttt{color}, \texttt{material}, and \texttt{shape}.
\item \textbf{attribute value}: \texttt{string}. The value of the attributes of the target object.
\end{itemize}

\item {\bf Partitions}.

\begin{itemize}
\item {\bf Partition 1}.
\begin{itemize}
\item {\bf Template}
\begin{itemize}
\item {\bf Q}: Where is the <target category> in the image?
\item {\bf A}: <absolute position>
\end{itemize}
\item {\bf Example}
\begin{itemize}
\item {\bf Q}: Where is the apple in the image? 
\item {\bf A}: back left
\end{itemize}
\end{itemize}

\item {\bf Partition 2}.
\begin{itemize}
\item {\bf Template}.
\begin{itemize}
\item {\bf Q}: Where is the <target category> with respect to the <reference category>? 
\item {\bf A}: <reference position>
\end{itemize}
\item {\bf Example}
\begin{itemize}
\item {\bf Q}: Where is the vacuum cleaner with respect to the backpack?
\item {\bf A}: left
\end{itemize}
\end{itemize}

\end{itemize}

\item {\bf Limitations}: The current setup is primarily designed for stationary objects and may not effectively assess dynamic scenarios or human actions, such as interactions with objects or motion-based tasks.

\item {\bf Recommendations}: A task generator includes compositional and contextual challenges that require deeper reasoning about object relation and recognition.
\vspace{-.25em}
\end{itemize}
\end{framed}
\end{figure}

\begin{figure}
\begin{framed}
 \centering
{\Large {\bf WhatAttributeGridTaskGenerator}}
\begin{itemize}[leftmargin=*]

\item {\bf Basic Information}. 
\begin{itemize}
\item {\bf Task Type}.  ImageQA
\item {\bf Question Type}.  what object
\item {\bf Answer Type}.  object category
\item {\bf Image Type}. 2D sticker image
\item {\bf The model capability to evaluate}. object recognition with / without reference
\end{itemize}

\item {\bf Source Data}. 
\begin{itemize}
\item rendering images of objects from Objaverse
\item Annotations regarding object category, attribute, and shape
\end{itemize}

\item {\bf Task Plan Schema}.  
\begin{itemize}
\item \textbf{question type}: \texttt{string}. The question type of these tasks will be "what attribute".
\item \textbf{grid number}: \texttt{integer}. The number of diagonal grids of the image, $N$ indicates there are $N\times N$ grids in the image. Support \{2, 3\}.
\item \textbf{target category}: \texttt{string}. The category name of the target object.
\item \textbf{absolute position}: \texttt{string}. The absolute position of the target object in the grid. It is a number ranging from 0 to 3 (grid number = 2) or 0 to 8 (grid number = 3).
\item \textbf{reference category}: \texttt{string}. The category name of the object that is used to reference the target object. 
\item \textbf{reference position}: \texttt{string}. The relative position of the target object from the reference object.
\item \textbf{attribute type}: \texttt{string}. The type of attributes of the target object, currently include: \texttt{color}, \texttt{material}, and \texttt{shape}.
\item \textbf{attribute value}: \texttt{string}. The value of the attributes of the target object.
\end{itemize}

\item {\bf Partitions}.

\begin{itemize}
\item {\bf Partition 1}.
\begin{itemize}
\item {\bf Template}
\begin{itemize}
\item {\bf Q}:  What is the <attribute type> of the object in the <absolute position> part of the image?
\item {\bf A}: <attribute value>
\end{itemize}
\item {\bf Example}
\begin{itemize}
\item {\bf Q}: What is the material of the object in the middle part of the image? 
\item {\bf A}: plastic
\end{itemize}
\end{itemize}

\item {\bf Partition 2}.
\begin{itemize}
\item {\bf Template}.
\begin{itemize}
\item {\bf Q}: What is the <attribute type> of the object to the left of the <reference category>?
\item {\bf A}: <attribute value>
\end{itemize}
\item {\bf Example}
\begin{itemize}
\item {\bf Q}: What is the color of the object to the left of the silverware?
\item {\bf A}: gold
\end{itemize}
\end{itemize}

\end{itemize}

\item {\bf Limitations}: The current setup is primarily designed for stationary objects and may not effectively assess dynamic scenarios or human actions, such as interactions with objects or motion-based tasks.

\item {\bf Recommendations}: A task generator includes compositional and contextual challenges that require deeper reasoning about object relation and recognition.
\vspace{-.25em}
\end{itemize}
\end{framed}
\end{figure}

\begin{figure}
\begin{framed}
 \centering
{\Large {\bf WhereAttributeGridTaskGenerator}}
\begin{itemize}[leftmargin=*]

\item {\bf Basic Information}. 
\begin{itemize}
\item {\bf Task Type}.  ImageQA
\item {\bf Question Type}.  what object
\item {\bf Answer Type}.  object category
\item {\bf Image Type}. 2D sticker image
\item {\bf The model capability to evaluate}. object recognition with / without reference
\end{itemize}

\item {\bf Source Data}. 
\begin{itemize}
\item rendering images of objects from Objaverse
\item Annotations regarding object category, attribute, and shape
\end{itemize}

\item {\bf Task Plan Schema}.  
\begin{itemize}
\item \textbf{question type}: \texttt{string}. The question type of these tasks will be "where attribute".
\item \textbf{grid number}: \texttt{integer}. The number of diagonal grids of the image, $N$ indicates there are $N\times N$ grids in the image. Support \{2, 3\}.
\item \textbf{target category}: \texttt{string}. The category name of the target object.
\item \textbf{absolute position}: \texttt{string}. The absolute position of the target object in the grid. It is a number ranging from 0 to 3 (grid number = 2) or 0 to 8 (grid number = 3).
\item \textbf{reference category}: \texttt{string}. The category name of the object that is used to reference the target object. 
\item \textbf{reference position}: \texttt{string}. The relative position of the target object from the reference object.
\item \textbf{attribute type}: \texttt{string}. The type of attributes of the target object, currently include: \texttt{color}, \texttt{material}, and \texttt{shape}.
\item \textbf{attribute value}: \texttt{string}. The value of the attributes of the target object.
\end{itemize}

\item {\bf Partitions}.

\begin{itemize}
\item {\bf Partition 1}.
\begin{itemize}
\item {\bf Template}
\begin{itemize}
\item {\bf Q}: Where is the <attribute value> object in the image?
\item {\bf A}: <absolute position>
\end{itemize}
\item {\bf Example}
\begin{itemize}
\item {\bf Q}: Where is the white object in the image?
\item {\bf A}: top right
\end{itemize}
\end{itemize}

\item {\bf Partition 2}.
\begin{itemize}
\item {\bf Template}.
\begin{itemize}
\item {\bf Q}: Where is the <attribute value> object with respect to the <reference category>? 
\item {\bf A}: <absolute position>
\end{itemize}
\item {\bf Example}
\begin{itemize}
\item {\bf Q}: Where is the gray object with respect to the lollipop? 
\item {\bf A}: top
\end{itemize}
\end{itemize}

\end{itemize}

\item {\bf Limitations}: The current setup is primarily designed for stationary objects and may not effectively assess dynamic scenarios or human actions, such as interactions with objects or motion-based tasks.

\item {\bf Recommendations}: A task generator includes compositional and contextual challenges that require deeper reasoning about object relation and recognition.
\vspace{-.25em}
\end{itemize}
\end{framed}
\end{figure}

\begin{figure}
\begin{framed}
 \centering
{\Large {\bf HowManyGridTaskGenerator}}
\begin{itemize}[leftmargin=*]

\item {\bf Basic Information}. 
\begin{itemize}
\item {\bf Task Type}.  ImageQA
\item {\bf Question Type}.  what object
\item {\bf Answer Type}.  object category
\item {\bf Image Type}. 2D sticker image
\item {\bf The model capability to evaluate}. object recognition with / without reference
\end{itemize}

\item {\bf Source Data}. 
\begin{itemize}
\item rendering images of objects from Objaverse
\item Annotations regarding object category, attribute, and shape
\end{itemize}

\item {\bf Task Plan Schema}.  
\begin{itemize}
\item \textbf{question type}: \texttt{string}. The question type of these tasks will be "how many".
\item \textbf{grid number}: \texttt{integer}. The number of diagonal grids of the image, $N$ indicates there are $N\times N$ grids in the image. Support \{2, 3\}.
\item \textbf{target category}: \texttt{string}. The category name of the target object.
\item \textbf{count} \texttt{integer}. The total number of the target objects in the image.
\item \textbf{attribute type}: \texttt{string}. The type of attributes of the target object, currently include: \texttt{color}, \texttt{material}, and \texttt{shape}.
\item \textbf{attribute value}: \texttt{string}. The value of the attributes of the target object.
\end{itemize}

\item {\bf Partitions}.

\begin{itemize}
\item {\bf Partition 1}.
\begin{itemize}
\item {\bf Template}
\begin{itemize}
\item {\bf Q}: How many <attribute value> objects are there in the image?
\item {\bf A}: <count>
\end{itemize}
\item {\bf Example}
\begin{itemize}
\item {\bf Q}: How many blue objects are there in the image?
\item {\bf A}: 2
\end{itemize}
\end{itemize}

\item {\bf Partition 2}.
\begin{itemize}
\item {\bf Template}.
\begin{itemize}
\item {\bf Q}: How many <target category> are there in the image? 
\item {\bf A}: <count>
\end{itemize}
\item {\bf Example}
\begin{itemize}
\item {\bf Q}: How many tables are there in the image? 
\item {\bf A}: 4
\end{itemize}
\end{itemize}

\item {\bf Partition 3}.
\begin{itemize}
\item {\bf Template}.
\begin{itemize}
\item {\bf Q}: How many <attribute value> <target category> are there in the image? 
\item {\bf A}: <count>
\end{itemize}
\item {\bf Example}
\begin{itemize}
\item {\bf Q}: How many pink beverages are there in the image? 
\item {\bf A}: 2
\end{itemize}
\end{itemize}

\end{itemize}

\item {\bf Limitations}: The current setup is primarily designed for stationary objects and may not effectively assess dynamic scenarios or human actions, such as interactions with objects or motion-based tasks.

\item {\bf Recommendations}: A task generator includes compositional and contextual challenges that require deeper reasoning about object relation and recognition.
\vspace{-.25em}
\end{itemize}
\end{framed}
\end{figure}


\begin{figure}
\begin{framed}
 \centering
{\Large {\bf What3DGridTaskGenerator}}
\begin{itemize}[leftmargin=*]

\item {\bf Basic Information}. 
\begin{itemize}
\item {\bf Task Type}.  ImageQA
\item {\bf Question Type}.  what object
\item {\bf Answer Type}.  object category
\item {\bf Image Type}. 3D tabletop image
\item {\bf The model capability to evaluate}. object recognition with / without reference
\end{itemize}

\item {\bf Source Data}. 
\begin{itemize}
\item rendering images of objects from Objaverse
\item Annotations regarding object category, attribute, and shape
\end{itemize}

\item {\bf Task Plan Schema}.  
\begin{itemize}
\item \textbf{question type}: \texttt{string}. The question type of these tasks will be "what".
\item \textbf{grid number}: \texttt{integer}. The number of diagonal grids of the image, $N$ indicates there are $N\times N$ grids in the image. Support \{2, 3\}.
\item \textbf{target category}: \texttt{string}. The category name of the target object.
\item \textbf{absolute position}: \texttt{string}. The absolute position of the target object in the grid. It is a number ranging from 0 to 3 (grid number = 2) or 0 to 8 (grid number = 3).
\item \textbf{reference category}: \texttt{string}. The category name of the object that is used to reference the target object. 
\item \textbf{reference position}: \texttt{string}. The relative position of the target object from the reference object.
\item \textbf{attribute type}: \texttt{string}. The type of attributes of the target object, currently include: \texttt{color}, \texttt{material}, and \texttt{shape}.
\item \textbf{attribute value}: \texttt{string}. The value of the attributes of the target object.
\end{itemize}

\item {\bf Partitions}.

\begin{itemize}
\item {\bf Partition 1}.
\begin{itemize}
\item {\bf Template}
\begin{itemize}
\item {\bf Q}: What is the object in the <absolute pos> part of the image?
\item {\bf A}: <target category>
\end{itemize}
\item {\bf Example}
\begin{itemize}
\item {\bf Q}: What is the object in the front right part of the image?
\item {\bf A}: scale
\end{itemize}
\end{itemize}

\item {\bf Partition 2}.
\begin{itemize}
\item {\bf Template}.
\begin{itemize}
\item {\bf Q}: What is the object <reference pos> the <reference category>?
\item {\bf A}: <target category>
\end{itemize}
\item {\bf Example}
\begin{itemize}
\item {\bf Q}: What is the object to the right of the mobile computer?
\item {\bf A}: bucket 
\end{itemize}
\end{itemize}

\end{itemize}

\item {\bf Limitations}: The current setup is primarily designed for stationary objects and may not effectively assess dynamic scenarios or human actions, such as interactions with objects or motion-based tasks.

\item {\bf Recommendations}: A task generator includes compositional and contextual challenges that require deeper reasoning about object relation and recognition.
\vspace{-.25em}
\end{itemize}
\end{framed}
\end{figure}

\begin{figure}
\begin{framed}
 \centering
{\Large {\bf Where3DGridTaskGenerator}}
\begin{itemize}[leftmargin=*]

\item {\bf Basic Information}. 
\begin{itemize}
\item {\bf Task Type}.  ImageQA
\item {\bf Question Type}.  what object
\item {\bf Answer Type}.  object category
\item {\bf Image Type}. 3D tabletop image
\item {\bf The model capability to evaluate}. object recognition with / without reference
\end{itemize}

\item {\bf Source Data}. 
\begin{itemize}
\item rendering images of objects from Objaverse
\item Annotations regarding object category, attribute, and shape
\end{itemize}

\item {\bf Task Plan Schema}.  
\begin{itemize}
\item \textbf{question type}: \texttt{string}. The question type of these tasks will be "where".
\item \textbf{grid number}: \texttt{integer}. The number of diagonal grids of the image, $N$ indicates there are $N\times N$ grids in the image. Support \{2, 3\}.
\item \textbf{target category}: \texttt{string}. The category name of the target object.
\item \textbf{absolute position}: \texttt{string}. The absolute position of the target object in the grid. It is a number ranging from 0 to 3 (grid number = 2) or 0 to 8 (grid number = 3).
\item \textbf{reference category}: \texttt{string}. The category name of the object that is used to reference the target object. 
\item \textbf{reference position}: \texttt{string}. The relative position of the target object from the reference object.
\item \textbf{attribute type}: \texttt{string}. The type of attributes of the target object, currently include: \texttt{color}, \texttt{material}, and \texttt{shape}.
\item \textbf{attribute value}: \texttt{string}. The value of the attributes of the target object.
\end{itemize}

\item {\bf Partitions}.

\begin{itemize}
\item {\bf Partition 1}.
\begin{itemize}
\item {\bf Template}
\begin{itemize}
\item {\bf Q}: Where is the <target category> in the image?
\item {\bf A}: <absolute position>
\end{itemize}
\item {\bf Example}
\begin{itemize}
\item {\bf Q}: Where is the vacuum cleaner in the image? 
\item {\bf A}: back left
\end{itemize}
\end{itemize}

\item {\bf Partition 2}.
\begin{itemize}
\item {\bf Template}.
\begin{itemize}
\item {\bf Q}: Where is the <target category> with respect to the <reference category>? 
\item {\bf A}: <reference position>
\end{itemize}
\item {\bf Example}
\begin{itemize}
\item {\bf Q}: Where is the vacuum cleaner with respect to the wine glass?
\item {\bf A}: left
\end{itemize}
\end{itemize}

\end{itemize}

\item {\bf Limitations}: The current setup is primarily designed for stationary objects and may not effectively assess dynamic scenarios or human actions, such as interactions with objects or motion-based tasks.

\item {\bf Recommendations}: A task generator includes compositional and contextual challenges that require deeper reasoning about object relation and recognition.
\vspace{-.25em}
\end{itemize}
\end{framed}
\end{figure}

\begin{figure}
\begin{framed}
 \centering
{\Large {\bf WhatAttribute3DGridTaskGenerator}}
\begin{itemize}[leftmargin=*]

\item {\bf Basic Information}. 
\begin{itemize}
\item {\bf Task Type}.  ImageQA
\item {\bf Question Type}.  what object
\item {\bf Answer Type}.  object category
\item {\bf Image Type}. 3D tabletop image
\item {\bf The model capability to evaluate}. object recognition with / without reference
\end{itemize}

\item {\bf Source Data}. 
\begin{itemize}
\item rendering images of objects from Objaverse
\item Annotations regarding object category, attribute, and shape
\end{itemize}

\item {\bf Task Plan Schema}.  
\begin{itemize}
\item \textbf{question type}: \texttt{string}. The question type of these tasks will be "what attribute".
\item \textbf{grid number}: \texttt{integer}. The number of diagonal grids of the image, $N$ indicates there are $N\times N$ grids in the image. Support \{2, 3\}.
\item \textbf{target category}: \texttt{string}. The category name of the target object.
\item \textbf{absolute position}: \texttt{string}. The absolute position of the target object in the grid. It is a number ranging from 0 to 3 (grid number = 2) or 0 to 8 (grid number = 3).
\item \textbf{reference category}: \texttt{string}. The category name of the object that is used to reference the target object. 
\item \textbf{reference position}: \texttt{string}. The relative position of the target object from the reference object.
\item \textbf{attribute type}: \texttt{string}. The type of attributes of the target object, currently include: \texttt{color}, \texttt{material}, and \texttt{shape}.
\item \textbf{attribute value}: \texttt{string}. The value of the attributes of the target object.
\end{itemize}

\item {\bf Partitions}.

\begin{itemize}
\item {\bf Partition 1}.
\begin{itemize}
\item {\bf Template}
\begin{itemize}
\item {\bf Q}:  What is the <attribute type> of the object in the <absolute position> part of the image?
\item {\bf A}: <attribute value>
\end{itemize}
\item {\bf Example}
\begin{itemize}
\item {\bf Q}: What is the color of the object in the back left part of the image?  
\item {\bf A}: red
\end{itemize}
\end{itemize}

\item {\bf Partition 2}.
\begin{itemize}
\item {\bf Template}.
\begin{itemize}
\item {\bf Q}: What is the <attribute type> of the object to the left of the <reference category>?
\item {\bf A}: <attribute value>
\end{itemize}
\item {\bf Example}
\begin{itemize}
\item {\bf Q}: What is the material of the object behind the plate?
\item {\bf A}: wood
\end{itemize}
\end{itemize}

\end{itemize}

\item {\bf Limitations}: The current setup is primarily designed for stationary objects and may not effectively assess dynamic scenarios or human actions, such as interactions with objects or motion-based tasks.

\item {\bf Recommendations}: A task generator includes compositional and contextual challenges that require deeper reasoning about object relation and recognition.
\vspace{-.25em}
\end{itemize}
\end{framed}
\end{figure}

\begin{figure}
\begin{framed}
 \centering
{\Large {\bf WhereAttribute3DGridTaskGenerator}}
\begin{itemize}[leftmargin=*]

\item {\bf Basic Information}. 
\begin{itemize}
\item {\bf Task Type}.  ImageQA
\item {\bf Question Type}.  what object
\item {\bf Answer Type}.  object category
\item {\bf Image Type}. 3D tabletop image
\item {\bf The model capability to evaluate}. object recognition with / without reference
\end{itemize}

\item {\bf Source Data}. 
\begin{itemize}
\item rendering images of objects from Objaverse
\item Annotations regarding object category, attribute, and shape
\end{itemize}

\item {\bf Task Plan Schema}.  
\begin{itemize}
\item \textbf{question type}: \texttt{string}. The question type of these tasks will be "where attribute".
\item \textbf{grid number}: \texttt{integer}. The number of diagonal grids of the image, $N$ indicates there are $N\times N$ grids in the image. Support \{2, 3\}.
\item \textbf{target category}: \texttt{string}. The category name of the target object.
\item \textbf{absolute position}: \texttt{string}. The absolute position of the target object in the grid. It is a number ranging from 0 to 3 (grid number = 2) or 0 to 8 (grid number = 3).
\item \textbf{reference category}: \texttt{string}. The category name of the object that is used to reference the target object. 
\item \textbf{reference position}: \texttt{string}. The relative position of the target object from the reference object.
\item \textbf{attribute type}: \texttt{string}. The type of attributes of the target object, currently include: \texttt{color}, \texttt{material}, and \texttt{shape}.
\item \textbf{attribute value}: \texttt{string}. The value of the attributes of the target object.
\end{itemize}

\item {\bf Partitions}.

\begin{itemize}
\item {\bf Partition 1}.
\begin{itemize}
\item {\bf Template}
\begin{itemize}
\item {\bf Q}: Where is the <attribute value> object in the image?
\item {\bf A}: <absolute position>
\end{itemize}
\item {\bf Example}
\begin{itemize}
\item {\bf Q}: Where is the wood object in the image?
\item {\bf A}: front right
\end{itemize}
\end{itemize}

\item {\bf Partition 2}.
\begin{itemize}
\item {\bf Template}.
\begin{itemize}
\item {\bf Q}: Where is the <attribute value> object with respect to the <reference category>? 
\item {\bf A}: <absolute position>
\end{itemize}
\item {\bf Example}
\begin{itemize}
\item {\bf Q}: Where is the white object with respect to the trophy? 
\item {\bf A}: left
\end{itemize}
\end{itemize}

\end{itemize}

\item {\bf Limitations}: The current setup is primarily designed for stationary objects and may not effectively assess dynamic scenarios or human actions, such as interactions with objects or motion-based tasks.

\item {\bf Recommendations}: A task generator includes compositional and contextual challenges that require deeper reasoning about object relation and recognition.
\vspace{-.25em}
\end{itemize}
\end{framed}
\end{figure}

\begin{figure}
\begin{framed}
 \centering
{\Large {\bf HowMany3DGridTaskGenerator}}
\begin{itemize}[leftmargin=*]

\item {\bf Basic Information}. 
\begin{itemize}
\item {\bf Task Type}.  ImageQA
\item {\bf Question Type}.  what object
\item {\bf Answer Type}.  object category
\item {\bf Image Type}. 3D tabletop image
\item {\bf The model capability to evaluate}. object recognition with / without reference
\end{itemize}

\item {\bf Source Data}. 
\begin{itemize}
\item rendering images of objects from Objaverse
\item Annotations regarding object category, attribute, and shape
\end{itemize}

\item {\bf Task Plan Schema}.  
\begin{itemize}
\item \textbf{question type}: \texttt{string}. The question type of these tasks will be "how many".
\item \textbf{grid number}: \texttt{integer}. The number of diagonal grids of the image, $N$ indicates there are $N\times N$ grids in the image. Support \{2, 3\}.
\item \textbf{target category}: \texttt{string}. The category name of the target object.
\item \textbf{count} \texttt{integer}. The total number of the target objects in the image.
\item \textbf{attribute type}: \texttt{string}. The type of attributes of the target object, currently include: \texttt{color}, \texttt{material}, and \texttt{shape}.
\item \textbf{attribute value}: \texttt{string}. The value of the attributes of the target object.
\end{itemize}

\item {\bf Partitions}.

\begin{itemize}
\item {\bf Partition 1}.
\begin{itemize}
\item {\bf Template}
\begin{itemize}
\item {\bf Q}: How many <attribute value> objects are there in the image?
\item {\bf A}: <count>
\end{itemize}
\item {\bf Example}
\begin{itemize}
\item {\bf Q}: How many blue objects are there in the image?
\item {\bf A}: 6
\end{itemize}
\end{itemize}

\item {\bf Partition 2}.
\begin{itemize}
\item {\bf Template}.
\begin{itemize}
\item {\bf Q}: How many <target category> are there in the image? 
\item {\bf A}: <count>
\end{itemize}
\item {\bf Example}
\begin{itemize}
\item {\bf Q}: How many plates are there in the image? 
\item {\bf A}: 5
\end{itemize}
\end{itemize}

\item {\bf Partition 3}.
\begin{itemize}
\item {\bf Template}.
\begin{itemize}
\item {\bf Q}: How many <attribute value> <target category> are there in the image? 
\item {\bf A}: <count>
\end{itemize}
\item {\bf Example}
\begin{itemize}
\item {\bf Q}: How many black furnitures are there in the image? 
\item {\bf A}: 4
\end{itemize}
\end{itemize}

\end{itemize}

\item {\bf Limitations}: The current setup is primarily designed for stationary objects and may not effectively assess dynamic scenarios or human actions, such as interactions with objects or motion-based tasks.

\item {\bf Recommendations}: A task generator includes compositional and contextual challenges that require deeper reasoning about object relation and recognition.
\vspace{-.25em}
\end{itemize}
\end{framed}
\end{figure}

\begin{figure}
\begin{framed}
 \centering
{\Large {\bf WhatDistance3DGridTaskGenerator}}
\begin{itemize}[leftmargin=*]

\item {\bf Basic Information}. 
\begin{itemize}
\item {\bf Task Type}.  ImageQA
\item {\bf Question Type}.  what object
\item {\bf Answer Type}.  object category
\item {\bf Image Type}. 3D tabletop image
\item {\bf The model capability to evaluate}. object recognition with / without reference
\end{itemize}

\item {\bf Source Data}. 
\begin{itemize}
\item rendering images of objects from Objaverse
\item Annotations regarding object category, attribute, and shape
\end{itemize}

\item {\bf Task Plan Schema}.  
\begin{itemize}
\item \textbf{question type}: \texttt{string}. The question type of these tasks will be "what distance".
\item \textbf{distance type}: \texttt{string}. The type of the distance between target object and the reference object, indicates whether it pertains to the "farthest" or "closest" distance.
\item \textbf{grid number}: \texttt{integer}. The number of diagonal grids of the image, $N$ indicates there are $N\times N$ grids in the image. Support \{2, 3\}.
\item \textbf{target category}: \texttt{string}. The category name of the target object.
\item \textbf{absolute position}: \texttt{string}. The absolute position of the target object in the grid. It is a number ranging from 0 to 3 (grid number = 2) or 0 to 8 (grid number = 3).
\item \textbf{reference category}: \texttt{string}. The category name of the object that is used to reference the target object. 
\item \textbf{reference position}: \texttt{string}. The relative position of the target object from the reference object.
\item \textbf{attribute type}: \texttt{string}. The type of attributes of the target object, currently include: \texttt{color}, \texttt{material}, and \texttt{shape}.
\item \textbf{attribute value}: \texttt{string}. The value of the attributes of the target object.
\end{itemize}

\item {\bf Partitions}.

\begin{itemize}
\item {\bf Partition 1}.
\begin{itemize}
\item {\bf Template}
\begin{itemize}
\item {\bf Q}: What is the object that is <distance type> from the <reference category>?
\item {\bf A}: <target category>
\end{itemize}
\item {\bf Example}
\begin{itemize}
\item {\bf Q}: What is the object that is farthest from the optical instrument? 
\item {\bf A}: juice
\end{itemize}
\end{itemize}

\end{itemize}

\item {\bf Limitations}: The current setup is primarily designed for stationary objects and may not effectively assess dynamic scenarios or human actions, such as interactions with objects or motion-based tasks.

\item {\bf Recommendations}: A task generator includes compositional and contextual challenges that require deeper reasoning about object relation and recognition.
\vspace{-.25em}
\end{itemize}
\end{framed}
\end{figure}

\begin{figure}
\begin{framed}
 \centering
{\Large {\bf WhereDistance3DGridTaskGenerator}}
\begin{itemize}[leftmargin=*]

\item {\bf Basic Information}. 
\begin{itemize}
\item {\bf Task Type}.  ImageQA
\item {\bf Question Type}.  what object
\item {\bf Answer Type}.  object category
\item {\bf Image Type}. 3D tabletop image
\item {\bf The model capability to evaluate}. object recognition with / without reference
\end{itemize}

\item {\bf Source Data}. 
\begin{itemize}
\item rendering images of objects from Objaverse
\item Annotations regarding object category, attribute, and shape
\end{itemize}

\item {\bf Task Plan Schema}.  
\begin{itemize}
\item \textbf{question type}: \texttt{string}. The question type of these tasks will be "where distance".
\item \textbf{distance type}: \texttt{string}. The type of the distance between target object and the reference object, indicates whether it pertains to the "farthest" or "closest" distance.
\item \textbf{grid number}: \texttt{integer}. The number of diagonal grids of the image, $N$ indicates there are $N\times N$ grids in the image. Support \{2, 3\}.
\item \textbf{target category}: \texttt{string}. The category name of the target object.
\item \textbf{absolute position}: \texttt{string}. The absolute position of the target object in the grid. It is a number ranging from 0 to 3 (grid number = 2) or 0 to 8 (grid number = 3).
\item \textbf{reference category}: \texttt{string}. The category name of the object that is used to reference the target object. 
\item \textbf{reference position}: \texttt{string}. The relative position of the target object from the reference object.
\item \textbf{attribute type}: \texttt{string}. The type of attributes of the target object, currently include: \texttt{color}, \texttt{material}, and \texttt{shape}.
\item \textbf{attribute value}: \texttt{string}. The value of the attributes of the target object.
\end{itemize}

\item {\bf Partitions}.

\begin{itemize}
\item {\bf Partition 1}.
\begin{itemize}
\item {\bf Template}
\begin{itemize}
\item {\bf Q}: Where is the object that is <distance type> from the <reference category> in the image?
\item {\bf A}: <reference position>
\end{itemize}
\item {\bf Example}
\begin{itemize}
\item {\bf Q}: Where is the object that is farthest from the bread in the image?
\item {\bf A}: middle
\end{itemize}
\end{itemize}

\end{itemize}

\item {\bf Limitations}: The current setup is primarily designed for stationary objects and may not effectively assess dynamic scenarios or human actions, such as interactions with objects or motion-based tasks.

\item {\bf Recommendations}: A task generator includes compositional and contextual challenges that require deeper reasoning about object relation and recognition.
\vspace{-.25em}
\end{itemize}
\end{framed}
\end{figure}

\begin{figure}
\begin{framed}
 \centering
{\Large {\bf WhatAttributeDistance3DGridTaskGenerator}}
\begin{itemize}[leftmargin=*]

\item {\bf Basic Information}. 
\begin{itemize}
\item {\bf Task Type}.  ImageQA
\item {\bf Question Type}.  what object
\item {\bf Answer Type}.  object category
\item {\bf Image Type}. 3D tabletop image
\item {\bf The model capability to evaluate}. object recognition with / without reference
\end{itemize}

\item {\bf Source Data}. 
\begin{itemize}
\item rendering images of objects from Objaverse
\item Annotations regarding object category, attribute, and shape
\end{itemize}

\item {\bf Task Plan Schema}.  
\begin{itemize}
\item \textbf{question type}: \texttt{string}. The question type of these tasks will be "what attribute distance".
\item \textbf{distance type}: \texttt{string}. The type of the distance between target object and the reference object, indicates whether it pertains to the "farthest" or "closest" distance.
\item \textbf{grid number}: \texttt{integer}. The number of diagonal grids of the image, $N$ indicates there are $N\times N$ grids in the image. Support \{2, 3\}.
\item \textbf{target category}: \texttt{string}. The category name of the target object.
\item \textbf{absolute position}: \texttt{string}. The absolute position of the target object in the grid. It is a number ranging from 0 to 3 (grid number = 2) or 0 to 8 (grid number = 3).
\item \textbf{reference category}: \texttt{string}. The category name of the object that is used to reference the target object. 
\item \textbf{reference position}: \texttt{string}. The relative position of the target object from the reference object.
\item \textbf{attribute type}: \texttt{string}. The type of attributes of the target object, currently include: \texttt{color}, \texttt{material}, and \texttt{shape}.
\item \textbf{attribute value}: \texttt{string}. The value of the attributes of the target object.
\end{itemize}

\item {\bf Partitions}.

\begin{itemize}
\item {\bf Partition 1}.
\begin{itemize}
\item {\bf Template}
\begin{itemize}
\item {\bf Q}:  What is the <attribute type> of the object that is <distance type> to the <target category>?
\item {\bf A}: <attribute value>
\end{itemize}
\item {\bf Example}
\begin{itemize}
\item {\bf Q}: What is the color of the object that is closest to the statue?
\item {\bf A}: beige
\end{itemize}
\end{itemize}

\end{itemize}

\item {\bf Limitations}: The current setup is primarily designed for stationary objects and may not effectively assess dynamic scenarios or human actions, such as interactions with objects or motion-based tasks.

\item {\bf Recommendations}: A task generator includes compositional and contextual challenges that require deeper reasoning about object relation and recognition.
\vspace{-.25em}
\end{itemize}
\end{framed}
\end{figure}


\begin{figure}
\begin{framed}
 \centering
{\Large {\bf WhatSize3DGridTaskGenerator}}
\begin{itemize}[leftmargin=*]

\item {\bf Basic Information}. 
\begin{itemize}
\item {\bf Task Type}.  ImageQA
\item {\bf Question Type}.  what object
\item {\bf Answer Type}.  object category
\item {\bf Image Type}. 3D tabletop image
\item {\bf The model capability to evaluate}. object recognition with / without reference
\end{itemize}

\item {\bf Source Data}. 
\begin{itemize}
\item rendering images of objects from Objaverse
\item Annotations regarding object category, attribute, and shape
\end{itemize}

\item {\bf Task Plan Schema}.  
\begin{itemize}
\item \textbf{question type}: \texttt{string}. The question type of these tasks will be "what size".
\item \textbf{size}: \texttt{string}. The type of the size of the target object, indicates whether it pertains to the "largest" or "smallest" in all the objects.
\item \textbf{grid number}: \texttt{integer}. The number of diagonal grids of the image, $N$ indicates there are $N\times N$ grids in the image. Support \{2, 3\}.
\item \textbf{target category}: \texttt{string}. The category name of the target object.
\item \textbf{absolute position}: \texttt{string}. The absolute position of the target object in the grid. It is a number ranging from 0 to 3 (grid number = 2) or 0 to 8 (grid number = 3).
\item \textbf{attribute type}: \texttt{string}. The type of attributes of the target object, currently include: \texttt{color}, \texttt{material}, and \texttt{shape}.
\item \textbf{attribute value}: \texttt{string}. The value of the attributes of the target object.
\end{itemize}

\item {\bf Partitions}.

\begin{itemize}
\item {\bf Partition 1}.
\begin{itemize}
\item {\bf Template}
\begin{itemize}
\item {\bf Q}: What is the <size> object in the image? 
\item {\bf A}: <target category>
\end{itemize}
\item {\bf Example}
\begin{itemize}
\item {\bf Q}: What is the smallest object in the image? 
\item {\bf A}: spatula
\end{itemize}
\end{itemize}

\end{itemize}

\item {\bf Limitations}: The current setup is primarily designed for stationary objects and may not effectively assess dynamic scenarios or human actions, such as interactions with objects or motion-based tasks.

\item {\bf Recommendations}: A task generator includes compositional and contextual challenges that require deeper reasoning about object relation and recognition.
\vspace{-.25em}
\end{itemize}
\end{framed}
\end{figure}

\begin{figure}
\begin{framed}
 \centering
{\Large {\bf WhereSize3DGridTaskGenerator}}
\begin{itemize}[leftmargin=*]

\item {\bf Basic Information}. 
\begin{itemize}
\item {\bf Task Type}.  ImageQA
\item {\bf Question Type}.  what object
\item {\bf Answer Type}.  object category
\item {\bf Image Type}. 3D tabletop image
\item {\bf The model capability to evaluate}. object recognition with / without reference
\end{itemize}

\item {\bf Source Data}. 
\begin{itemize}
\item rendering images of objects from Objaverse
\item Annotations regarding object category, attribute, and shape
\end{itemize}

\item {\bf Task Plan Schema}.  
\begin{itemize}
\item \textbf{question type}: \texttt{string}. The question type of these tasks will be "where size".
\item \textbf{size}: \texttt{string}. The type of the size of the target object, indicates whether it pertains to the "largest" or "smallest" in all the objects.
\item \textbf{grid number}: \texttt{integer}. The number of diagonal grids of the image, $N$ indicates there are $N\times N$ grids in the image. Support \{2, 3\}.
\item \textbf{target category}: \texttt{string}. The category name of the target object.
\item \textbf{absolute position}: \texttt{string}. The absolute position of the target object in the grid. It is a number ranging from 0 to 3 (grid number = 2) or 0 to 8 (grid number = 3).
\item \textbf{reference category}: \texttt{string}. The category name of the object that is used to reference the target object. 
\item \textbf{reference position}: \texttt{string}. The relative position of the target object from the reference object.
\item \textbf{attribute type}: \texttt{string}. The type of attributes of the target object, currently include: \texttt{color}, \texttt{material}, and \texttt{shape}.
\item \textbf{attribute value}: \texttt{string}. The value of the attributes of the target object.
\item \textbf{target-reference order}: \texttt{string}. Define the target object goes first or not in the question. It is related to grammar
\end{itemize}

\item {\bf Partitions}.

\begin{itemize}
\item {\bf Partition 1}.
\begin{itemize}
\item {\bf Template}
\begin{itemize}
\item {\bf Q}: Where is the <size> object in the image?
\item {\bf A}: <absolute position>
\end{itemize}
\item {\bf Example}
\begin{itemize}
\item {\bf Q}: Where is the largest object in the image? 
\item {\bf A}: middle
\end{itemize}
\end{itemize}

\item {\bf Partition 2}.
\begin{itemize}
\item {\bf Template}
\begin{itemize}
\item {\bf Q}: Where is the <size> object in the image with respect to the <reference category>?
\item {\bf A}: <reference position>
\end{itemize}
\item {\bf Example}
\begin{itemize}
\item {\bf Q}: Where is the smallest object in the image with respect to the car?
\item {\bf A}: middle
\end{itemize}
\end{itemize}

\end{itemize}

\item {\bf Limitations}: The current setup is primarily designed for stationary objects and may not effectively assess dynamic scenarios or human actions, such as interactions with objects or motion-based tasks.

\item {\bf Recommendations}: A task generator includes compositional and contextual challenges that require deeper reasoning about object relation and recognition.
\vspace{-.25em}
\end{itemize}
\end{framed}
\end{figure}

\begin{figure}
\begin{framed}
 \centering
{\Large {\bf WhatAttributeSize3DGridTaskGenerator}}
\begin{itemize}[leftmargin=*]

\item {\bf Basic Information}. 
\begin{itemize}
\item {\bf Task Type}.  ImageQA
\item {\bf Question Type}.  what object
\item {\bf Answer Type}.  object category
\item {\bf Image Type}. 3D tabletop image
\item {\bf The model capability to evaluate}. object recognition with / without reference
\end{itemize}

\item {\bf Source Data}. 
\begin{itemize}
\item rendering images of objects from Objaverse
\item Annotations regarding object category, attribute, and shape
\end{itemize}

\item {\bf Task Plan Schema}.  
\begin{itemize}
\item \textbf{question type}: \texttt{string}. The question type of these tasks will be "what attribute size".
\item \textbf{size}: \texttt{string}. The type of the size of the target object, indicates whether it pertains to the "largest" or "smallest" in all the objects.
\item \textbf{grid number}: \texttt{integer}. The number of diagonal grids of the image, $N$ indicates there are $N\times N$ grids in the image. Support \{2, 3\}.
\item \textbf{target category}: \texttt{string}. The category name of the target object.
\item \textbf{absolute position}: \texttt{string}. The absolute position of the target object in the grid. It is a number ranging from 0 to 3 (grid number = 2) or 0 to 8 (grid number = 3).
\item \textbf{attribute type}: \texttt{string}. The type of attributes of the target object, currently include: \texttt{color}, \texttt{material}, and \texttt{shape}.
\item \textbf{attribute value}: \texttt{string}. The value of the attributes of the target object.

\end{itemize}

\item {\bf Partitions}.

\begin{itemize}
\item {\bf Partition 1}.
\begin{itemize}
\item {\bf Template}
\begin{itemize}
\item {\bf Q}:  What is the <attribute type>  of the <size> object in the image?
\item {\bf A}: <attribute value>
\end{itemize}
\item {\bf Example}
\begin{itemize}
\item {\bf Q}: What is the color of the smallest object in the image?
\item {\bf A}: black
\end{itemize}
\end{itemize}

\end{itemize}

\item {\bf Limitations}: The current setup is primarily designed for stationary objects and may not effectively assess dynamic scenarios or human actions, such as interactions with objects or motion-based tasks.

\item {\bf Recommendations}: A task generator includes compositional and contextual challenges that require deeper reasoning about object relation and recognition.
\vspace{-.25em}
\end{itemize}
\end{framed}
\end{figure}


\begin{figure}
\begin{framed}
 \centering
{\Large {\bf WhatMovementVideoGridTaskGenerator}}
\begin{itemize}[leftmargin=*]

\item {\bf Basic Information}. 
\begin{itemize}
\item {\bf Task Type}.  VideoQA
\item {\bf Question Type}.  what object
\item {\bf Answer Type}.  object category
\item {\bf Image Type}. 3D tabletop video
\item {\bf The model capability to evaluate}. object recognition with / without reference
\end{itemize}

\item {\bf Source Data}. 
\begin{itemize}
\item rendering images of objects from Objaverse
\item Annotations regarding object category, attribute, and shape
\end{itemize}

\item {\bf Task Plan Schema}.  
\begin{itemize}
\item \textbf{question type}: \texttt{string}. The question type of these tasks will be "what move video".
\item \textbf{grid number}: \texttt{integer}. The number of diagonal grids of the image, $N$ indicates there are $N\times N$ grids in the image. Support \{2, 3\}.
\item \textbf{target category}: \texttt{string}. The category name of the target object.
\item \textbf{absolute position}: \texttt{string}. The absolute position of the target object in the grid. It is a number ranging from 0 to 3 (grid number = 2) or 0 to 8 (grid number = 3).
\item \textbf{attribute type}: \texttt{string}. The type of attributes of the target object, currently include: \texttt{color}, \texttt{material}, and \texttt{shape}.
\item \textbf{attribute value}: \texttt{string}. The value of the attributes of the target object.
\item \textbf{moving direction}: \texttt{string}. The moving direction of the target object, can be either 'left', 'right', 'up', or 'down'.
\item \textbf{are other objects moving}: \texttt{string}. Indicates that other objects in the video are moving or not, can be "Yes" or "No". If it is "Yes" moving, it should not be in the same direction of the target object's moving direction.
\end{itemize}

\item {\bf Partitions}.

\begin{itemize}
\item {\bf Partition 1}.
\begin{itemize}
\item {\bf Template}
\begin{itemize}
\item {\bf Q}: What is the object that is moving <moving direction> in the video?
\item {\bf A}: <target category>
\end{itemize}
\item {\bf Example}
\begin{itemize}
\item {\bf Q}: What is the object that is moving left in the video?
\item {\bf A}: serving tray
\end{itemize}
\end{itemize}

\item {\bf Partition 2}.
\begin{itemize}
\item {\bf Template}
\begin{itemize}
\item {\bf Q}: What is the moving object in the video? 
\item {\bf A}: <target category>
\end{itemize}
\item {\bf Example}
\begin{itemize}
\item {\bf Q}: What is the moving object in the video?
\item {\bf A}: barrel
\end{itemize}
\end{itemize}

\end{itemize}

\item {\bf Limitations}: The current setup is primarily designed for stationary objects and may not effectively assess dynamic scenarios or human actions, such as interactions with objects or motion-based tasks.

\item {\bf Recommendations}: A task generator includes compositional and contextual challenges that require deeper reasoning about object relation and recognition.
\vspace{-.25em}
\end{itemize}
\end{framed}
\end{figure}

\begin{figure}
\begin{framed}
 \centering
{\Large {\bf WhereMovementVideoGridTaskGenerator}}
\begin{itemize}[leftmargin=*]

\item {\bf Basic Information}. 
\begin{itemize}
\item {\bf Task Type}.  VideoQA
\item {\bf Question Type}.  what object
\item {\bf Answer Type}.  object category
\item {\bf Image Type}. 3D tabletop video
\item {\bf The model capability to evaluate}. object recognition with / without reference
\end{itemize}

\item {\bf Source Data}. 
\begin{itemize}
\item rendering images of objects from Objaverse
\item Annotations regarding object category, attribute, and shape
\end{itemize}

\item {\bf Task Plan Schema}.  
\begin{itemize}
\item \textbf{question type}: \texttt{string}. The question type of these tasks will be "where move video".
\item \textbf{grid number}: \texttt{integer}. The number of diagonal grids of the image, $N$ indicates there are $N\times N$ grids in the image. Support \{2, 3\}.
\item \textbf{target category}: \texttt{string}. The category name of the target object.
\item \textbf{absolute position}: \texttt{string}. The absolute position of the target object in the grid. It is a number ranging from 0 to 3 (grid number = 2) or 0 to 8 (grid number = 3).
\item \textbf{attribute type}: \texttt{string}. The type of attributes of the target object, currently include: \texttt{color}, \texttt{material}, and \texttt{shape}.
\item \textbf{attribute value}: \texttt{string}. The value of the attributes of the target object.
\item \textbf{moving direction}: \texttt{string}. The moving direction of the target object, can be either 'left', 'right', 'up', or 'down'.
\item \textbf{are other objects moving}: \texttt{string}. Indicates that other objects in the video are moving or not, can be "Yes" or "No". If it is "Yes" moving, it should not be in the same direction of the target object's moving direction.
\end{itemize}

\item {\bf Partitions}. 

\begin{itemize}
\item {\bf Partition 1}.
\begin{itemize}
\item {\bf Template}
\begin{itemize}
\item {\bf Q}:  Where is the object that is moving down located in the video? 
\item {\bf A}: <absolute position>
\end{itemize}
\item {\bf Example}
\begin{itemize}
\item {\bf Q}: Where is the object that is moving down located in the video? 
\item {\bf A}: back right
\end{itemize}
\end{itemize}

\item {\bf Partition 2}.
\begin{itemize}
\item {\bf Template}
\begin{itemize}
\item {\bf Q}:  Where is the moving object located in the video?
\item {\bf A}: <absolute position>
\end{itemize}
\item {\bf Example}
\begin{itemize}
\item {\bf Q}: Where is the moving object located in the video?
\item {\bf A}: back right
\end{itemize}
\end{itemize}

\end{itemize}

\item {\bf Limitations}: The current setup is primarily designed for stationary objects and may not effectively assess dynamic scenarios or human actions, such as interactions with objects or motion-based tasks.

\item {\bf Recommendations}: A task generator includes compositional and contextual challenges that require deeper reasoning about object relation and recognition.
\vspace{-.25em}
\end{itemize}
\end{framed}
\end{figure}

\begin{figure}
\begin{framed}
 \centering
{\Large {\bf WhatAttributeMovementVideoGridTaskGenerator}}
\begin{itemize}[leftmargin=*]

\item {\bf Basic Information}. 
\begin{itemize}
\item {\bf Task Type}.  VideoQA
\item {\bf Question Type}.  what object
\item {\bf Answer Type}.  object category
\item {\bf Image Type}. 3D tabletop video
\item {\bf The model capability to evaluate}. object recognition with / without reference
\end{itemize}

\item {\bf Source Data}. 
\begin{itemize}
\item rendering images of objects from Objaverse
\item Annotations regarding object category, attribute, and shape
\end{itemize}

\item {\bf Task Plan Schema}.  
\begin{itemize}
\item \textbf{question type}: \texttt{string}. The question type of these tasks will be "what attribute move video".
\item \textbf{size}: \texttt{string}. The type of the size of the target object, indicates whether it pertains to the "largest" or "smallest" in all the objects.
\item \textbf{grid number}: \texttt{integer}. The number of diagonal grids of the image, $N$ indicates there are $N\times N$ grids in the image. Support \{2, 3\}.
\item \textbf{target category}: \texttt{string}. The category name of the target object.
\item \textbf{absolute position}: \texttt{string}. The absolute position of the target object in the grid. It is a number ranging from 0 to 3 (grid number = 2) or 0 to 8 (grid number = 3).
\item \textbf{attribute type}: \texttt{string}. The type of attributes of the target object, currently include: \texttt{color}, \texttt{material}, and \texttt{shape}.
\item \textbf{attribute value}: \texttt{string}. The value of the attributes of the target object.

\end{itemize}

\item {\bf Partitions}.

\begin{itemize}
\item {\bf Partition 1}.
\begin{itemize}
\item {\bf Template}
\begin{itemize}
\item {\bf Q}: What is the <attribute type> of the object that is moving <moving direction> in the video? 
\item {\bf A}: <attribute value>
\end{itemize}
\item {\bf Example}
\begin{itemize}
\item {\bf Q}: What is the color of the object that is moving left in the video?  
\item {\bf A}: black
\end{itemize}
\end{itemize}

\item {\bf Partition 2}.
\begin{itemize}
\item {\bf Template}
\begin{itemize}
\item {\bf Q}: Where is the <attribute type> of the moving object in the video?
\item {\bf A}: <attribute value>
\end{itemize}
\item {\bf Example}
\begin{itemize}
\item {\bf Q}: What is the color of the moving object in the video? 
\item {\bf A}: white
\end{itemize}
\end{itemize}

\end{itemize}

\item {\bf Limitations}: The current setup is primarily designed for stationary objects and may not effectively assess dynamic scenarios or human actions, such as interactions with objects or motion-based tasks.

\item {\bf Recommendations}: A task generator includes compositional and contextual challenges that require deeper reasoning about object relation and recognition.
\vspace{-.25em}
\end{itemize}
\end{framed}
\end{figure}

\begin{figure}
\begin{framed}
 \centering
{\Large {\bf WhatRotationVideoGridTaskGenerator}}
\begin{itemize}[leftmargin=*]

\item {\bf Basic Information}. 
\begin{itemize}
\item {\bf Task Type}.  VideoQA
\item {\bf Question Type}.  what object
\item {\bf Answer Type}.  object category
\item {\bf Image Type}. 3D tabletop video
\item {\bf The model capability to evaluate}. object recognition with / without reference
\end{itemize}

\item {\bf Source Data}. 
\begin{itemize}
\item rendering images of objects from Objaverse
\item Annotations regarding object category, attribute, and shape
\end{itemize}

\item {\bf Task Plan Schema}.  
\begin{itemize}
\item \textbf{question type}: \texttt{string}. The question type of these tasks will be "what rotate video".
\item \textbf{size}: \texttt{string}. The type of the size of the target object, indicates whether it pertains to the "largest" or "smallest" in all the objects.
\item \textbf{grid number}: \texttt{integer}. The number of diagonal grids of the image, $N$ indicates there are $N\times N$ grids in the image. Support \{2, 3\}.
\item \textbf{target category}: \texttt{string}. The category name of the target object.
\item \textbf{absolute position}: \texttt{string}. The absolute position of the target object in the grid. It is a number ranging from 0 to 3 (grid number = 2) or 0 to 8 (grid number = 3).
\item \textbf{attribute type}: \texttt{string}. The type of attributes of the target object, currently include: \texttt{color}, \texttt{material}, and \texttt{shape}.
\item \textbf{attribute value}: \texttt{string}. The value of the attributes of the target object.
\end{itemize}

\item {\bf Partitions}.

\begin{itemize}
\item {\bf Partition 1}.
\begin{itemize}
\item {\bf Template}
\begin{itemize}
\item {\bf Q}: What is the <size> object in the image? 
\item {\bf A}: <target category>
\end{itemize}
\item {\bf Example}
\begin{itemize}
\item {\bf Q}: What is the smallest object in the image? 
\item {\bf A}: spatula
\end{itemize}
\end{itemize}

\end{itemize}

\item {\bf Limitations}: The current setup is primarily designed for stationary objects and may not effectively assess dynamic scenarios or human actions, such as interactions with objects or motion-based tasks.

\item {\bf Recommendations}: A task generator includes compositional and contextual challenges that require deeper reasoning about object relation and recognition.
\vspace{-.25em}
\end{itemize}
\end{framed}
\end{figure}

\begin{figure}
\begin{framed}
 \centering
{\Large {\bf WhereRotationVideoGridTaskGenerator}}
\begin{itemize}[leftmargin=*]

\item {\bf Basic Information}. 
\begin{itemize}
\item {\bf Task Type}.  VideoQA
\item {\bf Question Type}.  what object
\item {\bf Answer Type}.  object category
\item {\bf Image Type}. 3D tabletop video
\item {\bf The model capability to evaluate}. object recognition with / without reference
\end{itemize}

\item {\bf Source Data}. 
\begin{itemize}
\item rendering images of objects from Objaverse
\item Annotations regarding object category, attribute, and shape
\end{itemize}

\item {\bf Task Plan Schema}.  
\begin{itemize}
\item \textbf{question type}: \texttt{string}. The question type of these tasks will be "where rotate video".
\item \textbf{size}: \texttt{string}. The type of the size of the target object, indicates whether it pertains to the "largest" or "smallest" in all the objects.
\item \textbf{grid number}: \texttt{integer}. The number of diagonal grids of the image, $N$ indicates there are $N\times N$ grids in the image. Support \{2, 3\}.
\item \textbf{target category}: \texttt{string}. The category name of the target object.
\item \textbf{absolute position}: \texttt{string}. The absolute position of the target object in the grid. It is a number ranging from 0 to 3 (grid number = 2) or 0 to 8 (grid number = 3).
\item \textbf{reference category}: \texttt{string}. The category name of the object that is used to reference the target object. 
\item \textbf{reference position}: \texttt{string}. The relative position of the target object from the reference object.
\item \textbf{attribute type}: \texttt{string}. The type of attributes of the target object, currently include: \texttt{color}, \texttt{material}, and \texttt{shape}.
\item \textbf{attribute value}: \texttt{string}. The value of the attributes of the target object.
\item \textbf{target-reference order}: \texttt{string}. Define the target object goes first or not in the question. It is related to grammar
\end{itemize}

\item {\bf Partitions}.

\begin{itemize}
\item {\bf Partition 1}.
\begin{itemize}
\item {\bf Template}
\begin{itemize}
\item {\bf Q}: Where is the <size> object in the image?
\item {\bf A}: <absolute position>
\end{itemize}
\item {\bf Example}
\begin{itemize}
\item {\bf Q}: Where is the largest object in the image? 
\item {\bf A}: middle
\end{itemize}
\end{itemize}

\item {\bf Partition 2}.
\begin{itemize}
\item {\bf Template}
\begin{itemize}
\item {\bf Q}: Where is the <size> object in the image with respect to the <reference category>?
\item {\bf A}: <reference position>
\end{itemize}
\item {\bf Example}
\begin{itemize}
\item {\bf Q}: Where is the smallest object in the image with respect to the car?
\item {\bf A}: middle
\end{itemize}
\end{itemize}

\end{itemize}

\item {\bf Limitations}: The current setup is primarily designed for stationary objects and may not effectively assess dynamic scenarios or human actions, such as interactions with objects or motion-based tasks.

\item {\bf Recommendations}: A task generator includes compositional and contextual challenges that require deeper reasoning about object relation and recognition.
\vspace{-.25em}
\end{itemize}
\end{framed}
\end{figure}

\begin{figure}
\begin{framed}
 \centering
{\Large {\bf WhatAttributeRotationVideoGridTaaskGenerator}}
\begin{itemize}[leftmargin=*]

\item {\bf Basic Information}. 
\begin{itemize}
\item {\bf Task Type}.  VideoQA
\item {\bf Question Type}.  what object
\item {\bf Answer Type}.  object category
\item {\bf Image Type}. 3D tabletop video
\item {\bf The model capability to evaluate}. object recognition with / without reference
\end{itemize}

\item {\bf Source Data}. 
\begin{itemize}
\item rendering images of objects from Objaverse
\item Annotations regarding object category, attribute, and shape
\end{itemize}

\item {\bf Task Plan Schema}.  
\begin{itemize}
\item \textbf{question type}: \texttt{string}. The question type of these tasks will be "what attribute rotate video".
\item \textbf{size}: \texttt{string}. The type of the size of the target object, indicates whether it pertains to the "largest" or "smallest" in all the objects.
\item \textbf{grid number}: \texttt{integer}. The number of diagonal grids of the image, $N$ indicates there are $N\times N$ grids in the image. Support \{2, 3\}.
\item \textbf{target category}: \texttt{string}. The category name of the target object.
\item \textbf{absolute position}: \texttt{string}. The absolute position of the target object in the grid. It is a number ranging from 0 to 3 (grid number = 2) or 0 to 8 (grid number = 3).
\item \textbf{attribute type}: \texttt{string}. The type of attributes of the target object, currently include: \texttt{color}, \texttt{material}, and \texttt{shape}.
\item \textbf{attribute value}: \texttt{string}. The value of the attributes of the target object.

\end{itemize}

\item {\bf Partitions}.

\begin{itemize}
\item {\bf Partition 1}.
\begin{itemize}
\item {\bf Template}
\begin{itemize}
\item {\bf Q}:  What is the <attribute type>  of the <size> object in the image?
\item {\bf A}: <attribute value>
\end{itemize}
\item {\bf Example}
\begin{itemize}
\item {\bf Q}: What is the color of the smallest object in the image?
\item {\bf A}: black
\end{itemize}
\end{itemize}

\end{itemize}

\item {\bf Limitations}: The current setup is primarily designed for stationary objects and may not effectively assess dynamic scenarios or human actions, such as interactions with objects or motion-based tasks.

\item {\bf Recommendations}: A task generator includes compositional and contextual challenges that require deeper reasoning about object relation and recognition.
\vspace{-.25em}
\end{itemize}
\end{framed}
\end{figure}


\begin{figure}
\begin{framed}
 \centering
{\Large {\bf WhatObjectSceneGraphTaskGenerator}}
\begin{itemize}[leftmargin=*]

\item {\bf Basic Information}. 
\begin{itemize}
\item {\bf Task Type}.  ImageQA
\item {\bf Question Type}.  what object
\item {\bf Answer Type}.  object category
\item {\bf Image Type}. 3D tabletop image
\item {\bf The model capability to evaluate}. object recognition with / without reference
\end{itemize}

\item {\bf Source Data}. 
\begin{itemize}
\item rendering images of objects from Objaverse
\item Annotations regarding object category, attribute, and shape
\end{itemize}

\item {\bf Task Plan Schema}.  
\begin{itemize}
\item \textbf{question type}: \texttt{string}. The question type of these tasks will be "what object".
\item \textbf{object}        : \texttt{string}. The target object node of the question.
\item \textbf{subgraph}      : \texttt{string}. The subgraph with the target object node as its root, used to reference the target object node.
\item \textbf{scene graph id} : \texttt{string}. The identifier of the scene graph. 
\item \textbf{answers}: \texttt{list}. A list of object nodes in the scene graph that share the same subgraph structure, except the target object node and itself.
\end{itemize}

\item {\bf Partitions}.

\begin{itemize}
\item {\bf Partition 1}.
\begin{itemize}
\item {\bf Template}
\begin{itemize}
\item {\bf Q}: What is the <object and its attributes in the subgraph> that <obj reference(other reference objects, attributes, and relations in the subgraph)>?
\item {\bf A}: <target category>
\end{itemize}
\item {\bf Example}
\begin{itemize}
\item {\bf Q}: What is the flat object that is on the brown and wood table?  
\item {\bf A}: paper
\end{itemize}
\end{itemize}

\end{itemize}

\item {\bf Limitations}: The current setup is primarily designed for stationary objects and may not effectively assess dynamic scenarios or human actions, such as interactions with objects or motion-based tasks.

\item {\bf Recommendations}: A task generator includes compositional and contextual challenges that require deeper reasoning about object relation and recognition.
\vspace{-.25em}
\end{itemize}
\end{framed}
\end{figure}

\begin{figure}
\begin{framed}
 \centering
{\Large {\bf WhatAttributeSceneGraphTaskGenerator}}
\begin{itemize}[leftmargin=*]

\item {\bf Basic Information}. 
\begin{itemize}
\item {\bf Task Type}.  ImageQA
\item {\bf Question Type}.  what object
\item {\bf Answer Type}.  object category
\item {\bf Image Type}. 3D tabletop image
\item {\bf The model capability to evaluate}. object recognition with / without reference
\end{itemize}

\item {\bf Source Data}. 
\begin{itemize}
\item rendering images of objects from Objaverse
\item Annotations regarding object category, attribute, and shape
\end{itemize}

\item {\bf Task Plan Schema}.  
\begin{itemize}
\item \textbf{question type}: \texttt{string}. The question type of these tasks will be "what attribute".
\item \textbf{attribute type}        : \texttt{string}. The type of the target attribute.
\item \textbf{attribute}        : \texttt{string}. The target attribute node of the question.
\item \textbf{subgraph}      : \texttt{string}. The subgraph with the target attribute node as its root.
\item \textbf{scene graph id} : \texttt{string}. The identifier of the scene graph. 
\item \textbf{answers}: \texttt{list}. A list of attribute nodes in the scene graph that share the same subgraph structure, except the target attribute node and itself.
\end{itemize}

\item {\bf Partitions}.

\begin{itemize}
\item {\bf Partition 1}.
\begin{itemize}
\item {\bf Template}
\begin{itemize}
\item {\bf Q}: What is the <attribute type> of the <target attribute's corresponding object and object's other attributes in the subgraph> that <obj reference(other reference objects, attributes, and relations in the subgraph)>?
\item {\bf A}: <attribute>
\end{itemize}
\item {\bf Example}
\begin{itemize}
\item {\bf Q}: What is the material of the smooth object that is to the right of the yellow container?  
\item {\bf A}: plastic
\end{itemize}
\end{itemize}

\end{itemize}

\item {\bf Limitations}: The current setup is primarily designed for stationary objects and may not effectively assess dynamic scenarios or human actions, such as interactions with objects or motion-based tasks.

\item {\bf Recommendations}: A task generator includes compositional and contextual challenges that require deeper reasoning about object relation and recognition.
\vspace{-.25em}
\end{itemize}
\end{framed}
\end{figure}

\begin{figure}
\begin{framed}
 \centering
{\Large {\bf WhatRelationSceneGraphTaskGenerator}}
\begin{itemize}[leftmargin=*]

\item {\bf Basic Information}. 
\begin{itemize}
\item {\bf Task Type}.  ImageQA
\item {\bf Question Type}.  what object
\item {\bf Answer Type}.  object category
\item {\bf Image Type}. 3D tabletop image
\item {\bf The model capability to evaluate}. object recognition with / without reference
\end{itemize}

\item {\bf Source Data}. 
\begin{itemize}
\item rendering images of objects from Objaverse
\item Annotations regarding object category, attribute, and shape
\end{itemize}

\item {\bf Task Plan Schema}.  
\begin{itemize}
\item \textbf{question type}: \texttt{string}. The question type of these tasks will be "what relation".
\item \textbf{relation}: \texttt{string}. The target relation edge between source object node and target object node
\item \textbf{source object}: \texttt{string}. The source object node of the question.
\item \textbf{target object}        : \texttt{string}. The target object node of the question.
\item \textbf{source subgraph}      : \texttt{string}. The subgraph with the source object node as its root.
\item \textbf{target subgraph}      : \texttt{string}. The subgraph with the target object node as its root.
\item \textbf{scene graph id} : \texttt{string}. The identifier of the scene graph. 
\item \textbf{answers}: \texttt{list}. A list of relation edges in the scene graph that connect the same source subgraph and target subgraph.
\end{itemize}

\item {\bf Partitions}.

\begin{itemize}
\item {\bf Partition 1}.
\begin{itemize}
\item {\bf Template}
\begin{itemize}
\item {\bf Q}:  What is the relation from the <source object's attributes in the source subgraph> object, which <source obj reference(other reference objects, attributes, and relations in the source subgraph)>, to the <target object's attributes in the source subgraph> object, which <target obj reference(other reference objects, attributes, and relations in the target subgraph)>?
\item {\bf A}: <relation>
\end{itemize}
\item {\bf Example}
\begin{itemize}
\item {\bf Q}: What is the relation from the standing object, which the colorful and long snowboard is to the right of,
to the blue and long object, which is to the left of the patterned skis?
\item {\bf A}: holding
\end{itemize}
\end{itemize}

\end{itemize}

\item {\bf Limitations}: The current setup is primarily designed for stationary objects and may not effectively assess dynamic scenarios or human actions, such as interactions with objects or motion-based tasks.

\item {\bf Recommendations}: A task generator includes compositional and contextual challenges that require deeper reasoning about object relation and recognition.
\vspace{-.25em}
\end{itemize}
\end{framed}
\end{figure}


\begin{figure}
\begin{framed}
 \centering
{\Large {\bf WhatObjectVideoSceneGraphTaskGenerator}}
\begin{itemize}[leftmargin=*]

\item {\bf Basic Information}. 
\begin{itemize}
\item {\bf Task Type}.  VideoQA
\item {\bf Question Type}.  what object
\item {\bf Answer Type}.  object category
\item {\bf Image Type}. 3D tabletop image
\item {\bf The model capability to evaluate}. object recognition with / without reference
\end{itemize}

\item {\bf Source Data}. 
\begin{itemize}
\item rendering images of objects from Objaverse
\item Annotations regarding object category, attribute, and shape
\end{itemize}

\item {\bf Task Plan Schema}.  
\begin{itemize}
\item \textbf{question type}: \texttt{string}. The question type of these tasks will be "what object video".
\item \textbf{object}        : \texttt{string}. The target object the person in the video interacts with.
\item \textbf{relation}        : \texttt{string}. The relation between the person and the target object it interacts with.
\item \textbf{reference action}        : \texttt{string}. The reference action to locate the moment when a person is interacting with the target object.
\item \textbf{reference type}        : \texttt{string}. The target object of the relation between the person and the target object it interacts with, can be "spatial" or "contact"
\item \textbf{temporal reference type}      : \texttt{string}. Type of the temporal reference between the reference action and the moment when a person is interacting with the target object. Can be "before", "while", or "after"
\item \textbf{video scene graph id} : \texttt{string}. The identifier of the video scene graph. 
\end{itemize}

\item {\bf Partitions}.

\begin{itemize}
\item {\bf Partition 1}.
\begin{itemize}
\item {\bf Template}
\begin{itemize}
\item {\bf Q}: What is the object that the person is <reference> <temporal reference type> the person <reference action>?
\item {\bf A}: <object>
\end{itemize}
\item {\bf Example}
\begin{itemize}
\item {\bf Q}: What is the object that the person is behind after the person watching something in a mirror?
\item {\bf A}: floor
\end{itemize}
\end{itemize}

\end{itemize}

\item {\bf Limitations}: The current setup is primarily designed for stationary objects and may not effectively assess dynamic scenarios or human actions, such as interactions with objects or motion-based tasks.

\item {\bf Recommendations}: A task generator includes compositional and contextual challenges that require deeper reasoning about object relation and recognition.
\vspace{-.25em}
\end{itemize}
\end{framed}
\end{figure}

\begin{figure}
\begin{framed}
 \centering
{\Large {\bf WhatRelationVideoSceneGraphTaskGenerator}}
\begin{itemize}[leftmargin=*]

\item {\bf Basic Information}. 
\begin{itemize}
\item {\bf Task Type}.  VideoQA
\item {\bf Question Type}.  what object
\item {\bf Answer Type}.  object category
\item {\bf Image Type}. 3D tabletop image
\item {\bf The model capability to evaluate}. object recognition with / without reference
\end{itemize}

\item {\bf Source Data}. 
\begin{itemize}
\item rendering images of objects from Objaverse
\item Annotations regarding object category, attribute, and shape
\end{itemize}

\item {\bf Task Plan Schema}.  
\begin{itemize}
\item \textbf{question type}: \texttt{string}. The question type of these tasks will be "what relation video".
\item \textbf{object}        : \texttt{string}. The object the person in the video interacts by the target relation.
\item \textbf{relation}        : \texttt{string}. The target relation between the person and the target object it interacts with.
\item \textbf{reference action}        : \texttt{string}. The reference action to locate the moment when a person is interacting with the object.
\item \textbf{reference type}        : \texttt{string}. The type of the target relation between the person and the object it interacts with, can be "spatial" or "contact"
\item \textbf{temporal reference type}      : \texttt{string}. Type of the temporal reference between the reference action and the moment when a person is interacting with the object. Can be "before", "while", or "after"
\item \textbf{video scene graph id} : \texttt{string}. The identifier of the video scene graph. 
\end{itemize}

\item {\bf Partitions}.

\begin{itemize}
\item {\bf Partition 1}.
\begin{itemize}
\item {\bf Template}
\begin{itemize}
\item {\bf Q}: What is the spatial relation of the person to the <object> while the person <reference action>.
\item {\bf A}: <attribute>
\end{itemize}
\item {\bf Example}
\begin{itemize}
\item {\bf Q}: What is the spatial relation of the person to the closet while the person closing a closet?
\item {\bf A}: behind
\end{itemize}
\end{itemize}

\item {\bf Partition 2}.
\begin{itemize}
\item {\bf Template}
\begin{itemize}
\item {\bf Q}: What is the person doing to the <object> before the person <reference action>? 
\item {\bf A}: <attribute>
\end{itemize}
\item {\bf Example}
\begin{itemize}
\item {\bf Q}: What is the person doing to the blanket before the person putting a phone somewhere? 
\item {\bf A}: touching
\end{itemize}
\end{itemize}

\end{itemize}

\item {\bf Limitations}: The current setup is primarily designed for stationary objects and may not effectively assess dynamic scenarios or human actions, such as interactions with objects or motion-based tasks.

\item {\bf Recommendations}: A task generator includes compositional and contextual challenges that require deeper reasoning about object relation and recognition.
\vspace{-.25em}
\end{itemize}
\end{framed}
\end{figure}

\begin{figure}
\begin{framed}
 \centering
{\Large {\bf WhatActionVideoSceneGraphTaskGenerator}}
\begin{itemize}[leftmargin=*]

\item {\bf Basic Information}. 
\begin{itemize}
\item {\bf Task Type}.  VideoQA
\item {\bf Question Type}.  what object
\item {\bf Answer Type}.  object category
\item {\bf Image Type}. 3D tabletop image
\item {\bf The model capability to evaluate}. object recognition with / without reference
\end{itemize}

\item {\bf Source Data}. 
\begin{itemize}
\item rendering images of objects from Objaverse
\item Annotations regarding object category, attribute, and shape
\end{itemize}

\item {\bf Task Plan Schema}.  
\begin{itemize}
\item \textbf{question type}: \texttt{string}. The question type of these tasks will be "what action video".
\item \textbf{action}        : \texttt{string}. The target action that the person in the video performs.
\item \textbf{reference action}        : \texttt{string}. The reference action to locate the moment when a person is performing the target action.
\item \textbf{temporal reference type}      : \texttt{string}. Type of the temporal reference between the reference action and the moment when a person is performing the target action. Can be "before", "while", or "after"
\item \textbf{video scene graph id} : \texttt{string}. The identifier of the video scene graph. 
\end{itemize}

\item {\bf Partitions}.

\begin{itemize}
\item {\bf Partition 1}.
\begin{itemize}
\item {\bf Template}
\begin{itemize}
\item {\bf Q}: What action is the person doing while <reference action>?
\item {\bf A}: <action>
\end{itemize}
\item {\bf Example}
\begin{itemize}
\item {\bf Q}: What action is the person doing while laughing at something?
\item {\bf A}: sitting at a table
\end{itemize}
\end{itemize}

\end{itemize}

\item {\bf Limitations}: The current setup is primarily designed for stationary objects and may not effectively assess dynamic scenarios or human actions, such as interactions with objects or motion-based tasks.

\item {\bf Recommendations}: A task generator includes compositional and contextual challenges that require deeper reasoning about object relation and recognition.
\vspace{-.25em}
\end{itemize}
\end{framed}
\end{figure}

\clearpage 

\bibliographystyle{plain}
\bibliography{refs}

\end{document}